\documentclass[11pt]{article}

\usepackage[a4paper, total={6in, 8.5in}]{geometry}

\usepackage{amsmath,amsfonts}
\usepackage[subrefformat=parens,labelformat=parens]{subcaption}
 
\usepackage{tikz}
\usetikzlibrary{arrows,decorations.pathreplacing,calligraphy}
\usepackage{enumitem}
\usepackage{natbib}

\usepackage{bbm}
\usepackage[title,toc,titletoc,page]{appendix}

\usepackage[ruled,linesnumbered]{algorithm2e}
\SetKwInOut{Input}{Input}
\SetKwInOut{Inputs}{Inputs}
\SetKwInOut{Output}{Output}
\SetKwFor{While}{While}{}{}
\SetKwIF{If}{ElseIf}{Else}{If}{}{else if}{else}{}
\SetKw{Return}{Return}

\allowdisplaybreaks

\usepackage{color}
\definecolor{blue}{rgb}{0.0, 0.5, 0.5}

\usepackage{scalerel}

\usepackage{caption}

\usepackage{blkarray}
\newcommand\bigzero{\makebox(0,0){\text{\huge0}}}

\DeclareMathOperator{\Var}{Var}
\DeclareMathOperator{\Cov}{Cov}
\DeclareMathOperator{\An}{An}
\DeclareMathOperator{\Pa}{Pa}
\DeclareMathOperator{\pa}{pa}
\DeclareMathOperator{\De}{De}
\DeclareMathOperator{\PossDe}{PossDe}
\DeclareMathOperator{\PossPa}{PossPa}
\DeclareMathOperator{\PossAn}{PossAn}
\DeclareMathOperator{\Adj}{Adj}
\DeclareMathOperator{\PCO}{PCO}
\DeclareMathOperator{\Det}{Det}

\newcommand{\BlackBox}{\rule{1.5ex}{1.5ex}}  
\ifdefined\proof
    \renewenvironment{proof}{\par\noindent{\bf Proof\ }}{\hfill\BlackBox\\[2mm]}
\else
    \newenvironment{proof}{\par\noindent{\bf Proof\ }}{\hfill\BlackBox\\[2mm]}
\fi
\newtheorem{example}{Example} 
\newtheorem{theorem}{Theorem}
\newtheorem{lemma}[theorem]{Lemma} 
\newtheorem{proposition}[theorem]{Proposition} 
\newtheorem{remark}[theorem]{Remark}
\newtheorem{corollary}[theorem]{Corollary}
\newtheorem{definition}[theorem]{Definition}

\newcommand\given[1][]{\:#1\vert\:}

\newcommand{\mb}[1]{\mathbf{#1}}
\newcommand{\dsepp}{\perp_{d}}
\newcommand{\g}[1][G]{\mathcal{#1}}
\newcommand*\diff{\mathop{}\!\mathrm{d}}
\newcommand{\ind}{\perp\!\!\!\!\perp}
\newcommand{\E}{\mathbb{E}}

\newenvironment{proofof}[1][]{\noindent \textbf{Proof of #1.}}{\hfill\BlackBox}
\newenvironment{proofofnoqed}[1][]{\noindent \textbf{Proof of #1.}}{}

\newenvironment{proofsketch}{\noindent \textbf{Proof Sketch.}}{\hfill\BlackBox\\}

\long\def\acks#1{\vskip 0.3in\noindent{\large\bf Acknowledgments and Disclosure of Funding}\vskip 0.2in
\noindent #1}

\begin{document}

\title{Identifying Conditional Causal Effects in MPDAGs}

\author{Sara LaPlante  (slap47@uw.edu) \\
         Emilija Perkovi\'c  (perkovic@uw.edu) \\
         Department of Statistics\\
              University of Washington,\\
              Seattle, WA 98195, USA}
\date{}

\maketitle


\begin{abstract}
    We consider identifying a conditional causal effect when a graph is known up to a maximally oriented partially directed acyclic graph (MPDAG). An MPDAG represents an equivalence class of graphs that is restricted by background knowledge and where all variables in the causal model are observed. We provide three results that address identification in this setting: an identification formula when the conditioning set is unaffected by treatment, a generalization of the well-known do calculus to the MPDAG setting, and an algorithm that is complete for identifying these conditional effects.
\end{abstract}


\section{Introduction}
\label{sec:intro}

In finding causal effects, researchers may want to know the effect across an entire population, sometimes called a \textit{total} or \textit{unconditional causal effect}. For example, does free access to pre-kindergarten (pre-K) improve children's socio-emotional skills throughout elementary school \citep{moffett2023enrollment}? However, researchers may want to know the effect within subgroups of the population, or a \textit{conditional causal effect}. For instance, is there a subgroup of children who particularly benefit from free access to pre-K? Our research considers identifying these conditional effects from observational data.

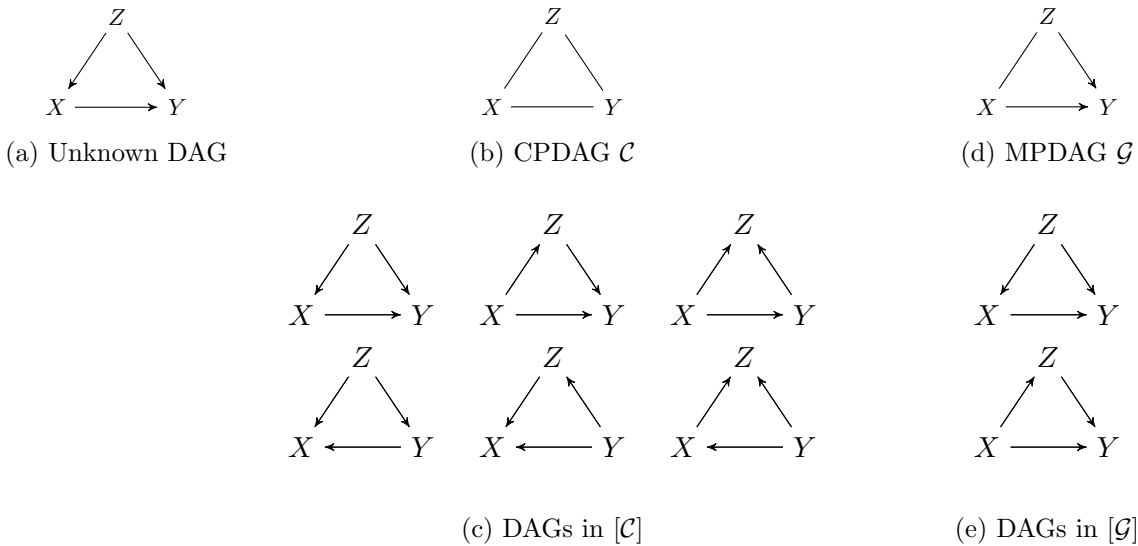
\begin{figure}
\tikzstyle{every edge}=[draw,>=stealth',->]
\newcommand\dagvariant[1]
{\begin{tikzpicture}[xscale=.8,yscale=.8]
    \node (x) at (0,0) {$X$};
    \node (y) at (2,0) {$Y$};
    \node (z) at (1,1.5) {$Z$};

    \draw (x) edge [-] (y);
    \draw (y) edge [-] (z);
    \draw (z) edge [-] (x);
    \draw #1;
\end{tikzpicture}}
\centering
\begin{subfigure}{.2\textwidth}
\centering
\begin{tikzpicture}[->,>=latex,shorten >=1pt,auto,node distance=0.8cm,scale=.8,transform shape]
\tikzstyle{state}=[inner sep=1pt, minimum size=12pt]
    \node (x) at (0,0) {$X$};
    \node (y) at (2,0) {$Y$};
    \node (z) at (1,1.5) {$Z$};

    \draw (x) edge  [->]  (y);
    \draw (z) edge  [->]  (y);
    \draw (z) edge  [->]  (x);
\end{tikzpicture}
\caption{Unknown DAG}
\label{fig:graphs-dag}
\vspace{4.98cm} 
\end{subfigure}
\hspace{.3cm}
\begin{subfigure}{.48\textwidth}
\centering
    \begin{tikzpicture}[->,>=latex,shorten >=1pt,auto,node distance=0.8cm,scale=.8,transform shape]
    \tikzstyle{state}=[inner sep=1pt, minimum size=12pt]
        \node (x) at (0,0) {$X$};
        \node (y) at (2,0) {$Y$};
        \node (z) at (1,1.5) {$Z$};

        \draw (x) edge [-] (y);
        \draw (y) edge [-] (z);
        \draw (z) edge [-] (x);
    \end{tikzpicture}
    \caption{CPDAG $\g[C]$}
    \label{fig:graphs-cpdag}
    \begin{center}
    \dagvariant{
        (x)  edge [->]  (y)
        (z)  edge [->]  (y)
        (z) edge [->] (x)
    }\hspace{.2cm}
    \dagvariant{
        (x)  edge [->]  (y)
        (z)  edge [->]  (y)
        (x) edge [->] (z)
    }\hspace{.2cm}
    \dagvariant{
        (x)  edge [->]  (y)
        (y)  edge [->]  (z)
        (x) edge [->] (z)
    }
    \\
    \dagvariant{
        (y)  edge [->]  (x)
        (z)  edge [->]  (y)
        (z) edge [->] (x)
    }\hspace{.2cm}
    \dagvariant{
        (y)  edge [->]  (x)
        (y)  edge [->]  (z)
        (z) edge [->] (x)
    }\hspace{.2cm}
    \dagvariant{
        (y)  edge [->]  (x)
        (y)  edge [->]  (z)
        (x) edge [->] (z)
    }
    \end{center}
    \caption{DAGs in $[\g[C]]$}
    \label{fig:graphs-dagsincpdag}
    \end{subfigure}
    \hspace{1.3cm}
    \begin{subfigure}{.17\textwidth}
    \centering
    \begin{tikzpicture}[->,>=latex,shorten >=1pt,auto,node distance=0.8cm,scale=.8,transform shape]
    \tikzstyle{state}=[inner sep=1pt, minimum size=12pt]
        \node (x) at (0,0) {$X$};
        \node (y) at (2,0) {$Y$};
        \node (z) at (1,1.5) {$Z$};

        \draw (x) edge [->] (y);
        \draw (z) edge [->] (y);
        \draw (z) edge [-] (x);
    \end{tikzpicture}
    \caption{MPDAG $\g$}
    \label{fig:graphs-mpdag}
    \begin{center}
    \dagvariant{
        (x)  edge [->]  (y)
        (z)  edge [->]  (y)
        (z) edge [->] (x)
    }\hspace{.2cm}
    \dagvariant{
        (x)  edge [->]  (y)
        (z)  edge [->]  (y)
        (x) edge [->] (z)
    }
    \end{center}
    \caption{DAGs in $[\g]$}
    \label{fig:graphs-dagsinmpdag}
    \end{subfigure}
    \caption{Illustrating an MPDAG. Let the DAG in \subref{fig:graphs-dag} represent an unknown causal model. We cannot learn this DAG from observational data alone, but we can learn the CPDAG in \subref{fig:graphs-cpdag}, which represents the equivalence class of DAGs shown in \subref{fig:graphs-dagsincpdag}. Adding background knowledge that $X$ and $Z$ precede $Y$ produces the MPDAG in \subref{fig:graphs-mpdag}, which represents the restricted class of DAGs shown in \subref{fig:graphs-dagsinmpdag}.}
    \label{fig:graphs}
\end{figure}

We assume knowledge of a causal graph. Though, in general, observational data alone are insufficient for learning a full directed acyclic graph (DAG), even when all variables in the model are observed. Thus, we consider a setting where the causal graph is known only up to an equivalence class of DAGs that can be learned from observational data \citep{spirtes2000causation, chickering2002optimal}. We allow the addition of expert knowledge as a way to restrict this class further. This restricted class of DAGs can be uniquely represented by a maximally oriented partially directed acyclic graph (MPDAG, \citealp{meek1995causal}), which is the focus of our work. For illustration, see Figure \ref{fig:graphs}.

Much of the literature focuses on identification in the unconditional setting, and many researchers consider identification through covariate adjustment. By definition, \textit{adjustment sets} identify the unconditional effect of $\mb{X}$ on $\mb{Y}$, since for any such set $\mb{S}$, it holds that $f(\mb{y} \,|\, do(\mb{x})) = \int f(\mb{y} \,|\, \mb{x}, \mb{s}) f(\mb{s}) \diff \mb{s}$, where $f$ is any density consistent with the model. Prior research provides methods of finding these sets by checking graphical relationships in a known graph---for example, a DAG \citep{pearl1995causal, shpitser2012validity}, an MPDAG \citep{perkovic2017interpreting}, or a \textit{partial ancestral graph} (PAG) that allows for latent variables \citep{van2014constructing, maathuis2015generalized, perkovic2018complete}. As an alternative to covariate adjustment, prior work also provides exact formulas for identifying unconditional effects when a DAG, MPDAG, or PAG is known \citep{perkovic2020identifying, jaber2018graphical}.

Other research focuses on identifying conditional effects, but in settings where graphs other than an MPDAG is known. For example, the well-known do calculus of \cite{pearl2009causality} offers a tool for transforming \textit{interventional densities} into observational densities based on \textit{d-separations} in a known DAG. \cite{zhang2008causal} extends this work by creating an analogous calculus for PAGs, and \cite{jaber2022causal} update the extension to make it complete. Based on this calculus, \cite{jaber2022causal} develop a sound and complete algorithm for conditional effect identification given a PAG. 

To the best of our knowledge, the only work that considers the identification of conditional effects when an MPDAG is known is that of \cite{laplante2024conditional}. These authors offer a graphical criterion for finding \textit{conditional adjustment sets}, where for any such set $\mb{S}$, it holds that $f(\mb{y} \,|\, do(\mb{x}), \mb{z}) = \int f(\mb{y} \,|\, \mb{x}, \mb{z}, \mb{s}) f(\mb{s} \,|\, \mb{z}) \diff \mb{s}$. But the authors note two limitations. First, there are identifiable effects where conditional adjustment sets do not exist (see Example \ref{ex:id-nocas} in Section \ref{sec:id-examples} below). Further, there are conditional adjustment sets that  \cite{laplante2024conditional}'s graphical criterion cannot find, since they require that the conditioning set must be unaffected by treatment (see Examples \ref{ex:alg-pass1}-\ref{ex:alg-pass2} Section \ref{sec:alg-examples} below).

Our results address this gap in the literature by providing three methods of identifying conditional causal effects in a setting where a causal MPDAG is known. We begin with an identification formula (Theorem \ref{thm:id-formula}), which provides an exact form of the interventional density in terms of observational densities. This result applies when the conditioning set is unaffected by treatment, which holds for the common case of pre-treatment covariates. Our second result (Theorem \ref{thm:do-calc-mpdag}) generalizes the well-known do calculus of \cite{pearl2009causality} to the MPDAG setting. That is, we provide a set of rules based on \textit{d-separations} in a causal graph with undirected edges that allow transformations of an interventional density under conditioning. We use this calculus in our final result: an identification algorithm (Algorithm \ref{alg:cidm}) that we show is complete for identifying conditional causal effects.

We organize this paper in the following way. Section \ref{sec:prelims} provides relevant preliminaries on graphs and related densities. Section \ref{sec:id} presents our conditional identification formula. Then Sections \ref{sec:do} and \ref{sec:alg} provide our do calculus for MPDAGs and conditional identification algorithm, respectively. We discuss the benefits, limitations, and possible usage of our results in Section \ref{sec:discussion}.


\section{Preliminaries}
\label{sec:prelims}

\textbf{Nodes, Edges, and Graphs.}
We use capital letters (e.g., $X$) to denote nodes in a graph as well as random variables that these nodes represent. We use bold capital letters (e.g., $\mb{X}$) to denote node sets. A \textit{graph} $\g=(\mb{V},\mb{E})$ consists of a set of nodes $\mb{V}$ and a set of edges $\mb{E}$. A \emph{directed graph} contains only directed edges ($\to$). A \emph{partially directed} graph may contain undirected edges ($-$) and directed edges ($\to$). An \textit{induced subgraph} $\g_{\mb{V'}} =(\mb{V'}, \mb{E'})$ of $\g$ consists of $\mb{V'} \subseteq \mb{V}$ and $\mb{E'} \subseteq \mb{E}$ where $\mb{E'}$ are all edges in $\mb{E}$ between nodes in $\mb{V'}$. In a partially directed graph, an edge is \textit{into} (\textit{out of}) a node $X$ if the edge is directed and has an arrowhead (tail) at $X$.

\textbf{Paths and Cycles.}
For disjoint node sets $\mb{X}$ and $\mb{Y}$, a \textit{path} from $\mb{X}$ to $\mb{Y}$ is a sequence of distinct nodes $\langle X, \dots,Y \rangle$ from some $X \in \mb{X}$ to some $Y \in \mb{Y}$ for which every pair of successive nodes is adjacent. An \textit{undirected path} is a path containing only undirected edges ($-$). A \textit{directed path} from $X$ to $Y$ is a path of the form $X \to \dots\to Y$.  A directed path from $X$ to $Y$ and the edge $Y\to X$ form a \textit{directed cycle}. A path from $\mb{X}$ to $\mb{Y}$ is \textit{proper} (w.r.t. $\mb{X}$) if only its first node is in $\mb{X}$. In a graph $\g$, the  path $p := \langle V_0, \dots, V_k \rangle$, $k \geq 1$, is \textit{possibly directed} if no edge $V_i \gets V_j, 0 \le i < j \le k$, is in $\g$ (\citealp{perkovic2017interpreting}).

\textbf{Subsequences, Subpaths, and Shields.}
A \textit{subsequence} of a path $p$ in a graph $\g$ is a path obtained by deleting a set of non-endpoint nodes from $p$ without changing the order of the remaining nodes. For a path $p = \langle X_1, \dots, X_m \rangle$, the \textit{subpath} from $X_i$ to $X_k$, for $1 \le i < k \le m$, is the path $p(X_i,X_k) = \langle X_i,X_{i+1}, \dots, X_{k}\rangle$. We use the notation $(-p)(X_k,X_i)$ to denote the path $\langle X_{k}, X_{k-1}, \dots, X_i \rangle$. The subpath $\langle X_{j-1}, X_j, X_{j+1} \rangle$, for $1<j<k$, is an \emph{unshielded triple} if $X_{j-1}$ and $X_{j+1}$ are not adjacent in $\g$. A path is \textit{unshielded} if all successive triples on the path are unshielded.

\textbf{Colliders, Non-colliders, and Definite Status Paths.}
The \textit{endpoints} of a path $p = \langle X_1, \dots, X_k \rangle$ in a graph $\g$ are the nodes $X_1$ and $X_k$. The node $X_i$, $1 < i < k$, is a \textit{collider} on $p$ if $p$ contains $X_{i-1} \to X_i \gets X_{i+1}$. The node $X_i$ is a \textit{definite non-collider} on $p$ if $p$ contains $X_{i-1} \gets X_i$ or $X_i \to X_{i+1}$, or if $\langle X_{i-1}, X_i, X_{i+1} \rangle$ is undirected and unshielded. A node is of \textit{definite status} on $p$ if it is an endpoint, collider, or definite non-collider on $p$. The path $p$ is of definite status if every node on $p$ is of definite status.

\textbf{Ancestral Relationships.}
If a graph $\g$ contains the edge $X \to Y$, then $X$ is a \textit{parent} of $Y$ in $\g$. If $\g$ contains $X - Y$ or $X \to Y$, then $X$ is a \textit{possible parent} of $Y$ in $\g$. If $\g$ contains a directed path from $X$ to $Y$, then $X$ is an \textit{ancestor} of $Y$ and $Y$ is a \textit{descendant} of $X$ in $\g$. If $\g$ contains a possibly directed path from $X$ to $Y$, then $X$ is a \textit{possible ancestor} of $Y$ and $Y$ is a \textit{possible descendant} of $X$ in $\g$. We use the convention that every node is an ancestor, descendant, possible ancestor, and possible descendant of itself. The sets of parents, ancestors, and descendants and the sets of possible parents, ancestors, and descendants of $X$ in $\g$ are denoted by $\Pa(X,\g)$, $\An(X,\g)$, $\De(X,\g)$, $\PossPa(X,\g)$, $\PossAn(X,\g)$, and $\PossDe(X,\g)$, respectively. For a set of nodes $\mb{X}$, we let $\An(\mb{X},\g) = \cup_{X \in \mb{X}} \An(X,\g)$, with analogous definitions for $\De(\mb{X},\g)$, $\PossAn(\mb{X},\g)$, and $\PossDe(\mb{X},\g)$. Unconventionally, we define $\Pa(\mb{X},\g) =(\cup_{X \in \mb{X}} \Pa(X,\g)) \setminus \mb{X}$.

\textbf{DAGs and MPDAGs.} 
A directed graph without directed cycles is a \textit{directed acyclic graph (DAG)}. A partially directed graph without directed cycles is a \textit{partially directed acyclic graph (PDAG)}. All DAGs over a node set $\mb{V}$ with the same adjacencies and unshielded colliders can be uniquely \textit{represented by} a \textit{completed PDAG} (CPDAG). These DAGs form a Markov equivalence class with the same set of d-separations. A \textit{maximally oriented PDAG} (MPDAG) is formed by taking a CPDAG, adding background knowledge (by directing undirected edges), and completing \cite{meek1995causal}'s orientation rules. We say a DAG is \textit{represented by} an MPDAG $\g$ if it has the same nodes, adjacencies, and directed edges as $\g$ and if it has no additional unshielded colliders from those in $\g$. The set of such DAGs---denoted by $[\g]$---forms a restriction of the Markov equivalence class so that all DAGs in $[\g]$ have same set of d-separations. Note that if $\g$ has the edge $A - B$, then $[\g]$ contains at least one DAG with $A \to B$ and one DAG with $A \gets B$ \citep{meek1995causal}. Further, note that all DAGs and CPDAGs are MPDAGs.

\textbf{Markov Compatibility and Positivity Assumption.}
An \textit{observational density} $f(\mb{v})$ is \textit{Markov compatible} with a DAG $\g[D] = (\mb{V},\mb{E})$ if $f(\mb{v})= \prod_{V_i \in \mb{V}}f(v_i|\pa(v_i,\g[D]))$. If $f(\mb{v})$ is Markov compatible with a DAG $\g[D]$, then it is Markov compatible with every DAG that is Markov equivalent to $\g[D]$ \citep{pearl2009causality}. Hence, we say that a density is \textit{Markov compatible} with an MPDAG $\g$ if it is Markov compatible with a DAG represented by $\g$. Throughout, we assume positivity. That is, we only consider densities that satisfy $f(\mb{v})>0$ for all valid values of $\mb{V}$ \citep{kivva2023identifiability}. Note that since, $f(\mb{v})= \prod_{V_i \in \mb{V}}f(v_i|\pa(v_i,\g[D]))$, assuming   $f(\mb{v})>0$ is equivalent to assuming $f(v_i|\pa(v_i, \g)) > 0$ for all $V_i \in \mb{V}$.

\textbf{D-connection, D-separation, and Probabilistic Implications.}
Let $\mb{X}$, $\mb{Y}$, and $\mb{Z}$ be pairwise disjoint node sets in a graph $\g$. A definite status path \textit{p} from $\mb{X}$ to $\mb{Y}$ is \textit{d-connecting} (or \textit{open}) given $\mb{Z}$ if every definite non-collider on $p$ is not in $\mb{Z}$ and every collider on $p$ has a descendant in $\mb{Z}$. Otherwise, $p$ is \textit{blocked} given $\mb{Z}$. If all definite status paths between $\mb{X}$ and $\mb{Y}$ in $\g$ are blocked given $\mb{Z}$, then $\mb{X}$ is \textit{d-separated} from $\mb{Y}$ given $\mb{Z}$ in $\g$ and we write $(\mb{X} \dsepp \mb{Y} | \mb{Z})_{\g}$. This d-separation implies that $\mb{X}$ and $\mb{Y}$ are conditionally independent given $\mb{Z}$ in any \textit{observational density} that is Markov compatible with $\g$ \citep{lauritzen1990independence, henckel2022graphical}. 

\textbf{Causal Graphs.}
An MPDAG $\g$ is a \textit{causal MPDAG} if every edge $V_i \to V_j$ in $\g$ represents a direct causal effect of $V_i$ on $V_j$ and if every edge $V_i - V_j$ represents a direct causal effect of unknown direction (either $V_i$ causes $V_j$ or $V_j$ causes $V_i$). In a causal MPDAG, any directed path is \textit{causal}, any possibly directed path is \textit{possibly causal}, and any other path is \textit{non-causal}.

\textbf{Causal Ordering.} Let $\mb{X} = \{X_1, \dots, X_k\}$, $k \ge 1$, be a node set in a causal DAG $\g[D]$. We say that $X_1 < \dots < X_k$ is a \textit{total causal ordering} of $\mb{X}$ consistent with $\g[D]$ if for every $X_i, X_j \in \mb{X}$ such that $X_i < X_j$ and such that $X_i$ and $X_j$ are adjacent in $\g[D]$, then $\g[D]$ contains $X_i \to X_j$. There can be more than one total causal ordering of a set of nodes in a DAG. For example, in the DAG $X_b \gets X_a \to X_c$, both $X_a < X_b < X_c$ and $X_a < X_c < X_b$ are total causal orderings consistent with the DAG.

Now let $\mb{X}$ be a node set in a causal MPDAG $\g$. Since $\g$ may contain undirected edges, there is generally no total causal ordering of $\mb{X}$ consistent with $\g$. Instead, we define a \textit{partial causal ordering} of $\mb{X}$ consistent with $\g$ to be an ordering of pairwise disjoint node sets $\mb{A_1}, \dots, \mb{A_k}$, $k \ge 1$, $\cup_{i=1}^k \mb{A_i} = \mb{X}$, that satisfies the following: for every $i,j \in \{1, \dots, k\}$ such that $\mb{A_i} < \mb{A_j}$ and there is an edge between $A_i \in \mb{A_i}$ and $A_j \in \mb{A_j}$ in $\g$, then $\g$ contains $A_i \to A_j$.

\textbf{Consistency.}
Let $f(\mb{v})$ be an observational density over a set of variables $\mb{V}$. The notation $do(\mb{X} = \mb{x})$, or $do(\mb{x})$ for short, represents an outside intervention that sets $\mb{X} \subseteq \mb{V}$ to fixed values $\mb{x}$. An \textit{interventional density} $f(\mb{v}|do(\mb{x}))$ is a density resulting from such an intervention.

Let $\mb{F^*}$ denote the set of all interventional densities $f(\mb{v}|do(\mb{x}))$ such that $\mb{X} \subseteq \mb{V}$ (including $\mb{X} = \emptyset$). A causal DAG $\g[D] = (\mb{V,E})$ is a \textit{causal Bayesian network compatible with} $\mb{F^*}$ if and only if for all $f(\mb{v} | do(\mb{x})) \in \mb{F^*}$, the following \textit{truncated factorization} holds:
\begin{align}
    f(\mb{v} | do(\mb{x})) = \prod_{V_i \in \mb{V} \setminus \mb{X}} f(v_i|\pa(v_i,\g[D])) \mathbbm{1}(\mb{X} = \mb{x}) \label{eq:trunc-fact}
\end{align}
\citep{pearl2009causality, bareinboim2012local}. We say an interventional density is \textit{consistent} with a causal DAG $\g[D]$ if it belongs to a set of interventional densities $\mb{F^*}$ such that $\g[D]$ is compatible with $\mb{F^*}$. Note that any observational density that is Markov compatible with $\g[D]$ is consistent with $\g[D]$. We say an interventional density is \textit{consistent} with a causal MPDAG $\g$ if it is consistent with each DAG in $[\g]$---were the DAG to be causal. Following convention, we omit $\mathbbm{1}(\mb{X} = \mb{x})$ from identifying expressions of interventional densities below.

\textbf{Identifiability.}
Let $\mb{X}$, $\mb{Y}$, and $\mb{Z}$ be pairwise disjoint node sets in a causal MPDAG $\g = (\mb{V,E})$, and let $\mb{F^*_i} = \{ f_i(\mb{v}|do(\mb{x'})) : \mb{X'} \subseteq \mb{V} \}$ be a set with which a DAG $\g[D]_i \in [\g]$ is compatible---were $\g[D]_i$ to be causal. Then the conditional causal effect of $\mb{X}$ on $\mb{Y}$ given $\mb{Z}$ is \textit{identifiable} in $\g$ if for any $\mb{F^*_1}, \mb{F^*_2}$ such that $f_1(\mb{v}) = f_2(\mb{v})$, we have $f_1(\mb{y}|do(\mb{x}), \mb{z}) = f_2(\mb{y}|do(\mb{x}), \mb{z})$ \citep{pearl2009causality}.


\section{Identification Formula}
\label{sec:id}

In this section, we provide our first result: a formula for identifying a conditional causal effect when a causal MPDAG is known and when the conditioning set is unaffected by treatment. This formula relies on concepts found in prior research on MPDAGs, which we review briefly below. After presenting our formula, we explore its use through special cases and examples. We close this section by providing a necessary and sufficient condition for identifying conditional effects in this setting.


\subsection{Review of the PCO Algorithm}
\label{sec:id-pco}

The identification formula we present in Section \ref{sec:id-formula} relies on output from the \textit{Partial Causal Ordering} (PCO) Algorithm of \cite{perkovic2020identifying}. We provide the full algorithm in Appendix \ref{app:existing}, but offer a brief description here for broad understanding of our results. The PCO Algorithm inputs a set of nodes from an MPDAG and outputs an ordered partition of that node set, which we call an \textit{ordered bucket decomposition}. That is, if $\mb{D}$ is a node set in an MPDAG $\g$, then $\text{PCO}(\mb{D},\g) = (\mb{B_1}, \dots, \mb{B_k})$, $k \ge 1$, where $\mb{B_1}, \dots, \mb{B_k}$ are pairwise disjoint subsets of $\g[D]$ that follow a \textit{partial causal ordering} consistent with $\g$ (see Section \ref{sec:prelims}). For clarity, we formally define \textit{buckets} and their ordered decompositions below.

\begin{definition}
\label{def:bucket}
{\normalfont (\textbf{Bucket}; \citealp{perkovic2020identifying})}
    Let $\mb{B}$ and $\mb{D}$ be node sets in an MPDAG $\g$ such that $\mb{B} \subseteq \mb{D}$. Then $\mb{B}$ is a bucket in $\mb{D}$ given $\g$ if it is the intersection of $\mb{D}$ with a maximal undirected connected set in $\g$. That is,
    \begin{itemize}
        \item $\g$ contains a undirected path from $B_i$ to $B_j$ for every $B_i, B_j \in \mb{B}$, and  
        \item $\g$ contains no undirected path from $\mb{B}$ to any node in $\mb{D} \setminus \mb{B}$.
    \end{itemize}
\end{definition}

\begin{definition}[Ordered Bucket Decomposition]
\label{def:obd}
    Let $\mb{D}$ be a node set in an MPDAG $\g$. Then $(\mb{B_1}, \dots, \mb{B_k}), k \ge 1$, is an ordered bucket decomposition of $\mb{D}$ in $\g$ if
    \begin{itemize}
        \item $\mb{B_i}$ is a bucket in $\mb{D}$ given $\g$ for all $i \in \{ 1, \dots, k\}$.
        \item $\mb{B_1} < \dots < \mb{B_k}$ is a partial causal ordering of $\mb{D}$ consistent with $\g$ so that
        \begin{itemize}
            \item $\cup_{i=1}^k \mb{B_i} = \mb{D}$, and
            \item any edge in $\g$ between $B_i \in \mb{B_i}$ and $B_j \in \mb{B_j}$ for $i,j \in \{1, \dots, k\}$, $i<j$, must take the form $B_i \to B_j$.
        \end{itemize}
    \end{itemize}
\end{definition}


\subsection{Identification Formula}
\label{sec:id-formula}

We turn to our identification formula, where we consider the conditional causal effect of a set of treatments $\mb{X}$ on outcomes $\mb{Y}$ given covariates $\mb{Z}$ that are unaffected by treatment.

\begin{theorem}[Conditional Identification Formula]
\label{thm:id-formula}
    Let $\mb{X}$, $\mb{Y}$, and $\mb{Z}$ be pairwise disjoint node sets in a causal MPDAG $\g =(\mb{V,E})$, where $\mb{Z} \cap \PossDe(\mb{X},\g) = \emptyset$ and where there is no proper possibly causal path from $\mb{X}$ to $\mb{Y}$ in $\g$ that starts with an undirected edge. Then for any density $f$ consistent with $\g$,
    \begin{align}
        f(\mb{y} \,|\, do(\mb{x}),\mb{z}) = \int \prod_{i\in \mb{I_E}} f(\mb{b_i} \,|\, \pa(\mb{b_i},\g)) \prod_{i \in \mb{I_N}} f(\mb{b_i} \,|\, \mb{b_i^N}, \mb{z}) \ \diff \mb{b}, \label{eq:id-formula}
    \end{align}
    for ${\mb{B} =\An(\mb{Y}, \g_{\mb{V} \setminus \mb{X}}) \setminus (\mb{Z} \cup \mb{Y})}$; ${(\mb{B_1}, \dots, \mb{B_k}) = \PCO(\An(\mb{Y}, \g_{\mb{V} \setminus \mb{X}}) \setminus \mb{Z}, \g)}$; ${\mb{B_0} = \emptyset}$; and $\mb{B_i^N}= \cup_{j=0}^{i-1}\mb{B_j} \setminus \PossDe(\mb{X},\g)$. For ${i \in \{1, \dots, k\}}$, let ${i \in \mb{I_E}}$ when ${\mb{Z} \cap \PossDe(\mb{B_i}, \g) = \emptyset}$ and let ${i \in \mb{I_N}}$ otherwise.
\end{theorem}

\begin{proofofnoqed}[Theorem \ref{thm:id-formula}]
    The first two equalities below follow by the law of total probability and the chain rule, respectively. The third equality follows from our Lemma \ref{lem:algo-4} (App. \ref{app:id-formula}), which we discuss following this proof.
    \begingroup
    \allowdisplaybreaks
    \begin{align*}
        \hspace{.65in}
        f(\mb{y} \,|\, do(\mb{x}), \mb{z})
            &= \int f(\mb{b}, \mb{y} \,|\, do(\mb{x}), \mb{z}) \diff \mb{b}&\\
            &= \int \prod_{i=1}^{k} f(\mb{b_i} \,|\, \mb{b_{i-1}}, \dots, \mb{b_0}, do(\mb{x}), \mb{z}) \diff \mb{b} &\\
            &= \int \prod_{i\in \mb{I_E}} f(\mb{b_i} \,|\, \pa(\mb{b_i},\g)) \prod_{i \in \mb{I_N}} f(\mb{b_i} \,|\, \mb{b_i^N}, \mb{z}) \diff \mb{b}. &\hspace{1.10in} \BlackBox
    \end{align*}
    \endgroup
\end{proofofnoqed}

\begin{proofsketch}
    Lemma \ref{lem:algo-4} and its proof can be found in Appendix \ref{app:id-formula}, but we provide an overview here for intuition. This lemma requires the same assumptions as Theorem \ref{thm:id-formula}---that $\mb{Z} \cap \PossDe(\mb{X},\g) = \emptyset$ and no proper possibly causal path from $\mb{X}$ to $\mb{Y}$ starts undirected. Under these assumptions, the lemma claims
    \begin{align}
        f(\mb{b_i} \,|\, \mb{b_{i-1}}, \dots, \mb{b_0}, do(\mb{x}),\mb{z}) = 
        \begin{cases} 
            f(\mb{b_i} \,|\, \pa(\mb{b_i},\g)) & \mb{Z} \cap \PossDe(\mb{B_i}, \g) = \emptyset\\
            f(\mb{b_i} \,|\, \mb{b_i^N}, \mb{z}) & \mb{Z} \cap \PossDe(\mb{B_i}, \g) \neq \emptyset.
        \end{cases} \label{lem:algo-4-claim}
    \end{align}
    The proof of this lemma relies on applying Rules 1-3 of \citeauthor{pearl2009causality}'s (\citeyear{pearl2009causality}) do calculus (Theorem \ref{thm:do-calc}, App. \ref{app:existing}) to an arbitrary DAG $\g[D] \in [\g]$. To show the top equality of \eqref{lem:algo-4-claim} holds, we use Rules 1 and 3 to restrict the conditioning set---including variables in the do intervention---to variables also in $\Pa(\mb{B_i},\g)$. Applying Rule 2 leaves a conditioning set of $\big(\cup_{j=0}^{i-1} \mb{B_j} \cup \mb{Z} \cup \mb{X} \big) \cap \Pa(\mb{B_i}, \g)$, which is simply $\Pa(\mb{B_i},\g)$. To show that the d-separations for these three rules hold, we assume, for sake of contradiction, that under a specific conditioning set, there is an open path in $\g[D]$ between $\mb{B_i}$ and variables in $\big(\cup_{j=1}^{i-1} \mb{B_j} \cup \mb{Z} \big) \setminus \Pa(\mb{B_i}, \g)$ and $\mb{X}$. Informally, we rely on the fact that $\g$ cannot contain a possibly causal path from $\mb{B_i}$ to these variables. We use similar logic to show the lower equality of \eqref{lem:algo-4-claim} holds, by applying Rules 1 and 3 to reduce the conditioning set to $\mb{B_i^N} \cup \mb{Z}$.
\end{proofsketch}

We want to highlight that Appendix \ref{app:mpdags} includes broad results for MPDAGs used in the proofs of Lemma \ref{lem:algo-4}. To keep the focus of our paper on our three identification methods, we do not include these results in the main text. However, we recommend them to researchers in this area, since the results may be useful for future work on MPDAGs. As a brief review, Appendix \ref{app:mpdags} includes an alternate definition for possibly causal paths, a claim about concatenating casual and possibly causal paths, examples when possibly causal paths in the MPDAG imply causal paths in the equivalence class of DAGs, and restrictions on possibly causal paths among buckets in the MPDAG.


\subsection{Special Cases}

Before providing examples, we return to the statement of Theorem \ref{thm:id-formula} to consider two settings where Equation \eqref{eq:id-formula} can be replaced with a simplified form. For the first setting, we compare our results to those of \cite{perkovic2020identifying}. Under similar assumptions, the author provides the following identification formula for the \textit{unconditional} causal effect of $\mb{X}$ on $\mb{Y}$:
\begin{align}
    f(\mb{y} \,|\, do(\mb{x})) = \int \prod_{i=1}^k f(\mb{b_i} \,|\, \pa(\mb{b_i},\g)) \diff \mb{b}. \label{eq:id-formula-unconditional}
\end{align}
To compare Equations \eqref{eq:id-formula} and \eqref{eq:id-formula-unconditional}, we note that Theorem \ref{thm:id-formula} holds for any conditioning set $\mb{Z}$, including when $\mb{Z} = \emptyset$. In this case, ${\mb{Z} \cap \PossDe(\mb{B_i}, \g) = \emptyset}$ for all $i \in \{1, \dots, k\}$, and thus, Equation \eqref{eq:id-formula} reduces to the exact form of the unconditional identification formula shown in Equation \eqref{eq:id-formula-unconditional}.


For the second setting, suppose, as in Theorem \ref{thm:id-formula}, that $\mb{Z} \cap \PossDe(\mb{X},\g) = \emptyset$. But instead of excluding a subset of possibly causal paths from $\mb{X}$ to $\mb{Y}$, consider a stronger assumption that $\mb{Y} \cap \PossDe(\mb{X},\g) = \emptyset$. In this setting, identification of the conditional effect reduces to an application of Rule 3 of the do calculus (Theorem \ref{thm:do-calc}, App. \ref{app:existing}). We show this formally below.

\begin{theorem}
\label{thm:id-rule3}
    Let $\mb{X}$, $\mb{Y}$, and $\mb{Z}$ be pairwise disjoint node sets in a causal MPDAG $\g$, where $\mb{Z} \cap \PossDe(\mb{X},\g) = \emptyset$ and $\mb{Y} \cap \PossDe(\mb{X},\g) = \emptyset$. Then for any $f$ consistent with $\g$,
    \begin{align}
        f(\mb{y} \,|\, do(\mb{x}),\mb{z}) = f(\mb{y} \,|\, \mb{z}). \label{eq:id-formula-rule3}
    \end{align}
\end{theorem}

\begin{proofof}[Theorem \ref{thm:id-rule3}]
    Let $\g[D]$ be an arbitrary DAG in $[\g]$. The result holds by Rule 3 of the do calculus (Theorem \ref{thm:do-calc}, App. \ref{app:existing}) if we show $(\mb{Y} \dsepp \mb{X} \,|\, \mb{Z})_{\g[D]_{\overline{\mb{W}}}}$ for $\mb{W} := \mb{X} \setminus \An(\mb{Z}, \g[D])$. Note that $\De(\mb{X}, \g[D]) \subseteq \PossDe(\mb{X}, \g)$ (Lemma \ref{lem:pc-imply2}, App. \ref{app:mpdags}) so that $\mb{Z} \cap \PossDe(\mb{X}, \g) = \emptyset$ implies $\mb{Z} \cap \De(\mb{X}, \g[D]) = \emptyset$. Thus, $\mb{W} = \mb{X}$ and we must show $(\mb{Y} \dsepp \mb{X} \,|\, \mb{Z})_{\g[D]_{\overline{\mb{X}}}}$.

    Let $p$ be an arbitrary path in $\g[D]_{\overline{\mb{X}}}$ from $X \in \mb{X}$ to $Y \in \mb{Y}$. By definition of $\g[D]_{\overline{\mb{X}}}$, $p$ begins $X \to$. But $p$ cannot be causal, since this would imply $Y \in \De(\mb{X}, \g[D]) \subseteq \PossDe(\mb{X}, \g)$, which is a contradiction. Thus, $p$ must have a collider. Form a set $\mb{S}$ with the collider on $p$ that is closest to $X$ on $p$ and all of its descendants in $\g[D]_{\overline{\mb{X}}}$. Since $\mb{S} \subseteq \De(X, \g[D]_{\overline{\mb{X}}}) \subseteq \De(\mb{X}, \g[D]) \subseteq \PossDe(\mb{X}, \g)$ and since $\mb{Z} \cap \PossDe(\mb{X}, \g) = \emptyset$, then $p$ must be d-separated given $\mb{Z}$.
\end{proofof}


\subsection{Examples}
\label{sec:id-examples}
The examples below demonstrate how to use Theorems \ref{thm:id-formula} and \ref{thm:id-rule3} to identify a conditional causal effect. As noted in Section \ref{sec:intro}, \cite{laplante2024conditional} already provide an identification method in the MPDAG setting using what the authors call \textit{conditional adjustment sets}. But as they point out, these sets are not present in every MPDAG. Our identification formula fills in this gap when the conditioning set is unaffected by treatment. To highlight our contribution, Example \ref{ex:id-cas} below considers a setting where a conditional adjustment set exists. We contrast this with Example \ref{ex:id-nocas}, where there is no such set. Example \ref{ex:id-rule3} shows a straightforward application of Theorem \ref{thm:id-rule3}.

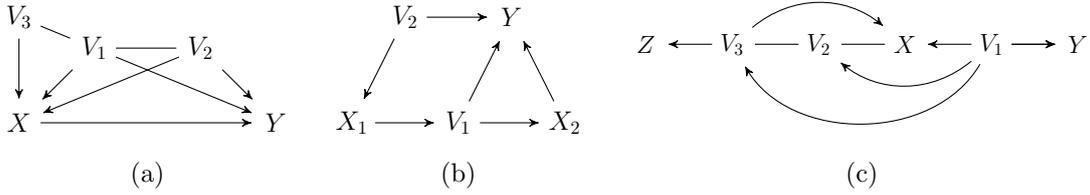
\begin{figure}
    \centering
    \begin{subfigure}{.26\textwidth}
        \vspace{1cm}
        \centering
        \begin{tikzpicture}[>=stealth',shorten >=1pt,auto,node distance=2cm,main node/.style={minimum size=0.8cm,font=\sffamily\Large\bfseries},scale=0.7,transform shape]
        \node[main node]   (X)                        {$X$};
        \node[main node]   (V1) [above right of = X]  {$V_1$};
        \node[main node]   (V2) [right of       = V1] {$V_2$};
        \node[main node]   (V3) [above of       = X]  {$V_3$};
        \node[main node]   (Y)  [below right of = V2] {$Y$};
        \draw[->]   (V3) edge (X);
        \draw[-]    (V3) edge (V1);
        \draw[->]   (X)  edge (Y);
        \draw[-]    (V1) edge (V2);
        \draw[->]   (V2) edge (X);
        \draw[->]   (V2) edge (Y);
        \draw[->]   (V1) edge (X);
        \draw[->]   (V1) edge (Y);
        \end{tikzpicture}
        \caption{}
        \label{fig:ex-id-cas}
    \end{subfigure}
    \begin{subfigure}{.26\textwidth}
        \centering
        \begin{tikzpicture}[>=stealth',shorten >=1pt,auto,node distance=2cm,main node/.style={minimum size=0.8cm,font=\sffamily\Large\bfseries},scale=.7,transform shape]
        \node[main node]   (X1)                  {$X_1$};
        \node[main node]   (V1) [right of = X1]  {$V_1$};
        \node[main node]   (Y)  at (3,2)         {$Y$};
        \node[main node]   (X2) [right of = V1]  {$X_2$};
        \node[main node]   (V2) at (1,2)         {$V_2$};
        \draw[->]   (X1) edge (V1);
        \draw[->]   (V1) edge (Y);
        \draw[->]   (V1) edge (X2);
        \draw[->]   (X2) edge (Y);
        \draw[->]   (V2) edge (X1);
        \draw[->]   (V2) edge (Y);
        \end{tikzpicture}
        \caption{}
        \label{fig:ex-id-nocas}
    \end{subfigure}
    \begin{subfigure}{.42\textwidth}
        \centering
        \begin{tikzpicture}[>=stealth',shorten >=1pt,auto,node distance=2cm,main node/.style={minimum size=0.8cm,font=\sffamily\Large\bfseries},scale=.65,transform shape]
        \node[main node]   (Z)  at (0,0)    {$Z$};
        \node[main node]   (V3) at (1.75,0)  {$V_3$};
        \node[main node]   (V2) at (3.5,0)    {$V_2$};
        \node[main node]   (X)  at (5.25,0)  {$X$};
        \node[main node]   (V1) at (7,0)    {$V_1$};
        \node[main node]   (Y)  at (8.75,0)  {$Y$};
        \draw[->]   (V1) edge (X);
        \draw[->]   (V1) edge (Y);
        \draw[-]    (X)  edge (V2);
        \draw[-]    (V2) edge (V3);
        \draw[->]   (V3) edge (Z);
        \draw[->]   (V1) edge (Y);
        \draw[->]   (V3) edge[bend left=40] node [left] {} (X);
        \draw[->]   (V1) edge[bend left=40] node [left] {} (V2);
        \draw[->]   (V1) edge[bend left=60] node [left] {} (V3);
        \end{tikzpicture}
        \caption{}
        \label{fig:ex-id-rule3}
    \end{subfigure}
    \caption{MPDAGs used in Examples \ref{ex:id-cas}-\ref{ex:id-rule3}}
    \label{fig:ex1}
\end{figure}


\begin{example}[Conditional Adjustment]
\label{ex:id-cas}
    Let $\g$ be the causal MPDAG in Figure \ref{fig:ex1}\subref{fig:ex-id-cas}, and let $\mb{X} = \{X\}$, $\mb{Y}=\{Y\}$, and $\mb{Z}=\{V_1\}$. To use Theorem \ref{thm:id-formula}, we confirm $\mb{Z} \cap \PossDe(\mb{X}, \g) = \emptyset$ and no possibly causal path from $\mb{X}$ to $\mb{Y}$ starts undirected. Then we find $\mb{B_1}=\{V_2\}$ (where $1 \in \mb{I_N}$), $\mb{B_2}=\{Y\}$ (where $2 \in \mb{I_E}$), and $\mb{B}= \{V_2\}$. Thus for $f$ consistent with $\g$,
    \vskip -.6cm
    \begin{align*}
        f(\mb{y} \,|\, do(\mb{x}),\mb{z})
            &=\scaleobj{.6}{\int} \,f\big(\mb{b_2} \,|\, \pa(\mb{b_2},\g)\big)\, f(\mb{b_1} \,|\, \mb{b_1^N},\mb{z}) \diff \mb{b} \\
            &=\scaleobj{.6}{\int} \,f(y \,|\, x, v_1, v_2) \,f(v_2 \,|\, v_1) \diff v_2.
    \end{align*}
    Compare this to Example 2 of \cite{laplante2024conditional}, where the authors find the identical form of $f(\mb{y}|do(\mb{x}),\mb{z})$ by showing that $\mb{S} := \{V_2\}$ is a conditional adjustment set.
\end{example}


\begin{example}[No Conditional Adjustment]
\label{ex:id-nocas}
    Let $\g$ be the causal DAG (and therefore, MPDAG) in Figure \ref{fig:ex1}\subref{fig:ex-id-nocas}, and let $\mb{X} = \{X_1, X_2\}$, $\mb{Y}=\{Y\}$, and $\mb{Z}=\{V_2\}$. To use Theorem \ref{thm:id-formula}, we confirm $\mb{Z} \cap \PossDe(\mb{X}, \g) = \emptyset$ and no possibly causal path from $\mb{X}$ to $\mb{Y}$ starts undirected. Then we find $\mb{B_1}=\{V_1\}$, $\mb{B_2}=\{Y\}$ (where $1,2 \in \mb{I_E}$), and $\mb{B}= \{V_1\}$. Thus for $f$ consistent with $\g$,
    \vskip -.6cm
    \begin{align}
        f(\mb{y} \,|\, do(\mb{x}),\mb{z})
            &=\scaleobj{.6}{\int} \,f\big(\mb{b_1} \,|\, \pa(\mb{b_1},\g)\big) \,f\big(\mb{b_2} \,|\, \pa(\mb{b_2},\g)\big) \diff \mb{b} \nonumber\\
            &=\scaleobj{.6}{\int} \,f(v_1 \,|\, x_1) \,f(y \,|\, x_2, v_1, v_2) \diff v_1. \nonumber
    \end{align}
    Note that \cite{laplante2024conditional} show there is no conditional adjustment set in this case (see their Figure 4).
\end{example}


\begin{example}[Using Theorem \ref{thm:id-rule3}]
\label{ex:id-rule3}
    Let $\g$ be the causal MPDAG in Figure \ref{fig:ex1}\subref{fig:ex-id-rule3}, and let $\mb{X} = \{X\}$, $\mb{Y}=\{Y\}$, and $\mb{Z}=\{Z\}$. To use Theorem \ref{thm:id-rule3}, we confirm $\mb{Y} \cap \PossDe(\mb{X}, \g) = \emptyset$ and $\mb{Z} \cap \PossDe(\mb{X}, \g) = \emptyset$. Thus for $f$ consistent with $\g$,
    \begin{align*}
        f(\mb{y} \,|\, do(\mb{x}),\mb{z})
            &= f(y \,|\, z).
    \end{align*}    
\end{example}


\subsection{Identifiability Condition}
\label{sec:id-condition}

We close the discussion of our identification formula (Theorem \ref{thm:id-formula}) by reflecting further on its second condition---the restriction on paths between $\mb{X}$ and $\mb{Y}$. The result below shows that this condition is necessary and sufficient for identification when the conditioning set is unaffected by treatment.

\begin{proposition}[Identifiability Condition, Restricted $\mb{Z}$]
\label{prop:id-condition}
    Let $\mb{X}$, $\mb{Y}$, and $\mb{Z}$ be pairwise disjoint node sets in a causal MPDAG $\g$ such that $\mb{Z} \cap \PossDe(\mb{X},\g) = \emptyset$. Then the conditional causal effect of $\mb{X}$ on $\mb{Y}$ given $\mb{Z}$ is identifiable in $\g$ if and only if there is no proper possibly causal path from $\mb{X}$ to $\mb{Y}$ in $\g$ that starts with an undirected edge.
\end{proposition}

\begin{proofsketch}
    The proof of Proposition \ref{prop:id-condition} can be found in Appendix \ref{app:id-condition}, but we provide an outline here for intuition. Note that $\Leftarrow$ follows from Theorem \ref{thm:id-formula}. We show $\Rightarrow$ through its contraposition. Thus, suppose there is a proper possibly causal path from $\mb{X}$ to $\mb{Y}$ in $\g$ that starts undirected. We show in a separate result (Lemma \ref{lem:twopaths-corrollary}, App. \ref{app:id-condition}) that this path implies there are DAGs $\g[D]^1,\g[D]^2 \in [\g]$ with corresponding paths $X \to \dots \to Y$ and $X \gets V_1 \to \dots \to Y$. To show the conditional effect is not identifiable, we find $\{ f_1(\mb{v}|do(\mb{x'})) : \mb{X'} \subseteq \mb{V} \}$ and $\{ f_2(\mb{v}|do(\mb{x'})) : \mb{X'} \subseteq \mb{V} \}$ that are compatible with $\g[D]^1$ and $\g[D]^2$, where $f_1(\mb{v})=f_2(\mb{v})$ but $f_1(\mb{y} | do(\mb{x}), \mb{z}) \neq f_2(\mb{y} | do(\mb{x}), \mb{z})$. We construct these sets using structural equation models (SEMs) based on DAGs with the same nodes as $\g[D]^i$ but with only the edges from the path of interest. To complete the proof, we show $E_1[Y | do(\mb{X}=\mb{1}), \mb{Z}] \neq E_2[Y | do(\mb{X}=\mb{1}), \mb{Z}]$ using the relevant DAGs, the do calculus (Theorem \ref{thm:do-calc}, App. \ref{app:existing}), and Wright's Rule (Lemma \ref{lem:wright}).
\end{proofsketch}


\section{Do Calculus for MPDAGs}
\label{sec:do}

The do calculus of \citeauthor{pearl2009causality} (\citeyear{pearl2009causality}, see Theorem \ref{thm:do-calc}, App. \ref{app:existing}) consists of three rules that justify transformations of an interventional density. These rules are based on d-separations in associated DAGs, and dictate, for example, when an interventional density is equivalent to an observational one. Thus, we often rely on these rules for causal identification.

But \citeauthor{pearl2009causality}'s do calculus was designed for densities consistent with a known DAG, and researchers rarely have knowledge of every edge orientation in a causal graph. For this reason, \cite{zhang2008causal} created a do calculus for the setting where the causal graph is known only up to a \textit{maximal} or \textit{partial ancestral graph} (MAG or PAG). Other extensions include \citeauthor{correa2020calculus}'s (\citeyear{correa2020calculus}) $\sigma$-calculus for inference under non-static interventions. Following this branch of research, we offer Theorem \ref{thm:do-calc-mpdag}: a do calculus for MPDAGs.


\subsection{The Calculus}
Just as Pearl's do calculus relies on manipulations of a known DAG (e.g., $\g[D]_{\overline{\mb{X}}}$), our do calculus relies on manipulations of a known MPDAG (e.g., $\g_{\overline{\mb{X}}}$). But our graphs differ from Pearl's in notable ways. In order to build intuition for our do calculus, we first pause to describe these graphs and their differences.

As a review, \citeauthor{pearl2009causality}'s do calculus requires that interventional densities must be consistent with a DAG $\g[D]$, and that d-separations must hold in specific manipulations of $\g[D]$. For example, $\g[D]_{\overline{\mb{X}}}$ denotes the graph obtained by taking $\g[D]$ and removing all edges into $\mb{X}$. This process has an intuitive relationship with a do intervention on $\mb{X}$ in that intervening on $\mb{X}$ would override the causal effect of its parents in the original model. Note further that these mutilated graphs are still DAGs, and their d-separations capture probabilistic relationships in densities consistent with the mutilated DAGs.

In our do calculus, we assume knowledge of an MPDAG $\g$ and consider d-separations in mutilated versions of $\g$. For example, we use $\g_{\overline{\mb{X}}}$ to denote the graph obtained by taking $\g$ and removing all edges into $\mb{X}$. But this process differs from Pearl's in several ways. First, removing edges from $\g$ into $\mb{X}$ does not have the same intuitive relationship with the intervention $do(\mb{x})$, since $\g_{\overline{\mb{X}}}$ may still contain edges adjacent to $\mb{X}$ that are undirected. In the true causal DAG, the corresponding edges may be directed into $\mb{X}$ and thus, represent a causal effect on $\mb{X}$. Second, these mutilated graphs may no longer be MPDAGs. For example, consider an MPDAG $\g$ that contains the paths $V_1 \to V_2 - X$ and $V_1 \to X$. The graph $\g_{\overline{X}}$ only contains the path $V_1 \to V_2 - X$, and thus, is not closed under \citeauthor{meek1995causal}'s (\citeyear{meek1995causal}) orientation rules. In Appendix \ref{app:do}, we note a third, more technical difference useful in the proofs of Theorem \ref{thm:do-calc-mpdag}. With this discussion, we turn to our result.

\begin{theorem}[Do Calculus for MPDAGs]
\label{thm:do-calc-mpdag} 
    Let $\mb{X}$, $\mb{Y}$, $\mb{Z}$, and $\mb{W}$ be pairwise disjoint node sets in a causal MPDAG $\g=(\mb{V,E})$. Let $\g_{\overline{\mb{X}}\underline{\mb{Z}}}$ denote the graph obtained by deleting all edges into $\mb{X}$ and all edges out of $\mb{Z}$ from $\g$. We write $\g_{\overline{\mb{X}}\underline{\mb{Z}}}$ as $\g_{\underline{\mb{Z}}}$ when $\mb{X}$ is empty and as $\g_{\overline{\mb{X}}}$ when $\mb{Z}$ is empty. Then the following hold for all densities $f$ consistent with $\g$.
    \vskip .075in
    
    \textbf{Rule~1.} If $(\mb{Y} \dsepp \mb{Z} | \mb{X}, \mb{W})_{\g_{\overline{\mb{X}}}}$, then
    \begin{align}
        f(\mb{y} | do(\mb{x}), \mb{z}, \mb{w}) = f(\mb{y} | do(\mb{x}), \mb{w}). \label{eq:rule1-do-mpdag}
    \end{align}

    \textbf{Rule~2.} If $(\mb{Y} \dsepp \mb{Z} | \mb{X}, \mb{W})_{\g_{\overline{\mb{X}}\underline{\mb{Z}}}}$, then
    \begin{align} \label{eq:rule2-do-mpdag}
        f(\mb{y} | do(\mb{z}), \mb{w}, do(\mb{x})) = f(\mb{y} | \mb{z}, \mb{w}, do(\mb{x})).
    \end{align}

    \textbf{Rule~3.} If $(\mb{Y} \dsepp \mb{Z} | \mb{X}, \mb{W})_{\g_{\overline{\mb{X}}, \overline{\mb{Z'(W)}}}}$, then
    \begin{align}
    \label{eq:rule3-do-mpdag}
        \begin{split}
            f(\mb{y} | do(\mb{z}), \mb{w}, do(\mb{x})) = f(\mb{y} | \mb{w}, do(\mb{x})),
        \end{split}
    \end{align}
    \hskip .86in where $\mb{Z'(W)} := \mb{Z} \setminus \PossAn(\mb{W}, \g_{\mb{V} \setminus \mb{X}})$.
\end{theorem}

\begin{proofsketch}
    Theorem \ref{thm:do-calc-mpdag} follows from the do calculus of \cite{pearl2009causality} as well as Lemmas \ref{lem:zhang-modify-1b}, \ref{lem:zhang-modify-2b}, and \ref{lem:zhang-modify-3c}. These lemmas and their proofs can be found in Appendix \ref{app:do}, but we provide an overview here for intuition. Consider Rule 1 as an example. The result follows directly from Lemma \ref{lem:zhang-modify-1b}, which claims that if the given d-separation---$\mb{Y} \dsepp \mb{Z} \,|\, \mb{X}, \mb{W}$---holds in the mutilated graph $\g_{\overline{\mb{X}}}$, then the same d-separation will hold in $\g[D]_{\overline{\mb{X}}}$ for any DAG $\g[D]$ in the equivalence class $[\g]$. We prove this by contraposition. Thus, we start by assuming there is a DAG $\g[D] \in [\g]$ such that $\g[D]_{\overline{\mb{X}}}$ has a path $\overline{p}$ from $\mb{Y}$ to $\mb{Z}$ that is open given $\mb{X} \cup \mb{W}$. By definition, no non-collider on $\overline{p}$ is in $\mb{X} \cup \mb{W}$. We then show that every collider on $\overline{p}$ is in $\An( \mb{X} \cup \mb{W}, \g_{\overline{\mb{X}}} )$. To do this, we draw repeatedly on the relationships between $\g[D]_{\overline{\mb{X}}}$, $\g[D]$, $\g$, and $\g_{\overline{\mb{X}}}$. We prove this implies that the sequence of nodes in $\g_{\overline{\mb{X}}}$ corresponding to $\overline{p}$ forms a path from $\mb{Y}$ to $\mb{Z}$ that is open given $\mb{X} \cup \mb{W}$, which completes the proof. We rely on a similar contraposition for showing that Rules 2 and 3 hold.
\end{proofsketch}

Before providing examples, we pause to compare Theorem \ref{thm:do-calc-mpdag} with the do calculus of \cite{pearl2009causality}. Note that since all DAGs are MPDAGs, Theorem \ref{thm:do-calc-mpdag} holds when a full DAG is known. We want to highlight that our do calculus and \citeauthor{pearl2009causality}'s are identical in this case. This claim follows without proof after considering a small detail found in Rule 3. Compare the definition of $\mb{Z'(W)}$ from Theorem \ref{thm:do-calc-mpdag} with that of $\mb{Z(W)}$ from Theorem \ref{thm:do-calc}. To show these sets are equivalent when a DAG $\g$ is known, we must show that $\mb{Z} \setminus \An(\mb{W}, \g_{\mb{V} \setminus \mb{X}}) = \mb{Z} \setminus \An(\mb{W}, \g_{\overline{\mb{X}}})$. To see this, note that $\An(\mb{W}, \g_{\mb{V} \setminus \mb{X}}) \subseteq \An(\mb{W}, \g_{\overline{\mb{X}}})$, where the only variables in $\An(\mb{W}, \g_{\overline{\mb{X}}})$ not in $\An(\mb{W}, \g_{\mb{V} \setminus \mb{X}})$ are in $\mb{X}$. But $\mb{Z}$ and $\mb{X}$ are disjoint. Therefore, removing $\An(\mb{W}, \g_{\mb{V} \setminus \mb{X}})$ from $\mb{Z}$ is equivalent to removing $\An(\mb{W}, \g_{\overline{\mb{X}}})$ from $\mb{Z}$, and the claim holds.


\subsection{Examples}
\label{sec:do-examples}

The rules of Theorem \ref{thm:do-calc-mpdag} offer a broad tool for transforming interventional densities in the MPDAG setting. An appealing use of this tool is identifying a conditional causal effect. We demonstrate how to use Theorem \ref{thm:do-calc-mpdag} for this purpose in the examples below. Example \ref{ex:do-multi} does this with the help of well-known probability rules, and it does this for an effect that our identification formula was able to find. We provide Examples \ref{ex:do-rule2} and \ref{ex:do-rule3} to highlight that our do calculus is able to find conditional effects that our identification formula cannot.

\begin{figure}
    \centering
    \begin{tikzpicture}[>=stealth',shorten >=1pt,auto,node distance=2cm,main node/.style={minimum size=0.8cm,font=\sffamily\Large\bfseries},scale=0.75,transform shape]
        \node[main node]   (X)  at (0,0)    {$X$};
        \node[main node]   (Y)  at (3,0)    {$Y$};
        \node[main node]   (V1) at (1.5,1)  {$V_1$};
        \node[main node]   (V2) at (0,-2)   {$V_2$};
        \node[main node]   (V3) at (3,-2)   {$V_3$};
        \draw[->]   (X)  edge (Y);
        \draw[-]    (V1) edge (V2);
        \draw[-]    (V2) edge (X);
        \draw[->]   (V1) edge (X);
        \draw[->]   (V1) edge (Y);
        \draw[->]   (V2) edge (V3);
        \draw[->]   (Y) edge (V3);
        \end{tikzpicture}
    \caption{MPDAG used in Example \ref{ex:do-rule2}}
    \label{fig:ex-rules1+2}
\end{figure}
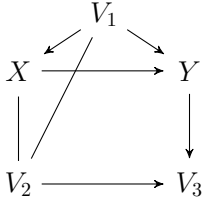

\begin{example}[Multiple Routes]
\label{ex:do-multi}
    Reconsider Example \ref{ex:id-cas}, where we found the form of a conditional effect using our identification formula. We could instead use Theorem \ref{thm:do-calc-mpdag}:
    \begin{align*}
        f(\mb{y}|do(\mb{x}),\mb{z})
            &=\scaleobj{.6}{\int} \,f(y, v_2|do(x), v_1) \diff v_2 \\
            &=\scaleobj{.6}{\int} \,f(y|do(x), v_1, v_2) \,f(v_2|do(x), v_1) \diff v_2 \\
            &=\scaleobj{.6}{\int} \,f(y|x, v_1, v_2) \,f(v_2|v_1) \diff v_2,
    \end{align*}
    The first two equalities hold by the law of total probability and the chain rule, respectively. The third holds by Rules 2 and 3, since $Y \dsepp X \,|\, V_1,V_2$ in $\g_{\underline{\mb{X}}}$ and $V_2 \dsepp X \,|\, V_1$ in $\g_{\overline{\mb{X}}}$. This example provides a case where all three methods---conditional adjustment, the identification formula, and the do calculus for MPDAGs---find the same form for the conditional effect.
\end{example}

\begin{example}[Using Rule 2]
\label{ex:do-rule2}
    Let $\g$ be the causal MPDAG in Figure \ref{fig:ex-rules1+2}, and suppose we want to identify the causal effect of $X$ on $Y$ given $\{V_1,V_2\}$. Note that we cannot use Theorem \ref{thm:id-formula}, since $\{V_1,V_2\} \cap \PossDe(X, \g) \neq \emptyset$. However, we can use Theorem \ref{thm:do-calc-mpdag} to show
    \begin{align*}
        f(y \,|\, do(x), v_1, v_2)
            &= f(y \,|\, x, v_1, v_2).
    \end{align*}
    The equality follows from Rule 2, since $Y \dsepp X \,|\, V_1, V_2$ in $\g_{\underline{\mb{X}}}$.
\end{example}

\begin{example}[Using Rule 3]
\label{ex:do-rule3}
    Let $\g$ be a causal MPDAG containing only $X - Z \to Y$, and suppose we want to identify the causal effect of $X$ on $Y$ given $Z$. Note that we cannot use Theorems \ref{thm:id-formula} or \ref{thm:id-rule3}, since $Z \in \PossDe(X, \g)$. However, we can use Theorem \ref{thm:do-calc-mpdag} to show
    \begin{align*}
        f(y \,|\, do(x), z)
            &= f(y \,|\, z).
    \end{align*}    
    This holds by Rule 3, since $Y \dsepp X \,|\, Z$ in $\g_{\overline{\mb{X}}}$.
\end{example}


\section{Identification Algorithm}
\label{sec:alg}

In Section \ref{sec:id}, we introduced our conditional identification formula for MPDAGs (Theorem \ref{thm:id-formula}), which is able to identify causal effects when the conditioning set is unaffected by treatment. In Section \ref{sec:do}, we introduced our do calculus for MPDAGs (Theorem \ref{thm:do-calc-mpdag})---a broad tool that is able to identify further effects.

In this section, we introduce our conditional identification algorithm for MPDAGs (Algorithm \ref{alg:cidm}), which combines Theorems \ref{thm:id-formula} and \ref{thm:do-calc-mpdag}. This algorithm does not restrict the conditioning set, and therefore, can identify further effects than our identification formula alone. In fact, we show below (Section \ref{sec:alg-condition}) that Algorithm \ref{alg:cidm} is complete, and thus, can find any conditional effect that is identifiable given knowledge of an MPDAG.

We start this section by describing the algorithm and providing examples of how to use the algorithm for identifying conditional effects, when possible. Then we highlight the algorithm's completeness, which we demonstrate through examples. We close this section by offering an extension to our algorithm that outputs an enumeration of possible effects in cases where the true effect is not identifiable.


\subsection{The Algorithm}

\begin{algorithm}[t]
    \vspace{.3cm}
    \Input{Disjoint node sets $\mb{X}, \mb{Y}, \mb{Z} \subseteq \mb{V}$ and causal MPDAG $\g=(\mb{V,E})$}
    \Output{Expression for $f(\mb{y} \,|\, do(\mb{x}), \mb{z})$, for any $f$ consistent with $\g$, or FAIL}
    \vspace{.3cm}
    \SetAlgoVlined
    Let $\mb{Z'} =  \mb{Z}$ and $\mb{X'} = \mb{X}$;\\ \vskip .2cm
    \While{
    $\exists$ a proper possibly causal path $\mb{X'}$ to $\mb{Y} \cup \mb{Z'}$ in $\g$ that starts undirected \label{alg:cidm-a}}{
        Pick $X \in \mb{X'}$ on such a path;\\
        \uIf{
        $(\, \mb{Y} \dsepp X \,\,|\,\, \mb{X'} \setminus \{X\}, \mb{Z'} \,)_{\, \g_{\overline{\mb{X'} \setminus \{X\}} \underline{X}}}$}{
            Add $X$ to $\mb{Z'}$;%
                \hspace*{20em}%
                \rlap{\smash{$\left.\begin{array}{@{}c@{}}\\{}\\{}\\{}\end{array}\right\}%
                \begin{tabular}{c}Applies Rule 2 of\\Theorem \ref{thm:do-calc-mpdag}\end{tabular}$}}\\
            Remove $X$ from $\mb{X'}$;
        }\Else{
        \Return FAIL\; \label{alg:cidm-b}
        }
    }\vskip .2cm
    \uIf{$(\, \mb{Y} \dsepp \mb{X'} \,\,|\,\, \mb{Z'} \,)_{\, \g_{\overline{\mb{X'}(\mb{Z'})}}}$ where $\mb{X'}(\mb{Z'}) := \mb{X'} \setminus \PossAn(\mb{Z'},\g)$ \label{alg:cidm-c}}{
        \Return $f(\mb{y} \,|\, \mb{z'})$;\label{alg:cidm-d}%
            \hspace*{19.85em}%
            \rlap{\smash{$\left.\begin{array}{@{}c@{}}\\{}\\{}\end{array}\right\}%
            \begin{tabular}{c}Applies Rule 3 of\\Theorem \ref{thm:do-calc-mpdag}\end{tabular}$}}
        \vskip .15cm
    }\Else{
        Let $\mb{Z^D} = \mb{Z'} \cap \PossDe(\mb{X'}, \g)$;\label{alg:cidm-e}\\
        Let $\mb{Z^N} = \mb{Z'} \setminus \PossDe(\mb{X'}, \g)$;\label{alg:cidm-f}\\
        Let $A =$ identification formula for $f(\mb{y},\mb{z^{\scriptscriptstyle{D}}} \,|\, do(\mb{x'}), \mb{z^{\scriptscriptstyle{N}}})$;\label{alg:cidm-g}%
            \hspace*{3.6em}%
            \rlap{\smash{$\left.\begin{array}{@{}c@{}}\\{}\\{}\\{}\\{}\\{}\end{array}\right\}%
            \begin{tabular}{c}Applies \\Theorem \ref{thm:id-formula}\end{tabular}$}}\\
        Let $B =$ identification formula for $f(\mb{z^{\scriptscriptstyle{D}}} \,|\, do(\mb{x'}), \mb{z^{\scriptscriptstyle{N}}})$;\label{alg:cidm-h}\\
        \Return $\frac{A}{B}$;\label{alg:cidm-i}
    }
\caption{Conditional Identification for MPDAGs (\texttt{CIDM})}
\label{alg:cidm}
\end{algorithm}

Our identification algorithm can be found in Algorithm \ref{alg:cidm}. It begins with pairwise disjoint node sets $\mb{X}$, $\mb{Y}$, and $\mb{Z}$ from a causal MPDAG $\g$. When possible, it outputs an identifying form for $f(\mb{y} \,|\, do(\mb{x}), \mb{z})$, where $f$ is any density consistent with $\g$. It does this in two broad steps: (1) manipulating $f(\mb{y} \,|\, do(\mb{x}), \mb{z})$ in preparation for identification, and (2) identifying $f(\mb{y} \,|\, do(\mb{x}), \mb{z})$. We describe these steps in more detail below.

Step (1) happens inside the \texttt{while} loop of lines \ref{alg:cidm-a}-\ref{alg:cidm-b}. Here the algorithm successively transfers nodes from $do(\mb{X})$ into $\mb{Z}$ using Rule 2 of the do calculus for MPDAGs (Theorem \ref{thm:do-calc-mpdag}). The \texttt{while} loop ends once there is no longer a proper possibly causal path from $\mb{X'}$ to $\mb{Y} \cup \mb{Z'}$ that starts undirected. In Section \ref{sec:alg-condition}, we show the absence of such a path is necessary for identification. Thus, if there is a node $X \in \mb{X'}$ with a possibly causal path to $\mb{Y} \cup \mb{Z'}$ that starts undirected and which Rule 2 of the do calculus cannot transfer from $do(\mb{X})$ into $\mb{Z}$, then the algorithm outputs a \texttt{FAIL}.

Step (2) happens inside the \texttt{if-else} statement of lines \ref{alg:cidm-c}-\ref{alg:cidm-i}. It begins by checking if the d-separation in Rule 3 of our do calculus holds. If so, then it provides the identifying expression $f(\mb{y} \,|\, do(\mb{x}), \mb{z}) = f(\mb{y} \,|\, \mb{z'})$. This step is not required for implementation, but we offer it as an off-ramp for a likely easy-to-estimate form of the density. If Rule 3 does not apply, then the algorithm relies on our identification formula (Theorem \ref{thm:id-formula}) in the following way. Lines \ref{alg:cidm-e}-\ref{alg:cidm-f} separate $\mb{Z'}$ into variables that \textit{are} and \textit{are not} possible descendants of $\mb{X'}$---namely, $\mb{Z^D}$ and $\mb{Z^N}$. Then lines \ref{alg:cidm-g}-\ref{alg:cidm-i} apply Theorem \ref{thm:id-formula} based on the following insight:
\begin{align}
    f(\mb{y} \,|\, do(\mb{x'}), \mb{z'}) &= \frac{f(\mb{y}, \mb{z^{\scriptscriptstyle{D}}} \,|\, do(\mb{x'}), \mb{z^{\scriptscriptstyle{N}}} )} {f(\mb{z^{\scriptscriptstyle{D}}} \,|\, do(\mb{x'}), \mb{z^{\scriptscriptstyle{N}}} )}. \label{eq:alg-frac}
\end{align}
This holds by the chain rule, where the denominator is non-zero since we assume positivity (see Section \ref{sec:prelims}). Thus, the algorithm applies Theorem \ref{thm:id-formula} to both the numerator and denominator of \eqref{eq:alg-frac}. To see that this is allowed, note that $\mb{Z^N} \cap \PossDe(\mb{X'}, \g) = \emptyset$, and by Step (1), there is no proper possibly causal path from $\mb{X'}$ to $\mb{Y} \cup \mb{Z^D}$ in $\g$ that starts undirected.


\subsection{Examples}
\label{sec:alg-examples}

In the examples below, we use Algorithm \ref{alg:cidm} in an attempt to identify the conditional causal effect of $\mb{X}$ on $\mb{Y}$ given $\mb{Z}$. In Examples \ref{ex:alg-pass1} and \ref{ex:alg-pass2}, Algorithm \ref{alg:cidm} succeeds in finding $f(\mb{y} \,|\, do (\mb{x}), \mb{z})$. But the algorithm outputs a \texttt{FAIL} in Example \ref{ex:alg-fail1}. Note that all three examples assume knowledge of an MPDAG $\g$ where $\mb{Z} \cap \PossDe(\mb{X}, \g) \neq \emptyset$ and thus, our identification formula (Theorem \ref{thm:id-formula}) cannot be applied directly.

\begin{figure}
    \centering
    \begin{subfigure}{.24\textwidth}
        \centering
        \begin{tikzpicture}[>=stealth',shorten >=1pt,auto,node distance=2cm,main node/.style={minimum size=0.8cm,font=\sffamily\Large\bfseries},scale=.7,transform shape]
        \node[main node]   (X1)  at (0,0)       {$X_1$};
        \node[main node]   (Z)   at (1.8,0)       {$Z$};
        \node[main node]   (X2)  at (1.8,1.5)     {$X_2$};
        \node[main node]   (Y)   at (3.6,0)       {$Y$};
        \draw[-]   (X1) edge (Z);
        \draw[->]  (Z)  edge (Y);
        \draw[->]  (X1) edge (X2);
        \draw[->]  (Z)  edge (X2);
        \draw[->]  (X2) edge (Y);
        \draw[->, out = 320, in=220]   (X1) edge (Y);
        \end{tikzpicture}
        \caption{}
        \label{fig:ex-alg-a}
    \end{subfigure}
    \begin{subfigure}{.24\textwidth}
        \centering
        \begin{tikzpicture}[>=stealth',shorten >=1pt,auto,node distance=2cm,main node/.style={minimum size=0.8cm,font=\sffamily\Large\bfseries},scale=0.7,transform shape]
        \node[main node]   (X)   at (0,0)       {$X$};
        \node[main node]   (Z)   at (1.8,0)       {$Z$};
        \node[main node]   (V1)  at (1.8,1.5)     {$V_1$};
        \node[main node]   (Y)   at (3.6,0)       {$Y$};
        \draw[->]   (X)  edge (Z);
        \draw[-]    (Z) edge (Y);
        \draw[->]   (V1) edge (X);
        \draw[->]   (V1) edge (Z);
        \draw[->]   (V1) edge (Y);
        \draw[->, out = 320, in=220]   (X) edge (Y);
        \end{tikzpicture}
        \caption{}
        \label{fig:ex-alg-b}
    \end{subfigure}    
    \begin{subfigure}{.24\textwidth}
        \centering
        \begin{tikzpicture}[>=stealth',shorten >=1pt,auto,node distance=2cm,main node/.style={minimum size=0.8cm,font=\sffamily\Large\bfseries},scale=.7,transform shape]
        \node[main node]   (X)   at (0,0)    {$X$};
        \node[main node]   (V2)  at (1.8,0)    {$Z$};
        \node[main node]   (V1)  at (1.8,1.5)  {$V_1$};
        \node[main node]   (Y)   at (3.6,0)    {$Y$};
        \draw[-]    (X)  edge (V2);
        \draw[->]   (V2) edge (Y);
        \draw[->]   (V1) edge (X);
        \draw[->]   (V1) edge (V2);
        \draw[->]   (V1) edge (Y);
        \draw[->, out = 320, in=220]   (X) edge (Y);
        \end{tikzpicture}
        \caption{}
        \label{fig:ex-alg-c}
    \end{subfigure}
    \caption{MPDAGs used in Examples \ref{ex:alg-pass1}-\ref{ex:alg-fail1}}
    \label{fig:ex-alg}
\end{figure}
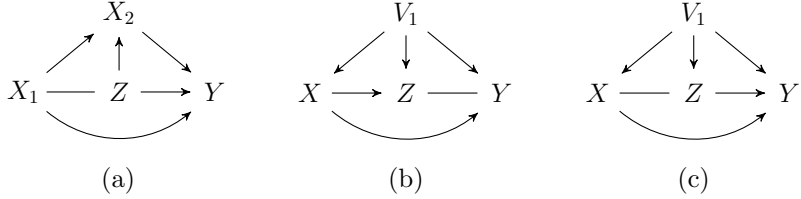

\begin{example}[Non-fractional Form]
\label{ex:alg-pass1}
    Let $\g$ be the causal MPDAG in Figure \ref{fig:ex-alg}\subref{fig:ex-alg-a}, and let $\mb{X} = \{X_1,X_2\}$, $\mb{Y} = \{Y\}$, and $\mb{Z} = \{Z\}$. To use Algorithm \ref{alg:cidm}, we first consider the \emph{\texttt{while}} loop of lines \ref{alg:cidm-a}-\ref{alg:cidm-b}. Since there is a possibly causal path from $X_1 \in \mb{X}$ to $\mb{Y} \cup \mb{Z}$ that starts undirected and does not contain $X_2$, we confirm that $\mb{Y} \dsepp X_1 \,|\, X_2, \mb{Z}$ in $\g_{\overline{X_2}\underline{X_1}}$. Thus, we define $\mb{X'} = \{X_2\}$ and $\mb{Z'} = \{X_1,Z\}$. Now that all possibly causal paths from $\mb{X'}$ to $\mb{Y} \cup \mb{Z'}$ start directed, we exit the \emph{\texttt{while}} loop. Then we skip lines \ref{alg:cidm-c}-\ref{alg:cidm-d} since $\mb{Y} \not\dsepp \mb{X'} \,|\, \mb{Z'}$ in $\g_{\overline{\mb{X'}}}$. Finally, by lines \ref{alg:cidm-e}-\ref{alg:cidm-i},
    \begin{align*}
        f(\mb{y} \,|\, do(\mb{x}), \mb{z}) 
            &= f(y \,|\, do(x_2), x_1, z)\\
            &= f(y \,|\, x_1,x_2,z).
    \end{align*}
\end{example}

\begin{example}[Fractional Form]
\label{ex:alg-pass2}
    Let $\g$ be the causal MPDAG in Figure \ref{fig:ex-alg}\subref{fig:ex-alg-b}, and let $\mb{X} = \{X\}$, $\mb{Y} = \{Y\}$, and $\mb{Z} = \{Z\}$. To use Algorithm \ref{alg:cidm}, we first bypass the \emph{\texttt{while}} loop of lines \ref{alg:cidm-a}-\ref{alg:cidm-b}, since all possibly causal paths from $\mb{X}$ to $\mb{Y} \cup \mb{Z}$ start directed. We also skip lines \ref{alg:cidm-c}-\ref{alg:cidm-d} since $\mb{Y} \not\dsepp \mb{X} \,|\, \mb{Z}$ in $\g$. Thus, by lines \ref{alg:cidm-e}-\ref{alg:cidm-i},
    \begin{align*}
        f(\mb{y} \,|\, do(\mb{x}), \mb{z}) 
            = \frac{ \int f(y,z|v_1,x) f(v_1) \diff v_1 }
              {\int f(z \,|\, v_1 ,x) f(v_1) \diff v_1}.
    \end{align*}
\end{example}

\begin{example}[Not Identifiable]
\label{ex:alg-fail1}
    Let $\g$ be the causal MPDAG in Figure \ref{fig:ex-alg}\subref{fig:ex-alg-c}, and let $\mb{X} = \{X\}$, $\mb{Y} = \{Y\}$, and $\mb{Z} = \{Z\}$. To use Algorithm \ref{alg:cidm}, we first consider the \emph{\texttt{while}} loop of lines \ref{alg:cidm-a}-\ref{alg:cidm-b}. Note that there is a possibly causal path from $X \in \mb{X}$ to $\mb{Y} \cup \mb{Z}$ that starts undirected, but $\mb{Y} \not\dsepp X \,|\, \mb{Z}$ in $\g_{\underline{X}}$. Thus, the algorithm returns a \emph{\texttt{FAIL}}. In Section \ref{sec:alg-condition}, we show this effect is not identifiable in $\g$ (see Example \ref{ex:alg-condition1}).
\end{example}


\subsection{Further Examples}

The three examples above rely on our identification algorithm by walking through steps that our other methods would not. But we want to highlight that since Algorithm \ref{alg:cidm} combines our identification formula with our do calculus for MPDAGs, applying our algorithm sometimes amounts to no more than applying one of these methods alone. We demonstrate this in the examples below, which revisit examples from Sections \ref{sec:id} and \ref{sec:do}.

\begin{example}[Reduces to Identification Formula]
    Reconsider Example \ref{ex:id-cas}, but now use Algorithm \ref{alg:cidm} to identify the causal effect. We skip lines \ref{alg:cidm-a}-\ref{alg:cidm-d}, since the following hold: $\mb{Z} \cap \PossDe(\mb{X}, \g) = \emptyset$, no proper possibly causal path from $\mb{X}$ to $\mb{Y}$ starts undirected, and $\mb{Y} \not\dsepp \mb{X} \,|\, \mb{Z}$ in $\g_{\overline{\mb{X}}}$. Thus, the algorithm reduces to lines \ref{alg:cidm-e}-\ref{alg:cidm-i}, and since $\mb{Z^N}=\mb{Z}$, these lines reduce to the same application of our identification formula shown in Example \ref{ex:id-cas}. Note that this logic also applies to Example \ref{ex:id-nocas}.
\end{example}

\begin{example}[Reduces to Rule 3]
    Reconsider Example \ref{ex:id-rule3}, but now use Algorithm \ref{alg:cidm} to identify the causal effect. We skip lines \ref{alg:cidm-a}-\ref{alg:cidm-b}, since no there is no possibly causal path from $\mb{X}$ to $\mb{Y} \cup \mb{Z}$. From lines \ref{alg:cidm-c}-\ref{alg:cidm-d}, we confirm that $\mb{Y} \dsepp \mb{X} \,|\, \mb{Z}$ in $\g_{\overline{\mb{X}}}$. Thus, the algorithm reduces to an application of Rule 3 of our do calculus. In this case, note that this is equivalent to the application of Theorem \ref{thm:id-rule3} shown in Example \ref{ex:id-rule3}.
\end{example}

\begin{example}[Reduces to Rule 2]
    Reconsider Example \ref{ex:do-rule2}. To use Algorithm \ref{alg:cidm}, we first consider the \emph{\texttt{while}} loop of lines \ref{alg:cidm-a}-\ref{alg:cidm-b}. Since there is a possibly causal path from $X \in \mb{X}$ to $\mb{Y} \cup \mb{Z}$ that starts undirected, we confirm that $\mb{Y} \dsepp X \,|\, \mb{Z}$ in $\g_{\underline{X}}$. Thus, we define $\mb{X'} = \emptyset$ and $\mb{Z'} = \{X,Z\}$. Formally, the algorithm must exit the \emph{\texttt{while}} loop and consider lines \ref{alg:cidm-c}-\ref{alg:cidm-d}, but note that it has already found an identifying form (that the additional steps will not change). Thus, the algorithm reduces to the same application of Rule 2 of our do calculus shown in Example \ref{ex:do-rule2}. Additionally note that the same logic applies to Example \ref{ex:do-rule3}.
\end{example}


\subsection{Soundness and Completeness}
\label{sec:alg-condition}

Algorithm \ref{alg:cidm} is sound for identifying conditional causal effects given an MPDAG. That is, any expression the algorithm outputs will be an identifying form of $f(\mb{y} \,|\, do(\mb{x}), \mb{z})$. This follows directly from Rules 2-3 of the do calculus for MPDAGs (Theorem \ref{thm:do-calc-mpdag}), the chain rule, our positivity assumption (see Section \ref{sec:prelims}), and our conditional identification formula (Theorem \ref{thm:id-formula}).

Algorithm \ref{alg:cidm} is also complete for identifying conditional causal effects given an MPDAG. To prove this, we offer the theorem below. This result shows that when our algorithm outputs a \texttt{FAIL}, the effect is not identifiable. That is, our algorithm is able to find an identifying form for any identifiable effect.

\begin{theorem}[Completeness of Algorithm \ref{alg:cidm}]
\label{thm:alg-complete}    
    Let $\mb{X}$, $\mb{Y}$, and $\mb{Z}$ be pairwise disjoint node sets in a causal MPDAG $\g$. If there is a proper possibly causal path from $\mb{X}$ to $\mb{Y} \cup \mb{Z}$ in $\g$ that starts with an undirected edge and contains any $X \in \mb{X}$ such that $\mb{Y} \not\dsepp X \,|\, \mb{X} \setminus \{X\}, \mb{Z}$ in $\g_{\overline{\mb{X} \setminus \{X\}} \underline{X}}$, then the conditional causal effect of $\mb{X}$ on $\mb{Y}$ given $\mb{Z}$ is not identifiable in $\g$.
\end{theorem}

\begin{proofsketch}
    The proof of Theorem \ref{thm:alg-complete} can be found in Appendix \ref{app:alg}, but we provide an overview here for intuition. We begin the proof by assuming that a possibly causal path---as given in the statement of Theorem \ref{thm:alg-complete}---exists in the MPDAG $\g$. Based on the given d-separation between $X \in \mb{X}$ and $\mb{Y}$, we consider the following non-empty sets:
    \begin{align}
        \mb{S_1} &= \left\{\begin{array}{ll}
                        \text{definite status paths in $\g_{\overline{\mb{X} \setminus \{X\}} \underline{X}}$ from $X$ to $\mb{Y}$ that are} \\
                        \text{d-connecting given $\mb{X} \setminus \{X\} \cup \mb{Z}$}
                    \end{array}\right\}\text{ and} \nonumber\\
        \mb{S_2} &= \{ \text{ paths in $\mb{S_1}$ with the fewest colliders } \}. \label{thm:alg-complete-s2}
    \end{align}
    The remainder of the proof falls into two cases: (a) when $\mb{S_2}$ only contains paths that start directed (i.e., of the form $X \gets \dots Y$) and (b) when $\mb{S_2}$ contains paths that start undirected (i.e., of the form $X - \dots Y$). Rather than providing an overview of each case, we offer two examples below. Example \ref{ex:alg-condition1} considers an MPDAG with a path that falls into (a), and Example \ref{ex:alg-condition2} considers an MPDAG with a path that falls into (b). In both examples, we show that the conditional causal effect is not identifiable.
\end{proofsketch}


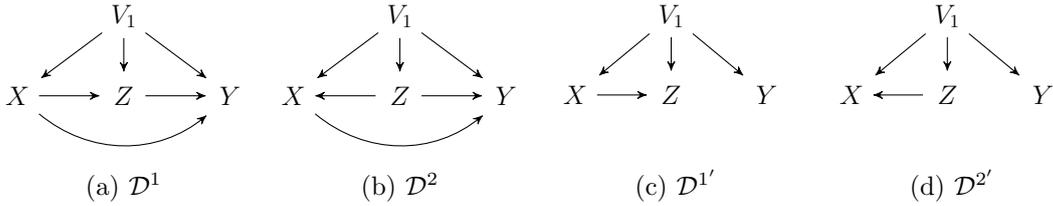
\begin{figure}
    \centering
    \begin{subfigure}{.23\textwidth}
        \centering
        \begin{tikzpicture}[>=stealth',shorten >=1pt,auto,node distance=2cm,main node/.style={minimum size=0.8cm,font=\sffamily\Large\bfseries},scale=.7,transform shape]
        \node[main node]   (X)   at (0,0)    {$X$};
        \node[main node]   (Z)  at (2,0)    {$Z$};
        \node[main node]   (V1)  at (2,1.5)  {$V_1$};
        \node[main node]   (Y)   at (4,0)    {$Y$};
        \draw[->]    (X)  edge (Z);
        \draw[->]   (Z) edge (Y);
        \draw[->]   (V1) edge (X);
        \draw[->]   (V1) edge (Z);
        \draw[->]   (V1) edge (Y);
        \draw[->, out = 320, in=220]   (X) edge (Y);
        \end{tikzpicture}
        \caption{$\g[D]^1$}
        \label{fig:alg-condition1-a}
    \end{subfigure}
    \begin{subfigure}{.23\textwidth}
        \centering
        \begin{tikzpicture}[>=stealth',shorten >=1pt,auto,node distance=2cm,main node/.style={minimum size=0.8cm,font=\sffamily\Large\bfseries},scale=.7,transform shape]
        \node[main node]   (X)   at (0,0)    {$X$};
        \node[main node]   (Z)  at (2,0)    {$Z$};
        \node[main node]   (V1)  at (2,1.5)  {$V_1$};
        \node[main node]   (Y)   at (4,0)    {$Y$};
        \draw[<-]    (X)  edge (Z);
        \draw[->]   (Z) edge (Y);
        \draw[->]   (V1) edge (X);
        \draw[->]   (V1) edge (Z);
        \draw[->]   (V1) edge (Y);
        \draw[->, out = 320, in=220]   (X) edge (Y);
        \end{tikzpicture}
        \caption{$\g[D]^2$}
        \label{fig:alg-condition1-b}
    \end{subfigure}
    \begin{subfigure}{.23\textwidth}
        \centering
        \begin{tikzpicture}[>=stealth',shorten >=1pt,auto,node distance=2cm,main node/.style={minimum size=0.8cm,font=\sffamily\Large\bfseries},scale=.7,transform shape]
        \node[main node]   (X)   at (0,0)    {$X$};
        \node[main node]   (Z)  at (1.8,0)    {$Z$};
        \node[main node]   (V1)  at (1.8,1.5)  {$V_1$};
        \node[main node]   (Y)   at (3.6,0)    {$Y$};
        \draw[->]   (X)  edge (Z);
        \draw[->]   (V1) edge (Z);
        \draw[->]   (V1) edge (Y);
        \draw[->] (V1) edge (X);
        \end{tikzpicture}
        \vspace{.5cm}
        \caption{$\g[D]^{1'}$}
        \label{fig:alg-condition1-c}
    \end{subfigure}
    \begin{subfigure}{.23\textwidth}
        \centering
        \begin{tikzpicture}[>=stealth',shorten >=1pt,auto,node distance=2cm,main node/.style={minimum size=0.8cm,font=\sffamily\Large\bfseries},scale=.7,transform shape]
        \node[main node]   (X)   at (0,0)    {$X$};
        \node[main node]   (Z)  at (1.8,0)    {$Z$};
        \node[main node]   (V1)  at (1.8,1.5)  {$V_1$};
        \node[main node]   (Y)   at (3.6,0)    {$Y$};
        \draw[<-]   (X)  edge (Z);
        \draw[->]   (V1) edge (Z);
        \draw[->]   (V1) edge (Y);
        \draw[->] (V1) edge (X);
        \end{tikzpicture}
        \vspace{.5cm}
        \caption{$\g[D]^{2'}$}
        \label{fig:alg-condition1-d}
    \end{subfigure}
    \caption{DAGs used in Example \ref{ex:alg-condition1}}
    \label{fig:alg-condition1}
\end{figure}

\begin{example}[Path in $\mb{S_2}$ that Starts Directed]
\label{ex:alg-condition1}
    Reconsider Example \ref{ex:alg-fail1}, where we attempted to use Algorithm \ref{alg:cidm} to identify the conditional effect of $\mb{X}$ on $\mb{Y}$ given $\mb{Z}$ but the algorithm output a \texttt{FAIL}. Note that the MPDAG $\g=(\mb{V},\mb{E})$ in this example contains the path $X \gets V_1 \to Y$, which belongs to the set $\mb{S_2}$ given in \eqref{thm:alg-complete-s2} and which starts directed.
    
    To show the conditional effect is not identifiable in $\g$, consider the DAGs $\g[D]^1, \g[D]^2 \in [\g]$ shown in Figure \ref{fig:alg-condition1}\subref{fig:alg-condition1-a}-\subref{fig:alg-condition1-b}. It suffices to show that each $\g[D]^i$ is compatible with a family of interventional densities $\mb{F^*_i} = \{ f_i(\mb{v}|do(\mb{x'})) : \mb{X'} \subseteq \mb{V} \}$ such that the following hold.
    \begin{align}
        f_1(\mb{v}) &\,=\, f_2(\mb{v}). \label{eq:alg-condition1-a}\\
        f_1(\mb{y} \,|\, do(\mb{x}), \mb{z}) & \,\neq\, f_2(\mb{y} \,|\, do(\mb{x}), \mb{z}). \label{eq:alg-condition1-b}
    \end{align}

    To define such families, consider the DAG $\g[D]^{1'}$ in Figure \ref{fig:alg-condition1}\subref{fig:alg-condition1-c}, and note that $\g[D]^{1'}$ is constructed by removing edges from $\g[D]^1$. Below we build an $\mb{F^*_1}$ such that $\g[D]^{1'}$ is compatible with $\mb{F^*_1}$. It follows that $\g[D]^1$ is compatible with $\mb{F^*_1}$ (see Lemma \ref{lem:markov-sub}, App. \ref{app:id-condition}). Start by considering the following structural equation model (SEM) with independent errors:
    \begin{align}
        V_1       &\gets  \varepsilon_{v_1};                   &\varepsilon_{v_1} &\sim \mathcal{N}(0,1); \label{ex:alg-condition1-sem1}\\
        X   &\gets  \tfrac{1}{2}V_1   +  \varepsilon_x;  &\varepsilon_x &\sim \mathcal{N}(0,\tfrac{3}{4});\nonumber\\
        Z   &\gets  \tfrac{1}{2} V_1 + \tfrac{1}{2} X + \varepsilon_z; &\varepsilon_z &\sim \mathcal{N}(0, \tfrac{1}{4});\nonumber\\
        Y   &\gets  \tfrac{1}{2} V_1 + \varepsilon_y;    &\varepsilon_y &\sim \mathcal{N}(0,\tfrac{3}{4}). \nonumber
    \end{align}
    We build $\mb{F^*_1}$ by letting $f_1(\mb{v})$ be the density of the multivariate normal generated by the SEM in \eqref{ex:alg-condition1-sem1}. For the remaining densities in $\mb{F^*_1}$, we let $f_1(\mb{v} \,|\, do(\mb{x'})) := f_*(\mb{v})$, where $f_*$ is  the density of the multivariate normal generated by taking the SEM in \eqref{ex:alg-condition1-sem1} and replacing $\mb{X'}$ with its interventional value $\mb{x'}$ \citep{pearl2009causality}.

    Now consider the DAG $\g[D]^{2'}$ in Figure \ref{fig:alg-condition1}\subref{fig:alg-condition1-d}, and note that $\g[D]^{2'}$ is constructed by removing edges from $\g[D]^2$. Next consider the following SEM with independent errors:
    \begin{align}
        V_1       &\gets  \varepsilon_{v_1};                   &\varepsilon_{v_1} &\sim \mathcal{N}(0,1); \label{ex:alg-condition1-sem2}\\
        X   &\gets  {-\tfrac{1}{7}V_1} + \tfrac{6}{7} Z + \varepsilon_x;  &\varepsilon_x &\sim \mathcal{N}(0,\tfrac{3}{7}); \nonumber\\
        Z   &\gets  \tfrac{3}{4} V_1 + \varepsilon_z;    &\varepsilon_z &\sim \mathcal{N}(0,\tfrac{7}{16}); \nonumber\\
        Y   &\gets  \tfrac{1}{2} V_1 + \varepsilon_y;    &\varepsilon_y &\sim \mathcal{N}(0,\tfrac{3}{4}), \nonumber
    \end{align}
    Just as we built $\mb{F^*_1}$ based on the SEM in \eqref{ex:alg-condition1-sem1}, we build $\mb{F^*_2}$ based on the SEM in \eqref{ex:alg-condition1-sem2}. It follows that $\g[D]^{2'}$---and therefore $\g[D]^2$---is compatible with $\mb{F^*_2}$.

    Thus we have two families of interventional densities such that $\g[D]^i$ is compatible with $\mb{F^*_i}$. Further, we can show that $f_1(\mb{v})=f_2(\mb{v})$ so that \eqref{eq:alg-condition1-a} holds. To complete the proof, we show that \eqref{eq:alg-condition1-b} holds. It suffices to show that $E[Y \,|\, do(X=1), Z=0]$ is not the same under $f_1$ and $f_2$. We start by calculating this expectation under $f_2$:    
    \begin{align*}
        \E_{f_2}[Y \,|\, do(X=1), Z=0]
            &\,=\, \left. \E_{f_2}[Y \,|\, Z] \,\, \right\rvert_{\, Z=0}\\
            &\,=\, \left. \E_{f_2}[Y] + \frac{\Cov_{f_2}(Y,Z)}{\Var_{f_2}(Z)} \cdot \Big(Z - \E_{f_2}[Z]\Big) \, \right\rvert_{\, Z=0}\\
            &\,=\, \left. \tfrac{3}{8} Z \, \right\rvert_{\, Z=0}\\
            &\,=\, 0.
    \end{align*}
    The first equality follows from \citeauthor{pearl2009causality}'s do calculus (Theorem \ref{thm:do-calc}), since $Y \dsepp X \,|\, Z$ in $\g[D]^{2'}_{\overline{X}}$ and since $f_2$ is consistent with $\g[D]^{2'}$. The second equality follows from properties of multivariate normals (see Lemma \ref{lem:mardia}). 
    
    To calculate the same expectation under $f_1$, recall that $f_1(\mb{v} \,|\, do(x)) := f_*(\mb{v})$, where $f_*(\mb{v})$ is the multivariate normal density generated by the following SEM with independent errors:
    \begin{align}
        V_1       &\gets  \varepsilon_{v_1};                   &\varepsilon_{v_1} &\sim \mathcal{N}(0,1); \label{ex:alg-condition1-sem3}\\
        X   &\gets  x; \nonumber\\
        Z   &\gets  \tfrac{1}{2} V_1 + \tfrac{1}{2} x + \varepsilon_z; &\varepsilon_z &\sim \mathcal{N}(0, \tfrac{1}{4});\nonumber\\
        Y   &\gets  \tfrac{1}{2} V_1 + \varepsilon_y;    &\varepsilon_y &\sim \mathcal{N}(0,\tfrac{3}{4}), \nonumber
    \end{align}
    It follows that $f_1(y \,|\, do(x),z) := f_*(y \,|\, z)$ is the conditional density of $Y \,|\, Z$ under $f_*(\mb{v})$. Based on this and properties of multivariate normals (see Lemma \ref{lem:mardia}),
    \begin{align*}
        \E_{f_1}[Y \,|\, do(X= 1), Z=0]
            &\,=\, \left. \E_{f_*}[Y \,|\, Z] \,\, \right\rvert_{\, x=1, \, Z=0}\\
            &\,=\, \left. \E_{f_*}[Y] + \frac{\Cov_{f_*}(Y,Z)}{\Var_{f_*}(Z)} \cdot \Big(Z - \E_{f_*}[Z]\Big) \, \right\rvert_{\, x=1, \, Z=0}\\
            &\,=\, \left. \tfrac{1}{2} \Big(Z - \frac{1}{2}x \Big) \, \right\rvert_{\, x=1, \, Z=0}\\
            &\,\neq\, 0.
    \end{align*}
\end{example}


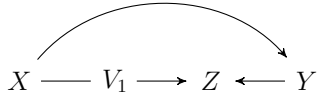
\begin{figure}
    \centering
    \begin{subfigure}{.33\textwidth}
        \centering
        \begin{tikzpicture}[>=stealth',shorten >=1pt,auto,node distance=2cm,main node/.style={minimum size=0.8cm,font=\sffamily\Large\bfseries},scale=.7,transform shape]
        \node[main node]   (X)  at (0,0)   {$X$};
        \node[main node]   (V1) at (1.8,0)   {$V_1$};
        \node[main node]   (Z) at (3.6,0)   {$Z$};
        \node[main node]   (Y)  at (5.4,0)   {$Y$};
        \draw[-]    (X)  edge (V1);
        \draw[->]   (V1) edge (Z);
        \draw[->]   (Y)  edge (Z);
        \draw[->, out = 50, in = 130]   (X) edge (Y);
        \end{tikzpicture}
    \end{subfigure}
    \caption{MPDAG used in Example \ref{ex:alg-condition2}}
    \label{fig:alg-condition2}
\end{figure}

\begin{example}[Path in $\mb{S_2}$ that Starts Undirected]
\label{ex:alg-condition2}    
    Let $\g=(\mb{V},\mb{E})$ be the MPDAG in Figure \ref{fig:alg-condition2}. If we attempt to identify the conditional effect of $X$ on $Y$ given $Z$, Algorithm \ref{alg:cidm} outputs a \texttt{FAIL}. We want to show this effect is not identifiable in $\g$. To do this, we follow the same strategy as Example \ref{ex:alg-condition1}, and thus, exclude duplicate technical details. However, note that---unlike Example \ref{ex:alg-condition1}---$\g$ contains the path $X - V_1 \to Z \gets Y$, which belongs to the set $\mb{S_2}$ given in \eqref{thm:alg-complete-s2} and which starts undirected.

    Let $\g[D]^1$, $\g[D]^2$ be the DAGs in $[\g]$ with edges $X \to V_1$ and $X \gets V_1$, respectively. Then let $\g[D]^{1'}$, $\g[D]^{2'}$ be the DAGs constructed by removing the edge $X \to Y$ from $\g[D]^1$, $\g[D]^2$, so that 
    \begin{align*}
        \g[D]^{1'}: \,\, &X \to V_1 \to Z \gets Y, \text{ and} \\
        \g[D]^{2'}: \,\, &X \gets V_1 \to Z \gets Y.
    \end{align*}
    Define $\mb{F^*_1}$ (as in Example \ref{ex:alg-condition1}) based on the following SEM with independent errors:
    \begin{align*}
        X   &\gets  \varepsilon_x;                     &\varepsilon_x &\sim \mathcal{N}(0,1);\\
        V_1 &\gets \tfrac{1}{2}X + \varepsilon_{v_1};  &\varepsilon_{v_1} &\sim \mathcal{N}(0,\tfrac{3}{4});\\
        Z   &\gets \tfrac{1}{2} V_1 + \tfrac{1}{2} Y + \varepsilon_z; &\varepsilon_z &\sim \mathcal{N}(0, \tfrac{1}{2});\\
        Y   &\gets \varepsilon_y;                      &\varepsilon_y &\sim \mathcal{N}(0,1).
    \end{align*}
    Analogously, define $\mb{F^*_2}$ based on the following SEM with independent errors:
    \begin{align*}
        X &\gets \tfrac{1}{2}V_1 + \varepsilon_x;        &\varepsilon_x &\sim \mathcal{N}(0,\tfrac{3}{4});\\
        V_1   &\gets  \varepsilon_{v_1};               &\varepsilon_{v_1} &\sim \mathcal{N}(0,1);\\
        Z   &\gets \tfrac{1}{2} V_1 + \tfrac{1}{2} Y + \varepsilon_z; &\varepsilon_z &\sim \mathcal{N}(0, \tfrac{1}{2});\\
        Y   &\gets \varepsilon_y;                      &\varepsilon_y &\sim \mathcal{N}(0,1).
    \end{align*}
    
    Thus we have two families of interventional densities where $\g[D]^i$ is compatible with $\mb{F^*_i}$ and where we can show $f_1(\mb{v})=f_2(\mb{v})$. To complete the proof, we show that $E[Y \,|\, do(X=1), Z=0]$ is not the same under $f_1$ and $f_2$.
    \begin{align*}
        \E_{f_1}[Y \,|\, do(X=1), Z=0] 
            &\,=\,      \E_{f_1}[Y \,|\, X=1, Z=0] 
             \,=\,      \left. {-\tfrac{2}{15}X} + \tfrac{8}{15}Z \,\right\rvert_{\, X=1, Z=0} 
             \,\neq\,   0.\\
        \E_{f_2}[Y \,|\, do(X=1), Z=0]
            &\,=\,      \E_{f_2}[Y \,|\, Z=0]
             \,=\,      \left. {\tfrac{1}{2}Z} \,\right\rvert_{\, Z=0} 
             \,=\,      0.
    \end{align*}
    The first equality of each line follows from Rules 2 and 3 of \citeauthor{pearl2009causality}'s do calculus (Theorem \ref{thm:do-calc}), respectively, since $Y \dsepp X \,|\, Z$ in $\g[D]^{1'}_{\underline{X}}$; $Y \dsepp X \,|\, Z$ in $\g[D]^{2'}_{\overline{X}}$; and $f_i$ is consistent with $\g[D]^{i'}$. The remaining equalities follow from properties of multivariate normals (see Lemma \ref{lem:mardia}).
\end{example}


\subsection{An Extension for Non-identifiable Effects}
\label{sec:enumeration}

Prior research on causal effect identification includes suggestions for estimation when an effect is not identifiable. The IDA algorithm of \cite{maathuis2009estimating} does this by outputting a multiset of all possible causal effects. It relies on the fact that even when an effect is not identifiable given a \textit{completed partially directed acyclic graph} (CPDAG) $\g[C]$, the effect is identifiable in each DAG in $[\g[C]]$. This multiset does not tell us which of its elements is the true effect, but it does provide some information, such as bounds on the causal effect. Subsequent research extended these results to consider additional settings and to improve computation \citep{maathuis2010predicting, perkovic2017interpreting, nandy2017estimating, fang2020ida, guo2021minimal}.

Following this line of research, we offer Algorithm \ref{alg:cida} (\texttt{CIDME})---an extension of our identification algorithm (Algorithm \ref{alg:cidm}) that outputs a multiset of possible expressions for the interventional density $f(\mb{y} \,|\, do(\mb{x}), \mb{z})$ given an MPDAG $\g$. When the conditional effect is identifiable, this multiset will be the identical output to that of Algorithm \ref{alg:cidm}. But when the conditional effect is not identifiable, this multiset will be a set of  possible expressions for $f(\mb{y} \,|\, do(\mb{x}), \mb{z})$---one for each partition of the equivalence class $[\g]$. 

Algorithm \ref{alg:cida} produces these partitions in the following way. Whenever there is a proper possibly causal path from $\mb{X'}$ to $\mb{Y} \cup \mb{Z'}$ in $\g$ that starts with an undirected edge $X-V_2$, the algorithm takes a shortest such a path and considers both orientations of $\langle X, V_2 \rangle$. Orienting $X \to V_2$ in $\g$ and completing R1-R4 of \cite{meek1995causal} forms a new MPDAG $\g_1$ with fewer possibly causal paths from $\mb{X'}$ to $\mb{Y} \cup \mb{Z'}$ that start undirected, and orienting $X \gets V_2$ forms an analogous MPDAG $\g_2$. Algorithm \ref{alg:cida} takes these new MPDAGs and begins anew, outputting \texttt{CIDME}($\mb{X',Z',Y}, \g_1$) and \texttt{CIDME}($\mb{X',Z',Y}, \g_2$). If the causal effect of $\mb{X'}$ on $\mb{Y}$ given $\mb{Z'}$ is identifiable given $\g_1$ and $\g_2$, then lines \ref{alg:cida-c}-\ref{alg:cida-d} of Algorithm \ref{alg:cida} output a multiset with two elements: expressions for $f(\mb{y} \,|\, do(\mb{x}), \mb{z})$ given $\g_1$ and $\g_2$, respectively. Otherwise, the algorithm iterates by continuing to orient the first edge of any possibly causal path from $\mb{X'}$ to $\mb{Y} \cup \mb{Z'}$ that starts undirected, until none remain.

\begin{algorithm}[t]
    \vspace{.3cm}
    \Input{Disjoint node sets $\mb{X}, \mb{Y}, \mb{Z} \subseteq \mb{V}$ and causal MPDAG $\g=(\mb{V,E})$}
    \Output{Set of possible expressions for $f(\mb{y} \,|\, do(\mb{x}), \mb{z})$, for any $f$ consistent with $\g$}
    \vspace{.3cm}
    \SetAlgoVlined
    Let $\mb{Z'} =  \mb{Z}$ and $\mb{X'} = \mb{X}$;\\ \vskip .2cm
    \While{
    $\exists$ a proper possibly causal path $\mb{X'}$ to $\mb{Y} \cup \mb{Z'}$ in $\g$ that starts undirected}{
        Pick $X \in \mb{X'}$ on such a path;\\
        \uIf{
        $(\, \mb{Y} \dsepp X \,\,|\,\, \mb{X'} \setminus \{X\}, \mb{Z'} \,)_{\, \g_{\overline{\mb{X'} \setminus \{X\}} \underline{X}}}$}{
            Add $X$ to $\mb{Z'}$;\\
            Remove $X$ from $\mb{X'}$;
        }\Else{
        Pick a shortest proper possibly causal path in $\g$ \\
            \hspace*{.4em} from  $\mb{X'}$ to $\mb{Y} \cup \mb{Z'}$
             that starts with $X - V_2$;
            \hspace*{25.4em}%
            \rlap{\smash{$\left.\begin{array}{@{}c@{}}\\{}\\{}\\{}\\{}\\{}\end{array}\right\}%
            \begin{tabular}{c}Partitions $[\g]$ \\by orienting $X - V_2$
            \end{tabular}$}}\\
            \vspace{-.5cm}
        Form $\g_1$: orient $X \to V_2$, complete R1-R4 of \cite{meek1995causal}; \label{alg:cida-meek1}\\
        Form $\g_2$: orient $X \gets V_2$, complete R1-R4 of \cite{meek1995causal}; \label{alg:cida-meek2}\\
        \Return \texttt{CIDME}($\mb{X',Z',Y}, \g_1$); \label{alg:cida-c}\\
        \Return \texttt{CIDME}($\mb{X',Z',Y}, \g_2$); \label{alg:cida-d}
        }
    }\vskip .2cm
    \uIf{$(\, \mb{Y} \dsepp \mb{X'} \,\,|\,\, \mb{Z'} \,)_{\, \g_{\overline{\mb{X'}(\mb{Z'})}}}$ where $\mb{X'}(\mb{Z'}) := \mb{X'} \setminus \PossAn(\mb{Z'},\g)$}{
        \Return $f(\mb{y} \,|\, \mb{z'})$;
        \vskip .15cm
    }\Else{
        Let $\mb{Z^D} = \mb{Z'} \cap \PossDe(\mb{X'}, \g)$;\\
        Let $\mb{Z^N} = \mb{Z'} \setminus \PossDe(\mb{X'}, \g)$;\\
        Let $A =$ identification formula for $f(\mb{y},\mb{z^{\scriptscriptstyle{D}}} \,|\, do(\mb{x'}), \mb{z^{\scriptscriptstyle{N}}})$;\\
        Let $B =$ identification formula for $f(\mb{z^{\scriptscriptstyle{D}}} \,|\, do(\mb{x'}), \mb{z^{\scriptscriptstyle{N}}})$;\\
        \Return $\frac{A}{B}$;
    }
\caption{Conditional Identification for MPDAGs + Enumeration (\texttt{CIDME})}
\label{alg:cida}
\end{algorithm}


\section{Discussion}
\label{sec:discussion}

In this paper, we propose strategies to identify a conditional causal effect from observational data. We assume knowledge of an MPDAG, a graph that can be learned from observational data and that allows for the addition of background knowledge. Our results include 
\begin{itemize}
    \item an identification formula (Theorem \ref{thm:id-formula}), which is sound and complete for identifying a conditional effect when the conditioning set is unaffected by treatment (Proposition \ref{prop:id-condition}),
    
    \item a do calculus for MPDAGs (Theorem \ref{thm:do-calc-mpdag}), which is sound for identifying a conditional effect, and
    
    \item an identification algorithm (Algorithm \ref{alg:cidm}), which is sound and complete for identifying a conditional effect (Theorem \ref{thm:alg-complete}) and which combines our preceding results.
\end{itemize} 
Our work builds on prior research in causal identification, and we detail these connections in the sections above. Most directly, our identification formula broadens the scope of \citeauthor{perkovic2020identifying}'s (\citeyear{perkovic2020identifying}) formula for unconditional effects given an MPDAG. Similarly, our do calculus is comparable to \cite{zhang2008causal} and \citeauthor{jaber2022causal}'s (\citeyear{jaber2022causal}) calculus for PAGs. And our identification algorithm (\texttt{CIDM}) relates to a broad history of conditional identification algorithms, such as the \texttt{IDC} algorithm of \cite{shpitser2006identification} and the \texttt{CIDP} algorithm of \cite{jaber2022causal}. Our extension of \texttt{CIDM} for non-identifiable effects follows a rich thread of algorithms for enumerating possible causal effects, including the algorithms of \cite{maathuis2009estimating, maathuis2010predicting, nandy2017estimating, fang2020ida}; and \cite{guo2021minimal}.

The main limitation of our work compared to the results of \cite{shpitser2006identification}, \cite{zhang2008causal}, and \cite{jaber2022causal} is that we consider a setting that does not allow latent confounding. But we want to highlight that research on identification in these graphs (e.g., PAGs) does not cover the general MPDAG setting. For example, our \texttt{CIDM} algorithm cannot be seen as a simplification of the \texttt{CIDP} algorithm of \cite{jaber2022causal}. That is, a naive translation of the \texttt{CIDP} algorithm to the MPDAG setting would not be complete for identification given an MPDAG (see Example \ref{ex:cidp}, in App. \ref{rem-cidp}). This holds since the completeness of the \texttt{CIDM} algorithm relies on properties of PAGs (e.g., undirected components must be chordal, undirected paths must be possibly causal) that do not hold generally for MPDAGs.

However, a key advantage of our work is that we consider a setting that allows the addition of expert knowledge, which follows research on the MPDAG setting seen in \cite{perkovic2020identifying} and \cite{laplante2024conditional}. This process of adding expert knowledge to causal graphs is largely excluded from research on identification in causal PAGs (e.g., \citealp{zhang2008causal, jaber2022causal}). Though we point out the work of \cite{venkateswaran2024towards}, which considers a setting that allows for both latent variables and the addition of expert knowledge. We hope that our results in combination with those of \cite{venkateswaran2024towards} and \cite{laplante2024conditional} can be used toward future research on conditional identification in the presence of both latent variables and expert knowledge.


\acks{This material is based upon work supported by the National Science Foundation under Grant No. 2210210.}

%
%

\newpage
\appendix

\centerline{\LARGE{Supplement to:}}
\centerline{\LARGE{Identifying Conditional Causal Effects in MPDAGs}}

\tableofcontents


\section{Further Definitions}
\label{app:defs}

\textbf{Concatenation.}
We denote the concatenation of paths by the symbol $\oplus$, so that for a path $p = \langle X_1,X_2, \dots, X_m \rangle$, $p = p(X_1, X_r) \oplus p(X_r, X_m)$, for $1\le r\le m$.

\begin{definition}
\label{def:bucket-decomp}
{\normalfont (\textbf{Bucket Decomposition}; \citealp{perkovic2020identifying})}
    Let $\mb{D}$ be a node set in an MPDAG $\g$. Then $\mb{B_1}, \dots, \mb{B_k}$, $k \ge 1$, is the bucket decomposition of $\mb{D}$ in $\g$ if $\mb{B_i}$, $i \in \{ 1, \dots, k\}$, is a bucket in $\mb{D}$ and $\cup_{i=1}^k \mb{B_i} = \mb{D}$.
\end{definition}

\begin{definition}
\label{def:distance}
{\normalfont (\textbf{Distance to $\mb{Z}$}; \citealp{perkovic2017interpreting})} 
    Let $\{X,Y\}$ and $\mb{Z}$ be disjoint node sets in an MPDAG $\g$, and let $p$ be a path in $\g$ from $X$ to $Y$ such that $\g$ has a possibly directed path (possibly of zero length) from every collider on $p$ to $\mb{Z}$. Further, let $\{C_1, \dots, C_k\}$ be the set of colliders on $p$, and let $\ell_i$, $i \in \{1, \dots, k\}$, be the length of a shortest possibly directed path in $\g$ from $C_i$ to $\mb{Z}$. Then the \emph{distance to $\mb{Z}$} for $p$ is $\sum_{i=1}^k \ell_i$.
\end{definition}


\section{Existing Results}
\label{app:existing}

\begin{theorem}
\label{thm:do-calc}
{\normalfont (\textbf{Rules of the do Calculus}; Theorem 3.4.1 of \cite{pearl2009causality})}
Let $\mb{X,Y,Z,}$ and $\mb{W}$ be pairwise disjoint (possibly empty) node sets in a causal DAG $\g[D]$. Let $\g[D]_{\overline{\mb{X}}}$ denote the graph obtained by deleting all edges into $\mb{X}$ from $\g[D]$. Similarly, let $\g[D]_{\underline{\mb{X}}}$ denote the graph obtained by deleting all edges out of $\mb{X}$ in $\g[D]$, and let $\g[D]_{\overline{\mb{X}}\underline{\mb{Z}}}$ denote the graph obtained by deleting all edges into $\mb{X}$ and all edges out of $\mb{Z}$ in $\g[D]$. The following rules hold for all densities consistent with $\g[D]$.

\textbf{Rule 1.} If $(\mb{Y} \dsepp \mb{Z} \given \mb{X}, \mb{W})_{\g[D]_{\overline{\mb{X}}}}$, then
    \begin{align}
        f(\mb{y} | do(\mb{x}), \mb{z,w}) = f(\mb{y} | do(\mb{x}), \mb{w}). \label{thm:do-rule1}
    \end{align}

\textbf{Rule 2.} If $(\mb{Y} \dsepp \mb{X} \given \mb{Z}, \mb{W})_{\g[D]_{\underline{\mb{X}}\overline{\mb{W}}}}$, then
    \begin{align}
        f(\mb{y} | do(\mb{x}),\mb{z},do(\mb{w})) = f(\mb{y} | \mb{x},\mb{z},do(\mb{w})). \label{thm:do-rule2}
    \end{align}

\textbf{Rule 3.} If $(\mb{Y} \dsepp \mb{X} \given \mb{Z}, \mb{W})_{\g[D]_{\overline{\mb{W}}, \overline{\mb{X}(\mb{Z})}}}$, then
    \begin{align}
        \begin{split}
            f(\mb{y} | do(\mb{x}), \mb{z}, do(\mb{w})) = f(\mb{y} | \mb{z}, do(\mb{w})), \label{thm:do-rule3}
        \end{split}
    \end{align}
where $\mb{X(Z)} := \mb{X} \setminus \An(\mb{Z}, \g[D]_{\overline{\mb{W}}})$.
\end{theorem}

\begin{lemma}
\label{lem:mardia}
{\normalfont (Theorem 3.2.4 of \citealp{mardia1980multivariate})}
    Let $\mb{X} = (\mb{X_1}^T, \mb{X_2}^T)^T$ be a $p$-dimensional multivariate Gaussian random vector with mean vector $\mb{\mu} = (\mb{\mu_1}^T, \mb{\mu_2}^T)^T$ and covariance matrix $\mb{\Sigma} = \begin{bmatrix}
    \mb{\Sigma_{11}} & \mb{\Sigma}_{12} \\
    \mb{\Sigma_{21}} & \mb{\Sigma}_{22}
    \end{bmatrix}$, so that $\mb{X_1}$ is a $q$-dimensional multivariate Gaussian random vector with mean vector $\mb{\mu_1}$ and covariance matrix $\mb{\Sigma_{11}}$ and $\mb{X_2}$ is a $(p-q)$-dimensional multivariate Gaussian random vector with mean vector $\mb{\mu_2}$ and covariance matrix $\mb{\Sigma_{22}}$. Then $E[\mb{X_2}|\mb{X_1} = \mb{x_1}] = \mb{\mu_2} + \mb{\Sigma_{21}} \mb{\Sigma_{11}}^{-1}(\mb{x_1} - \mb{\mu_1})$.
\end{lemma}

\begin{lemma}
\label{lem:wright}
{\normalfont (\textbf{Wright's Rule}; \citealp{wright1921correlation, wright1934method})}
    Consider the following structural equation model (SEM) over a vector of random variables $\mb{X} = (X_1, \dots, X_k)^T$. The set of equations is given by $\mb{X}=\mb{AX}+\mb{\varepsilon}$, where $\mb{A} \in \mathbb{R}^{k \times k}$ is a matrix with zeroes on the diagonal and coefficients from the SEM on the off-diagonals, $\mb{\varepsilon} = (\varepsilon_1, \dots, \varepsilon_k)^T$ is a vector of mutually independent errors with finite means, and $Var(X_i) = 1$ for all $i \in \{1, \dots, k\}$. Let $\g[D]$ be a DAG over $\mb{X}$ such that $X_i \to X_j$ is in $\g[D]$ if and only if $A_{ji} \neq 0$ for $i, j \in \{1, \dots, k\}$, $i \neq j$. (We call $A_{ji}$ the edge coefficient of $X_i \to X_j$ when $A_{ji} \neq 0$.) Further, let $\{p_1, \dots, p_s\}$ be the set of paths between $X_i$ and $X_j$ in $\g[D]$ that do not contain a collider, and let $\pi_r$ be the product of all edge coefficients along the path $p_r$, $r \in \{1, \dots, s\}$. Then $\Cov(X_i, X_j) = \sum_{r=1}^s \pi_r$.
\end{lemma}

\begin{lemma}
\label{lem:sem-covariance}
{\normalfont (cf. proof of Lemma B.4, \cite{henckel2022graphical})}
    Consider the following structural equation model (SEM) over a vector of random variables $\mb{X} = (X_1, \dots, X_k)^T$. The set of equations is given by $\mb{X}=\mb{AX}+\mb{\varepsilon}$, where $\mb{A} \in \mathbb{R}^{k \times k}$ is a matrix with zeroes on the diagonal and coefficients from the SEM on the off-diagonals, $\mb{\varepsilon} = (\varepsilon_1, \dots, \varepsilon_k)^T$ is a vector of mutually independent errors with finite means, and finite variances. Let $\g[D]$ be a DAG over $\mb{X}$ such that $X_i \to X_j$ is in $\g[D]$ if and only if $A_{ji} \neq 0$ for $i, j \in \{1, \dots, k\}$, $i \neq j$. (We call $A_{ji}$ the edge coefficient of $X_i \to X_j$ when $A_{ji} \neq 0$.) Then we can write
    \begin{align*}
        X_i             &\,\,=\,\,  \mspace{-50mu} \sum_{ \mspace{40mu} j: \, X_j \in \An(X_i, \g[D])} \mspace{-45mu} \tau_{ji} \varepsilon_j 
        \intertext{where we define $\tau_{ji}$ as follows. Let $\tau_{ii} = 1$. For the remaining $\tau_{ji}$, consider the set of all causal paths $\{p_1, \dots, p_s\}$ from  $X_j$ to $X_i$ in $\g[D]$, and let $\pi_r$ be the product of all edge coefficients along the path $p_r$, $r \in \{1, \dots, s\}$. Then let $\tau_{ji} = \sum_{r=1}^s \pi_r$. Further, we have}
        \Cov(X_i,X_j)   &\,\,=\,\,  \mspace{-120mu} \sum_{ \mspace{120mu} k: \, X_k \in \An(X_i, \g[D]) \, \cap\, \An(X_j, \g[D])} \mspace{-115mu} \tau_{ki} \tau_{kj} \Var(\varepsilon_k),
    \end{align*}
\end{lemma}

\begin{lemma} 
\label{lem:markov-equiv} 
{\normalfont (cf. Theorem 1 and Proposition 3 of \citealp{lauritzen1990independence})}
    Let $\g[D] = (\mb{V},\mb{E})$ be a DAG, and let $f$ be an observational density over $\mb{V}$. Then $f$ is Markov compatible with $\g[D]$ if and only if 
    \begin{align*}
        V_i \ind \Big[ \mb{V} \setminus \big( \De(V_i, \g[D]) \cup \Pa(V_i,\g[D]) \big) \Big] | \Pa(V_i, \g[D])
    \end{align*}
    for all $V_i \in \mb{V}$, where $\ind$ indicates independence with respect to $f$.
\end{lemma}

\begin{lemma} 
\label{lem:adding-edges-imply}
{\normalfont (cf.\ Lemma F.1 of \citealp{rothenhausler2018causal})}
    Let $X$ and $Y$ be nodes in an MPDAG $\g = (\mb{V,E})$ such that $X - Y$ is in $\g$. Let $\g'$ be an MPDAG constructed from $\g$ by adding $X \to Y$ and completing the orientation rules R1 - R4 of \cite{meek1995causal}. For any $Z,W \in \mb{V}$, if $Z - W$ is in $\g$ and $Z \rightarrow W$ is in $\g'$, then $W \in \De(Y,\g')$.
\end{lemma}

\begin{lemma}
\label{lem:undirected-imply}
{\normalfont (cf.\ Lemma F.2 of \citealp{rothenhausler2018causal})}
    Let $X$ be a node in an MPDAG $\g=(\mb{V}, \mb{E})$, and let $\mb{S}$ be a set such that for all $S \in \mb{S}$, $X - S$ is in $\g$. Then there is an MPDAG $\g' = (\mb{V}, \mb{E'})$ that is formed by taking $\g$, orienting $X \to S$ for all $S \in \mb{S}$, and completing R1-R4 of \cite{meek1995causal}.
\end{lemma}

\begin{lemma}
\label{lem:prop1meek}
{\normalfont (cf.\ Lemma 1 of \citealp{meek1995causal})}
    Let $X$, $Y$, and $Z$ be nodes in a CPDAG $\g[C]$. If $\g[C]$ contains $X \to Y - Z$, then $\g[C]$ contains $X \to Y$.
\end{lemma}

\begin{algorithm}[t]
    \vspace{.2cm}
    \Inputs{Node set $\mb{D} \subseteq \mb{V}$ and MPDAG $\g=(\mb{V,E})$}
    \Output{List of buckets $\mb{B} {=} (\mb{B_1}, \dots, \mb{B_k}), k \ge 1$, in $\mb{D}$ given $\g$}
    \vspace{.2cm}
    \SetAlgoLined
    Let $\mb{ConComp}$ be the bucket decomposition of $\mb{V}$ in $\g$ (Definition \ref{def:bucket-decomp});\\
    Let $\mb{B}$ be an empty list;\\
    \While{$\mb{ConComp}\neq \emptyset$}{
        Let $\mb{C} \in \mb{ConComp}$;\\
        Let $\overline{\mb{C}}$ be the set of nodes in $\mb{ConComp}$ that are not in $\mb{C}$;\\
        \If{all edges between $\mb{C}$ and $\overline{\mb{C}}$ are into $\mb{C}$ in $\g$}{
            Remove $\mb{C}$ from $\mb{ConComp}$;\\
            Let $\mb{B_{*}} = \mb{C} \cap \mb{D}$;\\
            \If{$\mb{B_{*}}\neq \emptyset$} { Add $\mb{B_{*}}$ to the beginning of $\mb{B}$;}
            }
        }
    \Return $\mb{B}$;
    \caption{PCO Algorithm of \cite{perkovic2020identifying}}
    \label{alg:pco}
\end{algorithm}

\begin{lemma}
\label{lem:pc-imply1b}
{\normalfont (Lemma 3.2 of \citealp{perkovic2017interpreting})}
    Let $X$ and $Y$ be nodes in an MPDAG $\g$ and let $p$ be a path from $X$ to $Y$ in $\g$. If there exists a DAG $\g[D] \in [\g]$ such that the path in $\g[D]$ corresponding to $p$ in $\g$ is causal, then $p$ is possibly causal.
\end{lemma}

\begin{lemma}
\label{lem:pc-def-stat}
{\normalfont (Lemma 3.5 of \citealp{perkovic2017interpreting})}
    Let $p = \langle V_1, \dots, V_k \rangle$, $k \ge 2$, be a definite status path in an MPDAG. Then $p$ is possibly causal if and only if $p$ has no edge $V_i \gets V_{i+1}$ for any $i \in \{1, \dots, k-1\}$. 
\end{lemma}

\begin{lemma}
\label{lem:PCO-buckets}
{\normalfont (Lemma 3.5 of \citealp{perkovic2020identifying})}
    Let $\mb{D}$ be a node set in an MPDAG $\g$ and let $\PCO(\mb{D}, \g)=(\mb{B_1}, \dots, \mb{B_k})$, $k \ge 1$, be the output of the PCO Algorithm (Algorithm \ref{alg:pco}). Then for $i,j \in \{1, \dots, k\}$, $\mb{B_i}$ and $\mb{B_j}$ are buckets in $\mb{D}$, and if $i < j$, then $\mb{B_i} < \mb{B_j}$ in $\g$. 
\end{lemma}

\begin{lemma}
\label{lem:bucket-req1}
{\normalfont (Lemma C.2 of \citealp{perkovic2020identifying})}
    Let $\mb{C}$ be a bucket in $\mb{V}$ in an MPDAG $\g = (\mb{V},\mb{E})$ and let $X \in \mb{V} \setminus \mb{C}$. If there is a causal path from $X$ to $\mb{C}$ in $\g$, then for every node $C \in \mb{C}$, there is a causal path from $X$ to $C$ in $\g$.
\end{lemma}

\begin{lemma}
\label{lem:bucket-req2}
{\normalfont (cf. Proof of Lemma C.2 of \citealp{perkovic2020identifying})}
    Let $\mb{C}$ be a bucket in $\mb{V}$ in an MPDAG $\g = (\mb{V},\mb{E})$ and let $X \in \mb{V} \setminus \mb{C}$. If $X \in \Pa(\mb{C}, \g)$, then $X \in \Pa(C, \g)$ for every node $C \in \mb{C}$.
\end{lemma}

\begin{lemma}
\label{lem:xtod}
{\normalfont (Lemma D.1(i) of \citealp{perkovic2020identifying})}
    Let $\mb{X}$ and $\mb{Y}$ be disjoint node sets in an MPDAG $\g =(\mb{V,E})$, where there is no proper possibly causal path from $\mb{X}$ to $\mb{Y}$ that starts with an undirected edge in $\g$. Then there is no proper possibly causal path from $\mb{X}$ to $\An(\mb{Y},\g_{\mb{V} \setminus \mb{X}})$ that starts with an undirected edge in $\g$.
\end{lemma}


\section{Broad MPDAG Results}
\label{app:mpdags}

This section presents broad results for MPDAGs used in the proofs of Appendix \ref{app:id-formula}. We include these results here rather than in Appendix \ref{app:id-formula}, because they are likely useful beyond the scope of this paper.


\subsection{Results on Possibly Causal Paths}

\begin{lemma}
\label{lem:shortest-subseq}
{\normalfont (cf.\ Lemma 3.6 of \citealp{perkovic2017interpreting})}
    Let $X$ and $Y$ be distinct nodes in an MPDAG $\g$ and let $p$ be a possibly causal path from $X$ to $Y$ in $\g$. Then any shortest subsequence of $p$ in $\g$ forms an unshielded, possibly causal path from $X$ to $Y$. 
\end{lemma}

\begin{proofof}[Lemma \ref{lem:shortest-subseq}]    
    Let $k$ be the number of nodes on $p$. Pick an arbitrary shortest subsequence of $p$ and call it $p^* :=\langle X=V_0, \dots, V_\ell=Y \rangle$, $0 < \ell \le k$. Note that there is no edge $V_i \gets V_j, 0 \le i < j \le k$ in $\g$, since this would contradict that $p$ is possibly causal. Thus, $p^*$ is also possibly causal. Further, note that $p^*$ is unshielded, since if any triple on the path is shielded, it either contradicts that $p^*$ is possibly causal (i.e., $V_i \gets V_{i+2}$ cannot be in $p^*$) or that $p^*$ is a shortest subsequence of $p$ (i.e., $V_i \to V_{i+2}$ and $V_i - V_{i+2}$ cannot be in $p^*$).
\end{proofof}


\begin{lemma} 
\label{lem:nobackpaths}
    Let $p=\langle P_0, \dots, P_k \rangle$ be a path in an MPDAG $\g$. Then $p$ is possibly causal if and only if $\g$ does not contain any path $P_i \gets \dots \gets P_j$, $0 \le i < j \le k$.
\end{lemma}

\begin{proofof}[Lemma \ref{lem:nobackpaths}]
    $\Leftarrow$ Follows immediately.
    $\Rightarrow$ Let $p$ be possibly causal, and for sake of contradiction, suppose $\g$ contains a path $q$ from $P_i$ to $P_j$, $0 \le i < j \le k$, of the form $P_i=Q_0 \gets \dots \gets Q_{\ell} =P_j$. Then define $r=\langle P_i = R_0, \dots, R_m= P_j\rangle$ to be a shortest subsequence of $p(P_i,P_j)$ in $\g$, and note that by Lemma \ref{lem:shortest-subseq}, $r$ is an unshielded, possibly causal path.

    To see that $r(R_0, R_1)$ takes the form $R_0 - R_1$, note that $\g$ cannot contain $R_0 \gets R_1$, since $r$ is possibly causal. Further, $\g$ cannot contain $R_0 \to R_1$, since $r$ being unshielded would imply, by R1 of \cite{meek1995causal}, that $\g$ contains the cycle $P_i= R_0 \to \dots \to R_m=P_j=Q_{\ell} \to \dots \to Q_0=P_i$. Then since $\g$ contains $R_0 - R_1$, there must be a DAG $\g[D] \in [\g]$ that contains $R_0 \to R_1$. But since $r$ is unshielded, this implies, by R1 of \cite{meek1995causal}, that $\g[D]$ contains the cycle $P_i= R_0 \to \dots \to R_m=P_j=Q_{\ell} \to \dots \to Q_0=P_i$, which is a contradiction.
\end{proofof}


\begin{lemma}
\label{lem:concat}
    Let $X$, $Y$, and $Z$ be distinct nodes in an MPDAG $\g$.
    \begin{enumerate}[label=(\roman*)]
        \item \label{lem:concat-pc-c} 
        If $p$ is a possibly causal path from $X$ to $Y$ and $q$ is a causal path from $Y$ to $Z$, then $p \oplus q$ is a possibly causal path from $X$ to $Z$.

        \item \label{lem:concat-c-pc} 
        If $p$ is a causal path from $X$ to $Y$ and $q$ is a possibly causal path from $Y$ to $Z$, then $p \oplus q$ is a possibly causal path from $X$ to $Z$.
    \end{enumerate}
\end{lemma}

\begin{proofof}[Lemma \ref{lem:concat}]
    Let $p= \langle X=P_0, \dots, P_k=Y \rangle$ and $q= \langle Y=Q_0, \dots, Q_r=Z \rangle$, where either \ref{lem:concat-pc-c} $p$ is possibly causal and $q$ is causal, or \ref{lem:concat-c-pc} $q$ is possibly causal and $p$ is causal. To see that $p \oplus q$ exists, suppose for sake of contradiction that $p$ and $q$ share at least one node other than $Y$. Let $\mb{S}$ denote the collection of such nodes, and consider the node in $\mb{S}$ with the lowest index on $q$. That is, consider $Q_j \in \mb{S}$ such that $j \le \ell$ for all $Q_{\ell} \in \mb{S}$. Let $Q_j=P_i$ for some $P_i \neq Y$ on $p$. Under \ref{lem:concat-pc-c}, $q$ is causal so that $\g$ contains $P_k=Q_0 \to \dots \to Q_j=P_i$. But by Lemma \ref{lem:nobackpaths}, this contradicts that $p$ is possibly causal. Similarly, under \ref{lem:concat-c-pc}, $p$ is causal so that $\g$ contains $Q_j=P_i \to \dots \to P_k = Q_0$, which contradicts that $q$ is possibly causal.    

    To complete the proof, we show that there is no backward edge between any two nodes on $p \oplus q$. By the choice of $p$ and $q$, note that there is no edge $P_{i_1} \gets P_{j_1}$ for $0 \le i_1 < j_1 \le k$ in $\g$, and there is no edge $Q_{i_2} \gets Q_{j_2}$ for $0 \le i_2 < j_2 \le r$ in $\g$. Thus, suppose for sake of contradiction that there exists an edge $P_i \gets Q_j$ in $\g$ for $i \in \{0, \dots, k-1\}$ and $j \in \{1, \dots, r\}$. Note that $P_i$ is on $p$ and not $q$, and analogously, $Q_j$ is on $q$ and not $p$, since we have shown $p$ and $q$ cannot share nodes other than $Y$.

    \textbf{CASE \ref{lem:concat-pc-c}:}
    Note that $p(P_i, Y)$ is possibly causal. Then pick an arbitrary shortest subsequence of $p(P_i, Y)$ and call it $t :=\langle P_i=T_0, \dots, T_m=Y \rangle$, $m \ge 1$. By Lemma \ref{lem:shortest-subseq}, $t$ forms an unshielded, possibly causal path. Since $t$ is possibly causal, $\g$ must contain $T_0 \to T_1$ or $T_0 - T_1$. Thus, there is a DAG $\g[D] \in [\g]$ that contains the edge $T_0 \to T_1$. Further, since $t$ is unshielded, $\g[D]$ contains $T_0 \to \dots \to T_m$ by R1 of \cite{meek1995causal}. But this is a contradiction, since $q$ is causal and thus $\g[D]$ contains the cycle $P_i=T_0 \to \dots \to {T_m=Y=Q_0} \to \dots \to Q_j \to P_i$.

    \textbf{CASE \ref{lem:concat-c-pc}:}
    Note that $q(Y, Q_j)$ is possibly causal. Then pick an arbitrary shortest subsequence of $q(Y, Q_j)$ and call it $t :=\langle Y=T_0, \dots, T_m=Q_j \rangle$, $m \ge 1$. By Lemma \ref{lem:shortest-subseq}, $t$ forms an unshielded, possibly causal path. Since $t$ is possibly causal, $\g$ must contain $T_0 \to T_1$ or $T_0 - T_1$. Thus, there is a DAG $\g[D] \in [\g]$ that contains the edge $T_0 \to T_1$. Further, since $t$ is unshielded, $\g[D]$ contains $T_0 \to \dots \to T_m$ by R1 of \cite{meek1995causal}. But this is a contradiction, since $p$ is causal and thus $\g[D]$ contains the cycle ${Y=T_0} \to \dots \to {T_m=Q_j} \to P_i \to \dots \to {P_k=Y}$.
\end{proofof}

\begin{figure}[h!t]
	\centering
	\begin{tikzpicture}[>=stealth',shorten >=1pt,auto,node distance=2cm,main node/.style={minimum size=0.6cm,font=\sffamily\Large\bfseries},scale=1.3,transform shape=2]
	\tikzstyle{state}=[inner sep=3pt, minimum size=5pt]
	
	\node[state] (Y) at (0,2) {$Y$};
	\node[state] (X) at (2,2) {$X$};
	\node[state] (V2) at (0,0) {$V_2$};
	\node[state] (V1) at (2,0) {$V_1$};
	
	\draw[-] (Y) edge (X);
	\draw[-] (X) edge (V1);
	\draw[-] (V2) edge (Y);
	\draw[-] (V1) edge (V2);
	\draw[-] (Y) edge (V1);
	\end{tikzpicture}
\caption{An MPDAG used in Remark \ref{rem:pc-imply1a}.}
\label{fig:pc-imply1a}
\end{figure}
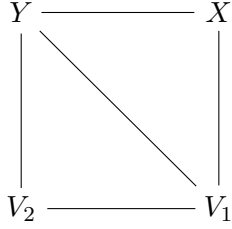

\begin{remark}
\label{rem:pc-imply1a} 
    Perhaps contrary to intuition, even if a path is possibly causal in an MPDAG $\g$, there may not exist a DAG in $[\g]$ where the corresponding path is causal.
    
    To see this, consider the MPDAG $\g$ in Figure \ref{fig:pc-imply1a} and let $p = \langle X, V_1, V_2, Y \rangle$. Note that $p$ is possibly causal. Let $\g[D]$ be an arbitrary DAG in $\g$ and define $p^*$ to be the path in $\g[D]$ corresponding to $p$ in $\g$. If $p^*$ begins $X \gets V_1$, it is non-causal. Thus, consider when $p^*$ begins $X \to V_1$. It follows that $\g[D]$ contains $V_1 \to V_2$ by R1 of \cite{meek1995causal} as well as $V_2 \gets Y$ by R4 of \cite{meek1995causal}, so that again, $p^*$ is non-causal.
\end{remark}

\begin{lemma} 
\label{lem:pc-imply2}
    Let $X$ and $Y$ be nodes in an MPDAG $\g$. Then there is a possibly causal path from $X$ to $Y$ in $\g$ if and only if there is at least one DAG in $[\g]$ with a causal path from $X$ to $Y$.
\end{lemma}

\begin{proofof}[Lemma \ref{lem:pc-imply2}]
    $\Leftarrow$ Holds by Lemma \ref{lem:pc-imply1b}.
    $\Rightarrow$ Let $p$ be a possibly causal path from $X$ to $Y$ in $\g=(\mb{V},\mb{E})$ and let $p^*=\langle X=V_0, \dots, V_k=Y \rangle$, $k \ge 1$, be a shortest subsequence of $p$ in $\g$. By Lemma \ref{lem:shortest-subseq}, $p^*$ forms an unshielded, possibly causal path. Since $p^*$ is possibly causal, $\g$ must contain $V_0 \to V_1$ or $V_0 - V_1$. Thus, there is a DAG $\g[D] \in [\g]$ that contains the edge $V_0 \to V_1$. Further, since $p^*$ is unshielded, $\g[D]$ contains $X=V_0 \to \dots \to V_k=Y$ by R1 of \cite{meek1995causal}.
\end{proofof}


\subsection{Remarks on the Ordering of Buckets}

\begin{remark} 
    Note that the output of the PCO Algorithm (Algorithm \ref{alg:pco}) is an ordered bucket decomposition (Definition \ref{def:obd}) of the input node set in the input MPDAG. This follows from the steps of Algorithm \ref{alg:pco} and Lemma \ref{lem:PCO-buckets}.
\end{remark}

\begin{definition}[Bucket Decomposition Matching]
\label{def:obd-matching}
    Let $\mb{D}$ be a node set in an MPDAG $\g=(\mb{V},\mb{E})$. Let $(\mb{B_1}, \dots, \mb{B_k}), k \ge 1$, be an ordered bucket decomposition of $\mb{D}$ in $\g$ (Definition \ref{def:obd}), and let $(\mb{C_1}, \dots, \mb{C_{\ell}})$, $\ell \ge k$, be an ordered bucket decomposition of $\mb{V}$ in $\g$. We say that $(\mb{C_1}, \dots, \mb{C_{\ell}})$ is an ordered bucket decomposition that \emph{matches the ordering} of $(\mb{B_1}, \dots, \mb{B_k})$ if the following holds. For any $i,j \in \{1, \dots, k \}$ and $m, n \in \{ 1, \dots, \ell \}$ such that $\mb{B_i} \cap \mb{C_m} \neq \emptyset$ and $\mb{B_j} \cap \mb{C_n} \neq \emptyset$, if $\mb{B_i} < \mb{B_j}$, then $\mb{C_m} < \mb{C_n}$.
\end{definition}

\begin{remark}
\label{rem:obd-matching}
    Let $\mb{D}$ be a node set in the MPDAG $\g = (\mb{V}, \mb{E})$, and let $PCO(\mb{D}, \g) =(\mb{B_1}, \dots, \mb{B_k})$, $k \ge 1$, be the output of the PCO Algorithm (Algorithm \ref{alg:pco}). Since the algorithm relies on the intersection of $\mb{D}$ with an ordering of the bucket decomposition of $\mb{V}$ in $\g$, there must exist an ordered bucket decomposition of $\mb{V}$ in $\g$ (Definition \ref{def:obd}) that matches the ordering (Definition \ref{def:obd-matching}) of $(\mb{B_1}, \dots, \mb{B_k})$.
\end{remark}


\subsection{Results on Paths Among Buckets}

\begin{lemma} 
\label{lem:bucket-req3}
    Let $\g=(\mb{V},\mb{E})$ be an MPDAG, and let $\mb{C_1}, \dots, \mb{C_{\ell}}$, $\ell \ge 1$, be the bucket decomposition of $\mb{V}$ in $\g$ (Definition \ref{def:bucket-decomp}). Further, let $X$ and $Y$ be nodes in $\g$ such that $X \in \mb{C_x}$ and $Y \in \mb{C_y}$ for some $x,y \in \{1, \dots, \ell\}$, $x \neq y$. If $p = \langle X=P_0, \dots, P_k=Y\rangle, k \ge 1$, is a proper possibly causal path from $\mb{C_x}$ to $\mb{C_y}$ in $\g$, then $\g$ contains a subsequence of $p$ that is a causal path from $X$ to $Y$.
\end{lemma}

\begin{proofof}[Lemma \ref{lem:bucket-req3}]
    Let $p = \langle X=P_0, \dots, P_k=Y\rangle, k \ge 1$, be a proper possibly causal path from $\mb{C_x}$ to $\mb{C_y}$ in $\g$. Since $p$ is proper, $X$ is the only node in $\mb{C_x}$ on $p$. Thus $P_1$ is the first node in the bucket $\mb{C_a}$ where $a \in \{1, \dots, \ell \} \setminus \{x\}$. By the definition of buckets and the fact that $p$ is possibly causal, $p$ contains $X \to P_1$. Then by Lemma \ref{lem:bucket-req2}, $\g$ contains $X \to A$ for all $A \in \mb{C_a}$.

    When $Y \in \mb{C_a}$, $\g$ contains the subsequence $X \to Y$ and we are done. When $Y \notin \mb{C_a}$, define $P_q$, $q \in \{1, \dots, k \}$, to be the last node on $p$ in $\mb{C_a}$. Note that $\g$ contains $X \to P_q$ and consider the node $P_{q+1}$. By Lemma \ref{lem:bucket-restrict1}, $p(P_1, P_q)$ is entirely contained in $\mb{C_a}$. Thus, $P_{q+1}$ is the first node in the bucket $\mb{C_b}$ where $b \in \{1, \dots, \ell\} \setminus \{a, x\}$. By the definition of buckets and the fact that $p$ is possibly causal, $p$ contains $P_q \to P_{q+1}$. Then by Lemma \ref{lem:bucket-req2}, $\g$ contains $P_q \to B$ for all $B \in \mb{C_b}$.

    When $Y \in \mb{C_b}$, $\g$ contains the subsequence $X \to P_q \to Y$ and we are done. When $Y \notin \mb{C_b}$, we can continue in this way until we reach the first node in $\mb{C_y}$---call it $P_{r+1}$. By the same logic as above, $p$ must contain $P_r \to P_{r+1}$, where $P_r \notin \mb{C_y}$ and where $\g$ contains $X \to P_q \to \dots \to P_r$. So again by Lemma \ref{lem:bucket-req2}, $\g$ contains $P_r \to W$ for all $W \in \mb{C_y}$ and thus $\g$ contains the subsequence $X \to P_q \to \dots \to P_r \to Y$. 
\end{proofof}

\begin{lemma}
\label{lem:bucket-restrict1}
    Let $\g=(\mb{V},\mb{E})$ be an MPDAG, and let $\mb{C_1}, \dots, \mb{C_k}, k \ge 1$, be the bucket decomposition of $\mb{V}$ in $\g$ (Definition \ref{def:bucket-decomp}). If $X, Y \in \mb{C_i}$, $i \in \{1, \dots, k\}$, then there is no possibly causal path from $X$ to $Y$ in $\g$ containing nodes in $\mb{V} \setminus \mb{C_i}$.
\end{lemma}

\begin{proofof}[Lemma \ref{lem:bucket-restrict1}]
    For sake of contradiction, suppose that there is a possibly causal path from $X$ to $Y$ in $\g$ containing nodes in $\mb{V} \setminus \mb{C_i}$---call this path $p=\langle X=P_0, \dots, P_u=Y \rangle$, $u \ge 1$. Let $P_s$, $s \in \{1, \dots, u-1\}$, be an arbitrary node on $p$ such that $P_s \in \mb{V} \setminus \mb{C_i}$. Let $P_r$, $r \in \{0, \dots, s-1\}$ be the node on $p$ closest to $P_s$ such that $P_r \in \mb{C_i}$. And let $P_t$, $t \in \{s+1, \dots, u\}$ be the node on $p$ closest to $P_s$ such that $P_t \in \mb{C_i}$.

    By the definition of $P_t$, we have that $P_{t-1} \notin \mb{C_i}$ and $P_t \in \mb{C_i}$. Consider the edge between $P_{t-1}$ and $P_t$. Since $\mb{C_i}$ is a maximal undirected connected subset of $\mb{V}$, then $p$ does not contain $P_{t-1} - P_t$. Further since $p$ is possibly causal, it does not contain $P_{t-1} \gets P_t$. Therefore $p$ must contain $P_{t-1} \to P_t$. But then by Lemma \ref{lem:bucket-req2}, $\g$ must contain $P_{t-1} \to P_r$, which contradicts that $p$ is possibly causal.
\end{proofof}

\begin{lemma} 
\label{lem:bucket-restrict2}
    Let $\g=(\mb{V},\mb{E})$ be an MPDAG and let $(\mb{C_1}, \dots, \mb{C_k}), k \ge 1$, be an ordered bucket decomposition of $\mb{V}$ in $\g$ (Definition \ref{def:obd}). Then for any $i, j \in \{1, \dots, k\}, i < j$, there is no possibly causal path from $\mb{C_j}$ to $\mb{C_i}$ in $\g$.
\end{lemma} 

\begin{proofof}[Lemma \ref{lem:bucket-restrict2}]
    For sake of contradiction, let there be a possibly causal path from $\mb{C_j}$ to $\mb{C_i}$ in $\g$. Choose a shortest such path $p :=\langle P_0, \dots, P_u \rangle$, $u \ge 1$, and let $\mb{S}$ be the smallest collection of buckets in $\mb{V}$ such that every node on $p$ is in one bucket in $\mb{S}$. Consider the order in which $p$ enters and exits each bucket in $\mb{S}$. $p$ begins in $\mb{C_j}$, but once it exits $\mb{C_j}$, it never returns. This holds by Lemma \ref{lem:bucket-restrict1} and the fact that $p$ is possibly causal. By the same logic, once $p$ enters any other bucket in $\mb{S}$, the first time it exits that bucket, it will never return. Thus, let $\mb{C_{\ell_1}}, \dots, \mb{C_{\ell_m}}$, $1 < m \le k$, be an arrangement of the buckets in $\mb{S}$ in order of their appearance on $p$.
    
    Note that $P_0$ is the only node on $p$ in $\mb{C_j}$ by the choice of $p$ as a shortest path. Therefore, $p$ is a proper possibly causal path from $\mb{C_j}$ to $\mb{C_i}$. It follows by Lemma \ref{lem:bucket-req3} and the choice of $p$ as a shortest path that $p$ is causal. Based on the existence of this causal path, any partial causal ordering of $\mb{V}$ consistent with $\g$ must require $\mb{C_{\ell_1}} < \dots < \mb{C_{\ell_m}}$. Since $\ell_1 = j$ and $\ell_m = i$ by the definition of $p$, then any partial causal ordering of $\mb{V}$ consistent with $\g$ must require $\mb{C_j} < \mb{C_i}$. But this contradicts the definition of $(\mb{C_1}, \dots, \mb{C_k})$ as an ordered bucket decomposition of $\mb{V}$ in $\g$, where $\mb{C_i} < \mb{C_j}$.
\end{proofof}

\begin{corollary} 
\label{cor:bucket-restrict2a}
    Let $\mb{D}$ be a node set in an MPDAG $\g$ and let $(\mb{B_1}, \dots, \mb{B_k}), k \ge 1$, be the output of $PCO(\mb{D}, \g)$. Then for any $i, j \in \{1, \dots, k\}, i < j$, there is no possibly causal path from $\mb{B_j}$ to $\mb{B_i}$ in $\g$.
\end{corollary}

\begin{proofof}[Corollary \ref{cor:bucket-restrict2a}]
    Let $(\mb{C_1}, \dots, \mb{C_{\ell}})$, $\ell \ge k$, be the ordered bucket decomposition of $\mb{V}$ in $\g=(\mb{V},\mb{E})$ that matches the ordering of $(\mb{B_1}, \dots, \mb{B_k})$ (see Remark \ref{rem:obd-matching}). Let $\mb{B_i} \cap \mb{C_m} \neq \emptyset$ and $\mb{B_j} \cap \mb{C_n} \neq \emptyset$, where $m,n \in \{1, \dots, \ell\}$. By construction, $m<n$. If there were a possibly causal path from $B_j \in \mb{B_j}$ to $B_i \in \mb{B_i}$ in $\g$, then there would be a possibly causal path from $B_j \in \mb{C_n}$ to $B_i \in \mb{C_m}$ in $\g$, which contradicts Lemma \ref{lem:bucket-restrict2}.
\end{proofof}

\begin{corollary} 
\label{cor:bucket-restrict2b}
    Let $\g=(\mb{V},\mb{E})$ be an MPDAG and let $(\mb{C_1}, \dots, \mb{C_k}), k \ge 1$, be an ordered bucket decomposition of $\mb{V}$ in $\g$. Then for any $i, j \in \{1, \dots, k\}, i < j$, there is no DAG in $[\g]$ with a causal path from $\mb{C_j}$ to $\mb{C_i}$.
\end{corollary}

\begin{proofof}[Corollary \ref{cor:bucket-restrict2b}]
    For sake of contradiction, suppose there is a DAG in $[\g]$ with a causal path $p$ from $\mb{C_j}$ to $\mb{C_i}$. Then by Lemma \ref{lem:pc-imply1b}, the path in $\g$ corresponding to $p$ is possibly causal from $\mb{C_j}$ to $\mb{C_i}$, which contradicts Lemma \ref{lem:bucket-restrict2}.
\end{proofof}

\begin{corollary} 
\label{cor:bucket-restrict2c}
    Let $\mb{D}$ be a node set in an MPDAG $\g$ and let $(\mb{B_1}, \dots, \mb{B_k}), k \ge 1$, be the output of $PCO(\mb{D}, \g)$. Then for any $i, j \in \{1, \dots, k\}, i < j$, there is no DAG in $[\g]$ with a causal path from $\mb{B_j}$ to $\mb{B_i}$.
\end{corollary}

\begin{proofof}[Corollary \ref{cor:bucket-restrict2c}]
    For sake of contradiction, suppose there is a DAG in $[\g]$ with a causal path $p$ from $\mb{B_j}$ to $\mb{B_i}$. Then by Lemma \ref{lem:pc-imply1b}, the path in $\g$ corresponding to $p$ is possibly causal from $\mb{B_j}$ to $\mb{B_i}$, which contradicts Corollary \ref{cor:bucket-restrict2a}.
\end{proofof}


\section{Proofs for Section \ref{sec:id-formula}: Identification Formula}
\label{app:id-formula}

This section includes the statement and proof of Lemma \ref{lem:algo-4}, which provides the core justification for the proof of Theorem \ref{thm:id-formula} found in Section \ref{sec:id-formula}. Three supporting results needed for the proof of Lemma \ref{lem:algo-4} follow.


\subsection{Main Result}

\begin{lemma}[\textbf{Setup for Theorem \ref{thm:id-formula}}]
\label{lem:algo-4} 
    Let $\mb{X}$, $\mb{Y}$, and $\mb{Z}$ be pairwise disjoint node sets in a causal MPDAG $\g =(\mb{V,E})$, where $\mb{Z} \cap \PossDe(\mb{X},\g) = \emptyset$ and where there is no proper possibly causal path from $\mb{X}$ to $\mb{Y}$ that starts with an undirected edge in $\g$. Let $\mb{D} =\An(\mb{Y},\g_{\mb{V} \setminus \mb{X}}) \setminus \mb{Z}$, let $(\mb{B_1}, \dots, \mb{B_k}) = \PCO(\mb{D},\g)$, $k \ge 1$, and let $\mb{B_0} = \emptyset$. Then the following hold for every density $f$ consistent with $\g$, and every $i \in \{1, \dots, k\}$.
    
    \begin{enumerate}[label=(\roman*)]
        \item \label{lem:algo-4-one} 
        If $\mb{Z} \cap \PossDe(\mb{B_i}, \g) = \emptyset$, then
        \begin{align}
            f(\mb{b_i}| \mb{b_{i-1}}, \dots, \mb{b_0}, do(\mb{x}),\mb{z}) = f(\mb{b_i}|\pa(\mb{b_i},\g))
        \end{align}
        for values $\pa(\mb{b_i},\g)$ of $\Pa(\mb{B_i},\g)$ that are in agreement with $\mb{x}$.
        
        \item \label{lem:algo-4-two}
        If $\mb{Z} \cap \PossDe(\mb{B_i}, \g) \neq \emptyset$, then
        \begin{align}
            f(\mb{b_i}| \mb{b_{i-1}}, \dots, \mb{b_0}, do(\mb{x}), \mb{z}) = f(\mb{b_i}| \mb{b_i^N}, \mb{z}),
        \end{align}
        where $\mb{B_i^N}= \cup_{j=0}^{i-1}\mb{B_j} \setminus \PossDe(\mb{X},\g)$.
    \end{enumerate}
\end{lemma}

\begin{proofof}[Lemma \ref{lem:algo-4}]
    Let $\g[D]$ be an arbitrary DAG in $[\g]$, let $f$ be an arbitrary density consistent with $\g$, and pick $i \in \{1, \dots, k\}$. Then let $(\mb{C_1}, \dots, \mb{C_{\ell}}), \ell \ge k$, be the ordered bucket decomposition of $\mb{V}$ in $\g$ that matches the ordering of $(\mb{B_1}, \dots, \mb{B_k})$ (see Remark \ref{rem:obd-matching}). Further, let $\mb{C_n} \cap \mb{B_i} \neq \emptyset$, $n \in \{1, \dots, \ell\}$, so that $\mb{B_i} \subseteq \mb{C_n}$ and $\cup_{j=0}^{i-1}\mb{B_j} \subseteq \cup_{j=1}^{n-1}\mb{C_j}$.

    \vskip .2in
    
    
    \textbf{PART \ref{lem:algo-4-one}}
    Note that this result is analogous to Lemma D.1 of \cite{perkovic2020identifying} for the unconditional causal effect setting. Start by defining the following sets.
    \begin{align*}
        \mb{P_i}      \,=\, &(\cup_{j=0}^{i-1}\mb{B_j} \cup \mb{Z}) \cap \Pa(\mb{B_i}, \g). &&
        \mb{X_{p_i}}  \,=\, \mb{X} \cap \Pa(\mb{B_i}, \g). \\
        \mb{N_i}      \,=\, &(\cup_{j=0}^{i-1} \mb{B_j} \cup \mb{Z}) \setminus \Pa(\mb{B_i}, \g). &&
        \mb{X_{n_i}}  \,=\, \mb{X} \setminus \Pa(\mb{B_i}, \g). \\[0.1in] 
        &&&
        \mb{X'_{n_i}} \,=\, \mb{X_{n_i}} \setminus \An(\mb{P_i}, \g[D]_{\overline{\mb{X_{p_i}}}}).
    \end{align*}
    We show the result in the three steps below.
    \begin{align}
        f(\mb{b_i}| \mb{b_{i-1}}, \dots, \mb{b_0}, do(\mb{x}),\mb{z})
                &= f(\mb{b_i}| \mb{p_i}, do(\mb{x})) \label{eq:algo-4-one-1}\\
                &= f(\mb{b_i}| \mb{p_i}, do(\mb{x_{p_i}})) \label{eq:algo-4-one-2}\\
                &= f(\mb{b_i}|\pa(\mb{b_i},\g)). \label{eq:algo-4-one-3}
    \end{align}
    In order to use Rules 1, 3, and 2 of the do calculus (Theorem \ref{thm:do-calc}) to show the equalities in \eqref{eq:algo-4-one-1}, \eqref{eq:algo-4-one-2}, and \eqref{eq:algo-4-one-3}, respectively, we must show the following.
    \begin{align}
        &(\mb{B_i} \dsepp \mb{N_i} \,\,\, | \mb{X}, \mb{P_i})_{\g[D]_{\overline{\mb{X}}}}. \label{eq:algo-4-one-4}\\
        &(\mb{B_i} \dsepp \mb{X_{n_i}} | \mb{X_{p_i}}, \mb{P_i})_{\g[D]_{\overline{\mb{X_{p_i}}\mb{X'_{n_i}}}}}. \label{eq:algo-4-one-5}\\
        &(\mb{B_i} \dsepp \mb{X_{p_i}} | \mb{P_i})_{\g[D]_{\underline{\mb{X_{p_i}}}}}. \label{eq:algo-4-one-6}
    \end{align}
    We first show a broader independence claim and complete the proof by showing that each of the independence statements above is a special case of the broader claim.

    \vskip .1in

    
    \textbf{Broader Claim:}
    Define the sets $\mb{N}$, $\mb{H}$, $\mb{X'}$, and $\mb{N'}$ such that $\mb{N}$, $\mb{H}$, $\mb{X'}$, $\mb{B_i}$, and $\mb{P_i}$ are pairwise disjoint and such that the following hold.
    \begin{align}
        \mb{N}  \phantom{,}&\subseteq\phantom{,} 
                \mb{N_i} \cup \mb{X_{n_i}}
                \,=\, \big(\cup_{j=1}^{i-1} \mb{B_j} \cup \mb{Z} \cup \mb{X} \big) \setminus \Pa(\mb{B_i}, \g). \nonumber\\
        \mb{H}  \phantom{,}&\subseteq\phantom{,}
                \mb{X_{p_i}}
                \,=\, \mb{X} \cap \Pa(\mb{B_i},\g). \nonumber\\
        \mb{X'} \phantom{,}&\subseteq\phantom{,} 
                \mb{X}. \nonumber\\
        \mb{N'} \phantom{,}&\subseteq\phantom{,} 
                \mb{X_{n_i}} \setminus \An(\mb{P_i}, \g[D]_{\overline{\mb{X'}}}). \nonumber\\[0.1in]
        \Pa(\mb{B_i},\g) \phantom{,}&\subseteq\, \mb{X'} \cup \mb{P_i} \cup \mb{H}. \nonumber\\
        \mb{N} \setminus \mb{N'} \phantom{,}&\subseteq\, \cup_{j=1}^{n-1} \mb{C_j} \cup \mb{Z}. \label{eq:algo-4-one-7}
    \end{align}
    The broader claim we will show is
    \begin{align}
        &(\mb{B_i} \dsepp \big[\mb{N} \cup \mb{H}\big] | \mb{X'}, \mb{P_i})_{\g[D]_{\overline{\mb{X'}\mb{N'}}\underline{\mb{H}}}}.\label{eq:algo-4-one-9}
    \end{align}
    
    Before beginning the proof of this broader claim, note that from the assumptions above, we also have the following.
    \begin{align*}
        \text{There is no poss}&\text{ibly causal path in } \g \text{ from } \label{eq:algo-4-one-8} \tag{$\ast$}\\
        \mb{B_i} \text{ to } &\big(\mb{N} \cup \mb{H} \cup \mb{P_i} \big) \setminus \mb{N'}. \nonumber
    \end{align*}
    To see this, note that $\mb{H} \subseteq \mb{X} \cap \Pa(\mb{B_i},\g)$, and by Lemmas \ref{lem:bucket-restrict2} and \ref{lem:algo-4efghi}\ref{lem:algo-4e}, $\big[ \mb{X} \cap \Pa(\mb{B_i},\g) \big] \cap \big( \cup_{j=n}^{\ell} \mb{C_j} \big) = \emptyset$. Thus, $\mb{H} \subseteq \cup_{j=1}^{n-1} \mb{C_j}$. Then by \eqref{eq:algo-4-one-7} and the definition of $\mb{P_i}$, we have $\big( \mb{N} \cup \mb{H} \cup \mb{P_i} \big) \setminus \mb{N'} \subseteq \cup_{j=1}^{n-1} \mb{C_j} \cup \mb{Z}$. The result follow by the fact that $\mb{B_i} \subseteq \mb{C_n}$, Lemma \ref{lem:bucket-restrict2}, and the fact that $\mb{Z} \cap \PossDe(\mb{B_i}, \g) = \emptyset$. 

    \vskip .1in
    
    We turn to prove the broader claim using a strategy similar to that of the proof of Lemma D.1 of \cite{perkovic2020identifying}. For sake of contradiction, suppose that there is a path from $\mb{B_i}$ to $\mb{N} \cup \mb{H}$ in $\g[D]_{\overline{\mb{X'}\mb{N'}}\underline{\mb{H}}}$ that is d-connecting given $\mb{X'} \cup \mb{P_i}$. Let $p =\langle B_i, \dots, N \rangle$, $B_i \in \mb{B_i}$, $N \in \mb{N} \cup \mb{H}$, be a shortest such path, and let $p^*$ be the corresponding path in $\g$.

    Consider when $p$ begins with an arrow out of $B_i$. If $p$ were causal, then $N \notin \mb{N'}$ since $p$ is in $\g[D]_{\overline{\mb{X'}\mb{N'}}\underline{\mb{H}}}$. Further, the corresponding path in $\g[D]$ would also be causal, which by Lemma \ref{lem:pc-imply1b} would imply that $p^*$ is a possibly causal path from $\mb{B_i}$ to $\big(\mb{N} \cup \mb{H} \big) \setminus \mb{N'}$. But this contradicts \eqref{eq:algo-4-one-8}. Thus, let $C$ be the closest collider to $B_i$ on $p$. Since $p$ is d-connecting given $\mb{X'} \cup \mb{P_i}$ in $\g[D]_{\overline{\mb{X'}\mb{N'}}\underline{\mb{H}}}$, then $C \in \An(\mb{P_i},\g[D]_{\overline{\mb{X'}\mb{N'}}\underline{\mb{H}}})$. But this implies that there is a causal path in $\g[D]_{\overline{\mb{X'}\mb{N'}}\underline{\mb{H}}}$---and therefore in $\g[D]$---from $\mb{B_i}$ to $\mb{P_i}$. By Lemma \ref{lem:pc-imply1b}, the corresponding path in $\g$ is possibly causal, which again contradicts \eqref{eq:algo-4-one-8}.

    Consider next when $p$ begins with $B_i \gets A$. By choice of $p$, $A \notin \mb{B_i}$ and so $A \in \Pa(\mb{B_i}, \g[D]_{\overline{\mb{X'}\mb{N'}}\underline{\mb{H}}})$. But $A \notin \Pa(\mb{B_i}, \g)$, since $\Pa(\mb{B_i}, \g) \subseteq \mb{X'} \cup \mb{P_i} \cup \mb{H}$ and $p$ is d-connecting given $\mb{X'} \cup \mb{P_i}$ in $\g[D]_{\overline{\mb{X'}\mb{N'}}\underline{\mb{H}}}$. Thus, $p^*$ begins with $B_i - A$. If $p^*$ were undirected, then $N \in \mb{C_n}$ and so $N \notin \mb{H} \subseteq \cup_{j=1}^{n-1} \mb{C_j}$. Thus by Lemma \ref{lem:algo-4abcd}\ref{lem:algo-4b}, $p^*$ would be a possibly causal path from $\mb{B_i}$ to $N \in \mb{N}$ and $(-p^*)$ would be a possibly causal path from $N \in \mb{N}$ to $\mb{B_i}$. The former contradicts \eqref{eq:algo-4-one-8} when $N \in \mb{N} \setminus \mb{N'}$. The latter contradicts Lemma \ref{lem:algo-4efghi}\ref{lem:algo-4e} when $N \in \mb{N'} \subseteq \mb{X}$.

    Since $p^*$ must contain a directed edge, let $T$ be the node closest to $B_i$ on $p^*$ such that $p^*(T, N)$ starts with such an edge. Let $S$ and $U$ be the nodes on $p^*$ immediately preceding and following $T$, respectively. Note that $T \notin \mb{H}$ by choice of $p$. 

    Consider when $p^*$ contains $T \to U$. Note that $p(T,N)$ also contains $T \to U$. When $p(T, N)$ contains at least one collider, then by Lemma \ref{lem:algo-4abcd}\ref{lem:algo-4d}, $\g$ contains a causal path from $T$ to $\mb{P_i}$. When instead $p(T,N)$ is causal, then $N \notin \mb{N'}$ since $p$ is in $\g[D]_{\overline{\mb{X'}\mb{N'}}\underline{\mb{H}}}$. Thus by Lemma \ref{lem:algo-4abcd}\ref{lem:algo-4c}, $\g$ contains a causal path from $T$ to $\big(\mb{N} \cup \mb{H} \big) \setminus \mb{N'}$. In either case, $\g$ contains a causal path from $T$ to $\big(\mb{N} \cup \mb{H} \cup \mb{P_i} \big) \setminus \mb{N'}$. Since $\g$ also contains $p^*(B_i, T)$---which is possibly causal by Lemma \ref{lem:algo-4abcd}\ref{lem:algo-4b} and the fact that $T \notin \mb{H}$---then by Lemma \ref{lem:concat}, $\g$ contains a possibly causal path from $B_i \in \mb{B_i}$ to $\big(\mb{N} \cup \mb{H} \cup \mb{P_i} \big) \setminus \mb{N'}$, which contradicts \eqref{eq:algo-4-one-8}.

    Finally, consider when $p^*$ contains $T \gets U$. Then by R1 of \cite{meek1995causal}, $\g$---and therefore $\g[D]$---contains the edge $\langle S, U \rangle$. Note that $\g[D]$ also contains $T \gets U$. By Lemma \ref{lem:algo-4abcd}\ref{lem:algo-4a}, the path in $\g[D]$ corresponding to $p(B_i, T)$ cannot contain colliders on the path in $\g[D]$ corresponding to $p$. Thus, $\g[D]$ contains $S \gets T \gets U$ and therefore $S \gets U$. Further since $\g[D]_{\overline{\mb{X'}\mb{N'}}\underline{\mb{H}}}$ must also contain $S \gets T \gets U$, then $S \notin \mb{X'} \cup \mb{N'}$ and $U \notin \mb{H}$, and so $\g[D]_{\overline{\mb{X'}\mb{N'}}\underline{\mb{H}}}$ contains $S \gets U$. But then $p(B_i, S) \oplus \langle S,U \rangle \oplus p(U, N)$ contradicts the choice of $p$.

    \vskip .1in

    
    \textbf{Special Cases:}
    With the broader claim shown, we now complete the proof by showing that the independence statements in \eqref{eq:algo-4-one-4}, \eqref{eq:algo-4-one-5}, and \eqref{eq:algo-4-one-6} are special cases of \eqref{eq:algo-4-one-9}.
    
    \eqref{eq:algo-4-one-4}
    Let $\mb{N} = \mb{N_i}$, $\mb{H} = \emptyset$, $\mb{X'} = \mb{X}$, and $\mb{N'} = \emptyset$. Since $\mb{X}$, $\mb{Z}$, and $\mb{D}$ are pairwise disjoint, then $\mb{N}$, $\mb{X'}$, $\mb{B_i}$, and $\mb{P_i}$ are pairwise disjoint by definition. Clearly, $\mb{N} \subseteq \mb{N_i} \cup \mb{X_{n_i}}$ and $\mb{X'} \subseteq \mb{X}$. Then note that $\Pa(\mb{B_i},\g) = \mb{X_{p_i}} \cup \mb{P_i} \subseteq \mb{X'} \cup \mb{P_i} \cup \mb{H}$. Finally, note that $\mb{N} \setminus \mb{N'} = \mb{N_i}$, where $\mb{N_i} \subseteq \cup_{j=1}^{i-1} \mb{B_j} \cup \mb{Z} \subseteq \cup_{j=1}^{n-1} \mb{C_j} \cup \mb{Z}$ by definition.

    \vskip .1in
    
    \eqref{eq:algo-4-one-5}
    Let $\mb{N} = \mb{X_{n_i}}$, $\mb{H} = \emptyset$, $\mb{X'} = \mb{X_{p_i}}$, and $\mb{N'} = \mb{X'_{n_i}}$. Since $\mb{X}$, $\mb{Z}$, and $\mb{D}$ are pairwise disjoint, then $\mb{N}$, $\mb{X'}$, $\mb{B_i}$, and $\mb{P_i}$ are pairwise disjoint by definition. Clearly, $\mb{N} \subseteq \mb{N_i} \cup \mb{X_{n_i}}$ and $\mb{X'} \subseteq \mb{X}$. Further, $\mb{N'} = \mb{X_{n_i}} \setminus \An(\mb{P_i}, \g[D]_{\overline{\mb{X_{p_i}}}}) \subseteq \mb{X_{n_i}} \setminus \An(\mb{P_i}, \g[D]_{\overline{\mb{X'}}})$. Then note that $\Pa(\mb{B_i},\g) = \mb{X_{p_i}} \cup \mb{P_i} \subseteq \mb{X'} \cup \mb{P_i} \cup \mb{H}$. 
    
    Finally, to see that $\mb{N} \setminus \mb{N'} \subseteq \cup_{j=1}^{n-1} \mb{C_j}$, let $N$ be an arbitrary node in $\mb{N} \setminus \mb{N'} = \mb{X_{n_i}} \cap \An(\mb{P_i}, \g[D]_{\overline{\mb{X_{p_i}}}})$. This implies that there is a causal path in $\g[D]_{\overline{\mb{X_{p_i}}}}$ from $N$ to some $P_i \in \mb{P_i} \subseteq \Pa(\mb{B_i},\g)$. This path is also in $\g[D]$, and thus by Lemma \ref{lem:pc-imply1b}, its corresponding path in $\g$ is possibly causal from $N$ to $\Pa(\mb{B_i},\g)$. Therefore by Lemma \ref{lem:concat}, there is a possibly causal path in $\g$ from $N$ to $\mb{B_i}$. Since $\mb{B_i} \subseteq \mb{C_n}$, then $N \notin \cup_{j=n+1}^{k} \mb{C_j}$ by Lemma \ref{lem:bucket-restrict2}. Further, since $N \in \mb{X}$ and $\mb{B_i} \subseteq \mb{D}$, then $N \notin \mb{C_n}$ by Lemma \ref{lem:algo-4efghi}\ref{lem:algo-4e}.

    \vskip .1in
    
    \eqref{eq:algo-4-one-6}
    Let $\mb{N} = \emptyset$, $\mb{H} = \mb{X_{p_i}}$, $\mb{X'} = \emptyset$, and $\mb{N'} = \emptyset$. Since $\mb{X}$, $\mb{Z}$, and $\mb{D}$ are pairwise disjoint, then $\mb{H}$, $\mb{B_i}$, and $\mb{P_i}$ are pairwise disjoint by definition. Clearly, $\mb{H} \subseteq \mb{X_{p_i}}$. Then note that $\Pa(\mb{B_i},\g) = \mb{X_{p_i}} \cup \mb{P_i} \subseteq \mb{X'} \cup \mb{P_i} \cup \mb{H}$. Finally, note that $\mb{N} \setminus \mb{N'} = \emptyset \subseteq \cup_{j=1}^{n-1} \mb{C_j} \cup \mb{Z}$.

    \vskip .2in
    

    \textbf{PART \ref{lem:algo-4-two}}
    Let $\mb{B_i^D}= \cup_{j=0}^{i-1}\mb{B_j} \cap \PossDe(\mb{X},\g)$. We show the result in the steps below.
    \begin{align}
        f(\mb{b_i}| \mb{b_{i-1}}, \dots, \mb{b_0}, do(\mb{x}), \mb{z})
                &= f(\mb{b_i}| \mb{b_i^N}, do(\mb{x}),\mb{z}) \label{eq:algo-4-two-1}\\
                &= f(\mb{b_i}| \mb{b_i^N}, \mb{z}). \label{eq:algo-4-two-2}
    \end{align}
    In order to use Rules 1 and 3 of the do calculus (Theorem \ref{thm:do-calc}) to show the equalities in \eqref{eq:algo-4-two-1} and \eqref{eq:algo-4-two-2}, respectively, we must show the following.
    \begin{align}
        &(\mb{B_i} \dsepp \mb{B_i^D} | \mb{X}, \mb{Z}, \mb{B_i^N})_{\g[D]_{\overline{\mb{X}}}}. \label{eq:algo-4-two-3}\\
        &(\mb{B_i} \dsepp \mb{X} \,\,\,\,| \mb{B_i^N}, \mb{Z})_{\g[D]_{\overline{\mb{W}}}}, \label{eq:algo-4-two-4}
    \end{align}
    where we let $\mb{W} = \mb{X} \setminus \An(\mb{B_i^N} \cup \mb{Z}, \g[D])$. We prove these independencies below.

    \vskip .1in
    
    \eqref{eq:algo-4-two-3} 
    Suppose for sake of contradiction that there is a path in $\g[D]_{\overline{\mb{X}}}$ from $\mb{B_i}$ to $\mb{B_i^D}$ that is d-connecting given $\mb{X} \cup \mb{Z} \cup \mb{B_i^N}$. Let $q=\langle B_i=Q_0, \dots, Q_m=B_i^D \rangle$, $m \ge 1$, be a shortest such path and let $q^*$ be the corresponding path in $\g$. Further, let $\mb{C_h}$, $h \in \{1, \dots, n-1\}$, be the bucket in $\mb{V}$ such that $B_i^D \in \mb{C_h}$. 

    Before beginning the proof, note the following.
    \begin{align*}
        \text{There is no ca}&\text{usal path in } \g[D]_{\overline{\mb{X}}} \text{ from } \label{eq:algo-4-one-10} \tag{$\ast\ast$}\\
        \mb{C_h} \text{ to } &\mb{B_i} \cup \mb{Z} \cup \mb{B_i^N}. \nonumber
    \end{align*}    
    To see this, suppose for sake of contradiction that such a path $p$ did exist, and let $p^*$ be the corresponding path in $\g$. Since the corresponding path in $\g[D]$ is also causal, then by Lemma \ref{lem:pc-imply1b}, $p^*$ is a possibly causal path from $\mb{C_h}$ to $\mb{B_i} \cup \mb{Z} \cup \mb{B_i^N}$, where by Lemma \ref{lem:algo-4efghi}\ref{lem:algo-4e}, $\mb{C_h} \subseteq \De(\mb{X}, \g)$. Thus by Lemma \ref{lem:concat}, $\g$ contains a possibly causal path from $\mb{X}$ to $\mb{B_i} \cup \mb{Z} \cup \mb{B_i^N}$. But by assumption, definition, and Lemma \ref{lem:algo-4efghi}\ref{lem:algo-4f}, $(\mb{B_i} \cup \mb{Z} \cup \mb{B_i^N}) \cap \PossDe(\mb{X}, \g) = \emptyset$. With the claim shown, we turn to find a contradiction in the two cases below.
    
    \textbf{Case 1:} 
    Consider when $q$ contains $Q_{m-1} \gets B_i^D$. Note that $(-q)$ cannot be causal from $B_i^D \in \mb{C_h}$ to $B_i \in \mb{B_i}$ by \eqref{eq:algo-4-one-10} and therefore, must contain a collider. Let $\mb{L}$ be the set containing the collider closest to $B_i^D$ on $q$ and all of its descendants in $\g[D]_{\overline{\mb{X}}}$. Since $q$ is d-connecting given $\mb{X} \cup \mb{Z} \cup \mb{B_i^N}$ in $\g[D]_{\overline{\mb{X}}}$, then $\mb{L}$ must contain an element in $\mb{Z} \cup \mb{B_i^N}$. But since $\mb{L} \subseteq \De(\mb{C_h},\g[D]_{\overline{\mb{X}}})$, this contradicts \eqref{eq:algo-4-one-10}.

    \textbf{Case 2:}
    Consider instead when $q$ contains $Q_{m-1} \to B_i^D$. Note that $q$ is not causal from $B_i \in \mb{C_n}$ to $B_i^D \in \mb{C_h}$, since otherwise the corresponding path in $\g[D]$ would also be causal, which would contradict Corollary \ref{cor:bucket-restrict2b}. Thus, $q$ contains a node $Q_c$, $c \in \{1, \dots, m-1\}$, such that $q$ contains $Q_{c-1} \gets Q_c \to \dots \to B_i^D$. Let $\mb{C_e}$, $e \in \{1, \dots, n-1\}$, be the bucket in $\mb{V}$ such that $Q_c \in \mb{C_e}$.

    For sake of contradiction, suppose $e \neq h$. That is, $Q_c$ and $B_i^D$ are in the distinct buckets $\mb{C_e}$ and $\mb{C_h}$, respectively. Define $Q_d$, where $d \in \{1, \dots, m-1\}$, $d \ge c$, to be the last node on $q$ in $\mb{C_e}$. Note that since $q(Q_d, B_i^D)$ is causal, the corresponding path in $\g[D]$ is also causal. Therefore, by Lemma \ref{lem:pc-imply1b}, $q^*(Q_d, B_i^D)$ is a proper possibly causal path from $\mb{C_e}$ to $\mb{C_h}$ and so by Lemma \ref{lem:bucket-req3}, $\g$ contains at least one subsequence of $q^*(Q_d, B_i^D)$ that is causal from $Q_d$ to $B_i^D$. Let $r^*$ be an arbitrary such path.

    Note that since $q$ is d-connecting given $\mb{X} \cup \mb{Z} \cup \mb{B_i^N}$, no node on $q(Q_d, B_i^D)$---and therefore no node on $r^*$---is in $\mb{X}$. That is, $r^*$ is a causal path in $\g$ from $Q_d$ to $B_i^D \in \An(\mb{Y}, \g_{\mb{V} \setminus \mb{X}}) \setminus \mb{Z}$ that does not contain any nodes in $\mb{X}$. Thus, $Q_d \in \mb{Z} \cup \mb{D}$. By the same logic, $Q_d \notin \mb{Z} \cup \mb{B_i^N}$, and further, by the choice of $q$ as a shortest path, $Q_d \notin \mb{B_i^D}$. But this implies that $Q_d \in \cup_{j=i}^{k} \mb{B_j}$, which contradicts Corollary \ref{cor:bucket-restrict2a}.

    Note that we have shown $q$ contains $Q_{c-1} \gets Q_c$, where $Q_c \in \mb{C_h}$. We can derive our final contradictions by using identical logic to that of Case 1 above.
        
    \vskip .2 in
    

    \eqref{eq:algo-4-two-4}
    Note that by Lemma \ref{lem:pc-imply2}, $\De(\mb{X}, \g[D]) \subseteq \PossDe(\mb{X}, \g)$. From this and since $(\mb{B_i^N} \cup \mb{Z}) \cap \PossDe(\mb{X}, \g) = \emptyset$, it follows that $(\mb{B_i^N} \cup \mb{Z}) \cap \De(\mb{X}, \g[D]) = \emptyset$ and thus $\mb{W} = \mb{X}$. Therefore, we want to show $(\mb{B_i} \dsepp \mb{X}| \mb{B_i^N}, \mb{Z})_{\g[D]_{\overline{\mb{X}}}}$.
    
    Let $p$ be an arbitrary path in $\g[D]_{\overline{\mb{X}}}$ from $X \in \mb{X}$ to $B_i \in \mb{B_i}$. By the definition of $\g[D]_{\overline{\mb{X}}}$, $p$ begins with $X \to$. But note that $p$ cannot be causal, since this would imply that $B_i \in \De(\mb{X}, \g[D]) \subseteq \PossDe(\mb{X}, \g)$, which would contradict Lemma \ref{lem:algo-4efghi}\ref{lem:algo-4f}. Thus, $p$ must have at least one collider.
    
    Let $\mb{S}$ be the set containing the collider that is closest to $X$ on $p$ and all of its descendants in $\g[D]_{\overline{\mb{X}}}$. Since $\mb{S} \subseteq \De(X, \g[D]_{\overline{\mb{X}}}) \subseteq \De(\mb{X}, \g[D]) \subseteq \PossDe(\mb{X}, \g)$ and since $(\mb{B_i^N} \cup \mb{Z}) \cap \PossDe(\mb{X}, \g) = \emptyset$, then $p$ must be d-separated given $\mb{B_i^N} \cup \mb{Z}$.
\end{proofof}


\subsection{Supporting Results}


\begin{lemma}[\textbf{Setup for Proof of Lemma \ref{lem:algo-4}\ref{lem:algo-4-one}}]
\label{lem:algo-4abcd}
    Let $\mb{N}$, $\mb{H}$, $\mb{X'}$, $\mb{B}$, and $\mb{P}$ be pairwise disjoint node sets in an MPDAG $\g =(\mb{V,E})$, and let $\g[D]$ be a DAG in $[\g]$. Suppose there is a path from $\mb{B}$ to $\mb{N} \cup \mb{H}$ that is d-connecting given $\mb{X'} \cup \mb{P}$ in $\g[D]_{\overline{\mb{X'}\mb{N'}}\underline{\mb{H}}}$, where $\mb{N'}$ is a node set in $\g$ such that $\mb{N'} \subseteq \mb{N}$ and $\mb{N'} \cap \An(\mb{P},\g[D]_{\overline{\mb{X'}}}) = \emptyset$. Let $p_1 =\langle B, \dots, N \rangle$, $B \in \mb{B}$, $N \in \mb{N} \cup \mb{H}$, be a shortest such path, and let $p^*$ and $p_2$ be the corresponding paths in $\g$ and $\g[D]$, respectively. Finally, let $q_1 = \langle Q_0, \dots, Q_k \rangle$, $k \ge 1$, be a subpath of $p_1$, and let $q^*$ and $q_2$ be the corresponding paths in $\g$ and $\g[D]$, respectively.
    \begin{enumerate}[label=(\roman*)]
        \item \label{lem:algo-4a} 
        If $q^*$ is undirected, then $q_2$ does not contain a collider on $p_2$.

        \item \label{lem:algo-4b} 
        If $q^*$ is undirected and $Q_k \notin \mb{H}$, then $q^*$ and $(-q^*)$ are possibly causal.

        \item \label{lem:algo-4c} 
        If $q_1$ is causal and $q^*$ begins with $Q_0 \to Q_1$, then $q^*$ is causal.

        \item \label{lem:algo-4d} 
        If $q_1$ contains at least one collider and $q^*$ begins with $Q_0 \to Q_1$, then $\g$ contains a causal path from $Q_0$ to $\mb{P}$.
    \end{enumerate}
\end{lemma}

\begin{proofof}[Lemma \ref{lem:algo-4abcd}]
    Let $Q_{-1}$ designate the node immediately preceding $Q_0$ on $p_1$ when $Q_0 \neq B$, and let $Q_{k+1}$ designate the node immediately following $Q_k$ on $p_1$ when $Q_k \neq N$.

    \vskip .1in


    \ref{lem:algo-4a}
    Let $q^*$ be undirected. For sake of contradiction, suppose that $q_2$ contains a collider on $p_2$ at some node $Q_j$, $0 \le j \le k$. But since $q^*$ is undirected in $\g$, then $Q_j$ is not a collider on $p^*$. Thus by properties of MPDAGs, $\g$---and therefore $\g[D]$---must contain an edge $\langle Q_{j-1}, Q_{j+1} \rangle$. We show below that this edge also exists in $\g[D]_{\overline{\mb{X'}\mb{N'}}\underline{\mb{H}}}$.
    
    Note that $Q_j$ is also a collider on $p_1$. Since $p_1$ is d-connecting given $\mb{X'} \cup \mb{P}$, then $Q_j$ and therefore $Q_{j-1}$ and $Q_{j+1}$ are in $\An(\mb{P}, \g[D]_{\overline{\mb{X'}\mb{N'}}\underline{\mb{H}}}) \subseteq \An(\mb{P}, \g[D]_{\overline{\mb{X'}}})$. So since $\mb{N'} \cap \An(\mb{P}, \g[D]_{\overline{\mb{X'}}}) = \emptyset$, then $Q_{j-1}, Q_{j+1} \notin \mb{N'}$. Further, since $Q_{j-1}$ and $Q_{j+1}$ are not colliders on $p_1$, they cannot be nodes in $\mb{X'}$ (or $\mb{P}$). Finally, since $\g[D]_{\overline{\mb{X'}\mb{N'}}\underline{\mb{H}}}$ contains $Q_{j-1} \to Q_j \gets Q_{j+1}$, then $Q_{j-1}, Q_{j+1} \notin \mb{H}$. Therefore $\g[D]_{\overline{\mb{X'}\mb{N'}}\underline{\mb{H}}}$ must also contain an edge $\langle Q_{j-1}, Q_{j+1} \rangle$. 

    Define the path $m_1 = p_1(B, Q_{j-1}) \oplus \langle Q_{j-1}, Q_{j+1} \rangle \oplus p_1(Q_{j+1}, N)$ in $\g[D]_{\overline{\mb{X'}\mb{N'}}\underline{\mb{H}}}$. By the choice of $p_1$, $m_1$ must be blocked given $\mb{X'} \cup \mb{P}$. Since $Q_{j-1}, Q_{j+1} \notin \mb{X'} \cup \mb{P}$, then $m_1$ must be blocked given $\mb{X'} \cup \mb{P}$ by a collider at $Q_{j-1}$ or $Q_{j+1}$. But this contradicts that $Q_{j-1}, Q_{j+1} \in \An(\mb{P}, \g[D]_{\overline{\mb{X'}\mb{N'}}\underline{\mb{H}}})$.

    \vskip .1in
    
    
    \ref{lem:algo-4b}
    Let $q^*$ be undirected and $Q_k \notin \mb{H}$. We show that $q^*$ is possibly causal. The proof for $(-q^*)$ follows analogous logic. Note that by part \ref{lem:algo-4a}, $q_2$ does not contain a collider on $p_2$, and therefore, $q_1$ does not contain a collider on $p_1$.
    
    For sake of contradiction, suppose that $q^*$ is non-causal. Then there exists an edge $Q_g \gets Q_h$ in $\g$---and therefore in $\g[D]$---for some $g, h \in \{0, \dots, k\}$, $h>g+1$. To see that $\g[D]_{\overline{\mb{X'}\mb{N'}}\underline{\mb{H}}}$ also contains $Q_g \gets Q_h$, note that $Q_h \notin \mb{H}$ by assumption and choice of $p_1$. Additionally, $Q_g \notin \mb{X'} \cup \mb{N'}$, since $q_2$ does not contain a collider on $p_2$ and $\g[D]$ does not contain a cycle, which implies that $\g[D]$---and therefore $\g[D]_{\overline{\mb{X'}\mb{N'}}\underline{\mb{H}}}$-- must contain $Q_g \gets Q_{g+1}$.

    Define a path $u_1 = p_1(B, Q_g) \oplus \langle Q_g, Q_h \rangle \oplus p_1(Q_h, N)$ in $\g[D]_{\overline{\mb{X'}\mb{N'}}\underline{\mb{H}}}$. By the choice of $p_1$, $u_1$ must be blocked given $\mb{X'} \cup \mb{P}$ at $Q_g$ or $Q_h$. However, note that since $q_1$ cannot contain colliders on $p_1$, then $Q_g$ and $Q_h$ are non-colliders on $p_1$. Further, since $\g[D]_{\overline{\mb{X'}\mb{N'}}\underline{\mb{H}}}$ contains $Q_g \gets Q_h$ and $Q_g \gets Q_{g+1}$, then $Q_g$ and $Q_h$ are also non-colliders on $u_1$. So if $u_1$ is blocked given $\mb{X'} \cup \mb{P}$ at $Q_g$ or $Q_h$, then so is $p_1$, which is a contradiction.
    
    \vskip .1in
    
    
    \ref{lem:algo-4c} 
    Let $q_1$ be causal and let $q^*$ begin with $Q_0 \to Q_1$. Note that $q_2$ is also causal. Suppose for sake of contradiction that $q^*$ is shielded. That is, $\g$---and therefore $\g[D]$---contains at least one edge $\langle Q_d, Q_{d+2} \rangle$ for some $d \in \{0, \dots, k-2 \}$. Since $q_2$ is causal, $\g[D]$ must contain $Q_d \to Q_{d+2}$. Further, because $\g[D]_{\overline{\mb{X'}\mb{N'}}\underline{\mb{H}}}$ contains $Q_d \to Q_{d+1} \to Q_{d+2}$, then $Q_{d+2} \notin \mb{X'} \cup \mb{N'}$ and $Q_d \notin \mb{H}$, and so $\g[D]_{\overline{\mb{X'}\mb{N'}}\underline{\mb{H}}}$ must also contain $Q_d \to Q_{d+2}$. But then the path $p_1(B,Q_d) \oplus \langle Q_d, Q_{d+2} \rangle \oplus p_1(Q_{d+2},N)$ contradicts the choice of $p_1$. Thus, $q^*$ is unshielded. 

    Since $q_2$ is causal, $q^*$ can only contain edges of the form $Q_i - Q_{i+1}$ and $Q_i \to Q_{i+1}$ for all $i \in \{0, \dots, k-1\}$. Since additionally $q^*$ begins with $Q_0 \to Q_1$ and is unshielded, the result follows by R1 of \cite{meek1995causal}.
    
    \vskip .1in


    \ref{lem:algo-4d}
    Let $q_1$ contain at least one collider and $q^*$ begin with $Q_0 \to Q_1$. Note that $q_1$ also begins with $Q_0 \to Q_1$. Then let $T_0$ be the closest collider to $Q_0$ on $q_1$. Note that $T_0$ is also a collider on $p_1$. Since $p_1$ is d-connecting given $\mb{X'} \cup \mb{P}$ in $\g[D]_{\overline{\mb{X'}\mb{N'}}\underline{\mb{H}}}$, then $T_0 \in \An(\mb{P}, \g[D]_{\overline{\mb{X'}\mb{N'}}\underline{\mb{H}}})$. Thus let $t_1= \langle T_0, \dots, T_w \rangle$, $w \ge 0$, $T_w \in \mb{P}$, be a shortest causal path in $\g[D]_{\overline{\mb{X'}\mb{N'}}\underline{\mb{H}}}$ from $T_0$ to $\mb{P}$, and let $t^*$ and $t_2$ be the corresponding paths in $\g$ and $\g[D]$, respectively.
    
    Note that $q_1(Q_0, T_0)$ is causal, and therefore, by part \ref{lem:algo-4c}, the subpath $q^*(Q_0, T_0)$ is causal. We want to show that there is a causal path from $Q_0$ to $T_w$ in $\g$. If $t^*$ is causal, then $q^*(Q_0, T_0) \oplus t^*$ is such a path and we are done.

    Consider when $t^*$ is not causal, and focus on the nodes adjacent to $T_0$ on $q_1$. Let $S$ be the node immediately preceding and $U$ the node immediately following $T_0$ on $q_1$. Note that since $S$ and $U$ are non-colliders on $p_1$ and $p_1$ is d-connecting given $\mb{X'} \cup \mb{P}$, then $S, U \notin \mb{X'} \cup \mb{P}$. Also since $\g[D]_{\overline{\mb{X'}\mb{N'}}\underline{\mb{H}}}$ contains $S \to T_0 \gets U$, then $S,U \notin \mb{H}$. Further since $T_0 \in \An(\mb{P}, \g[D]_{\overline{\mb{X'}\mb{N'}}\underline{\mb{H}}})$, then $S,U \in \An(\mb{P}, \g[D]_{\overline{\mb{X'}\mb{N'}}\underline{\mb{H}}}) \subseteq \An(\mb{P}, \g[D]_{\overline{\mb{X'}}})$. Therefore, $S,U \notin \mb{N'}$. Putting this together, we have that $S,U \notin \mb{X'} \cup \mb{P} \cup \mb{H} \cup \mb{N'}$.

    For sake of contradiction, suppose $\langle S, T_0, U \rangle$ is shielded in $\g$---and therefore in $\g[D]$. Since $S,U \notin \mb{X'} \cup \mb{N'} \cup \mb{H}$, then $\g[D]_{\overline{\mb{X'}\mb{N'}}\underline{\mb{H}}}$ also contains $\langle S,U \rangle$. Define a path $m_1 = p_1(B, S) \oplus \langle S, U \rangle \oplus p(U, N)$ in $\g[D]_{\overline{\mb{X'}\mb{N'}}\underline{\mb{H}}}$. By the choice of $p_1$, $m_1$ must be blocked given $\mb{X'} \cup \mb{P}$. Since $S, U \notin \mb{X'} \cup \mb{P}$, then either $S$ or $U$ must be a collider on $m_1$ that is not in $\An(\mb{P}, \g[D]_{\overline{\mb{X'}\mb{N'}}\underline{\mb{H}}})$, which is a contradiction. Thus, $\langle S, T_0, U \rangle$ is unshielded in $\g$. Since $\g[D]$ must also contain $S \to T_0 \gets U$, then by properties of MPDAGs, so does $\g$.

    Turn to consider $t^*$. Suppose for sake of contradiction that $\g$---and therefore $\g[D]$---contains at least one edge $\langle T_f, T_{f+2} \rangle$ for some $f \in \{0, \dots, w-2 \}$. Note that $t_2$ is causal, and so $\g[D]$ must contain $T_f \to T_{f+2}$. Since $\g[D]_{\overline{\mb{X'}\mb{N'}}\underline{\mb{H}}}$ contains an edge into and out of each node on $t_1$ except for $T_w \in \mb{P}$, then no node on $t_1$ is in $\mb{X'} \cup \mb{N'} \cup \mb{H}$ and so $\g[D]_{\overline{\mb{X'}\mb{N'}}\underline{\mb{H}}}$ must also contain $T_f \to T_{f+2}$. But then the path $t_1(T_0, T_f) \oplus \langle T_f, T_{f+2} \rangle \oplus t_1(T_{f+2}, T_w)$ contradicts the choice of $t_1$, and thus, $t^*$ is unshielded. 

    Next, define a node $T_v, v \in \{0, \dots, w\}$, as follows. When $t^*$ is entirely undirected, let $T_v = T_w$. Otherwise, let $T_v$ be the first node on $t^*$ followed by a directed arrow. Note that since $t_2$ is causal, $t^*$ can only contain edges of the form $T_i - T_{i+1}$ or $T_i \to T_{i+1}$ for $i \in \{0, \dots, w-1\}$. Thus, when $T_v \neq T_w$, $t^*$ contains $T_v \to T_{v+1}$. Then since $t^*$ is unshielded, by R1 of \cite{meek1995causal}, $t^*$ takes the form $T_0 - \dots - T_v \to \dots \to T_w$.

    We now show that $\g$ contains the edge $S \to T_v$, which will make $q^*(Q_0,S) \oplus \langle S, T_v\rangle \oplus t^*(T_v,T_w)$ a causal path in $\g$ from $Q_0$ to $\mb{P}$. We have already shown that $\g$ and $\g[D]$ contain $S \to T_0 \gets U$. For sake of induction, suppose that $\g$---and therefore $\g[D]$---contains $S \to T_d \gets U$ for some $d \in \{0, \dots, v-1\}$. Since $\g$ contains the paths $S \to T_d - T_{d+1}$ and $U \to T_d - T_{d+1}$, then $\g$---and therefore $\g[D]$---must contain the edges $\langle S, T_{d+1} \rangle$ and $\langle U,T_{d+1} \rangle$ by R1 of \cite{meek1995causal}. Since $\g[D]$ contains $S \to T_0 \to \dots \to T_{d+1}$ and $U \to T_0 \to \dots \to T_{d+1}$, it must also contain $S \to T_{d+1} \gets U$. Then because $\langle S, T_0, U \rangle$ is unshielded in $\g$, so is $\langle S, T_{d+1}, U \rangle$. Thus, by properties of MPDAGs, $\g$ must also contain $S \to T_{d+1} \gets U$.
\end{proofof}


\begin{lemma}[\textbf{Setup for Proof of Lemma \ref{lem:algo-4}}]
\label{lem:algo-4efghi} 
    Let $\mb{X}$, $\mb{Y}$, and $\mb{Z}$ be pairwise disjoint node sets in an MPDAG $\g =(\mb{V,E})$, where $\mb{Z} \cap \PossDe(\mb{X},\g) = \emptyset$ and where there is no proper possibly causal path from $\mb{X}$ to $\mb{Y}$ that starts with an undirected edge in $\g$. Further, let $\mb{D} =\An(\mb{Y},\g_{\mb{V} \setminus \mb{X}}) \setminus \mb{Z}$.
    \begin{enumerate}[label=(\roman*)] 
        \item \label{lem:algo-4e}
        Let $X \in \mb{X}$, $D \in \mb{D} \cap \PossDe(X, \g)$, and $\mb{C_d}$ be the bucket in $\mb{V}$ such that $D \in \mb{C_d}$. Then $\mb{C_d} \subseteq \De(X, \g)$.
    
        \item \label{lem:algo-4f}
        Let $(\mb{B_1}, \dots, \mb{B_k}) = \PCO(\mb{D},\g)$, $k \ge 1$, and let $i \in \{1, \dots, k\}$ such that $\mb{Z} \cap \PossDe(\mb{B_i}, \g) \neq \emptyset$. Then $\mb{B_i} \cap \PossDe(\mb{X}, \g) = \emptyset$.
    \end{enumerate}
\end{lemma}

\begin{proofof}[Lemma \ref{lem:algo-4efghi}]
    
    \ref{lem:algo-4e}
    We start by showing that $D \in \De(X, \g)$. To see this, pick an arbitrary possibly causal path in $\g$ from $X$ to $D$, and let $p = \langle X=P_0, \dots, P_r=D \rangle, r \ge 1$, be a shortest subsequence in $\g$ of that path. By Lemma \ref{lem:shortest-subseq}, $p$ is a possibly causal, unshielded path from $X$ to $D$. Since $\mb{D} \subseteq \An(\mb{Y}, \g_{\mb{V} \setminus \mb{X}})$, then by Lemma \ref{lem:xtod}, there is no proper possibly causal path from $X$ to $D$ that starts with an undirected edge in $\g$, and so $p$ must contain $P_0 \to P_1$. Then by R1 of \cite{meek1995causal}, $p$ takes the form $P_0 \to \dots \to P_r$. 

    Now suppose for sake of contradiction that $X \in \mb{C_d}$. Let $X'$ be the last node on $p$ in $\mb{X}$, and let $H$ be the node on $p$ immediately following $X'$. That is, $\g$ contains $X \to \dots \to X' \to H \to \dots \to D$, where it may be that $X'=X$ and $H=D$. Note that by Lemma \ref{lem:bucket-restrict1}, $p$ is entirely contained in $\mb{C_d}$ and so $X',H \in \mb{C_d}$. 

    By the definition of $\mb{C_d}$ as a bucket in $\mb{V}$, $\g$ must contain an undirected path from $X'$ to $H$. Let $u= \langle X' = U_0, \dots, U_r=H \rangle, r \ge 2$ be a shortest such path. We show by induction that $\g$ contains $X' \to U_c$ for all $c \in \{2, \dots, r \}$. We already have that $\g$ contains $X' \to U_r$. So suppose $\g$ contains $X' \to U_{d+1}$ for some $d \in \{2, \dots, r-1 \}$. Since $\g$ contains $X' \to U_{d+1} - U_d$, $\g$ must contain an edge $\langle X',U_d \rangle$. By the choice of $u$, $\g$ cannot contain $X' - U_d$. Neither can $\g$ contain $X' \gets U_d$, since by R2 of \cite{meek1995causal} this would force $u$ to contain $U_d \to U_{d+1}$. Therefore $\g$ contains $X' \to U_d$. 
        
    Turn to note that, by definition of $H$, $p(H,D)$ is a causal path with no nodes in $\mb{X}$, where $D \in \An(\mb{Y}, \g_{ \mb{V} \setminus \mb{X} }) \setminus \mb{Z}$. Thus, $\g$ contains a causal path from $H$ to $\mb{Y}$ with no nodes in $\mb{X}$. Since $\g$ contains $X' \to H$ and $\mb{Z} \cap \PossDe(\mb{X}, \g) = \emptyset$, then $H \notin \mb{Z}$. Thus, $H \in \mb{D}$, and by Lemma \ref{lem:xtod}, there is no proper possibly causal path from $\mb{X}$ to $H$ in $\g$ that starts with an undirected edge. Note that $u$ may not be proper, but there is a subpath of $u$ that is a proper path from $\mb{X}$ to $H$, which therefore cannot be possibly causal. Thus, $\g$ must contain $U_a \gets U_b$ for some $a, b \in \{0, \dots, r\}$, $a+1<b$. Pick the earliest such $U_a$ on $u$. 
    
    Suppose for sake of contradiction that $U_a \neq X'$. Then $\g$ contains $U_{a-1} - U_a \gets U_b$ and so R1 of \cite{meek1995causal}, $\g$ must contain an edge $\langle U_{a-1}, U_b \rangle$. $\g$ cannot contain $U_{a-1} - U_b$ since this would contradict the choice of $u$. Neither can $\g$ contain $U_{a-1} \to U_b$ since by R2 of \cite{meek1995causal}, this would force $u$ to contain the directed edge $U_{a-1} \to U_a$. Thus $\g$ contains $U_{a-1} \gets U_b$, but this contradicts the choice of $U_a$. Therefore $\g$ contains $X' \gets U_b$, $b \in \{2, \dots, r\}$, but this contradicts that $\g$ contains $X' \to U_c$ for all $c \in \{2, \dots, r\}$.

    We have shown that $X \notin \mb{C_d}$, where $\g$ contains a causal path from $X$ to $D \in \mb{C_d}$. The result follows from Lemma \ref{lem:bucket-req1}.
    
    \vskip .1in

    \ref{lem:algo-4f}    
    Suppose for sake of contradiction that $\mb{B_i} \cap \PossDe(\mb{X}, \g) \neq \emptyset$. By part \ref{lem:algo-4e}, there is a causal path in $\g$ from $\mb{X}$ to every node in $\mb{B_i}$. Since $\mb{Z} \cap \PossDe(\mb{B_i},\g) \neq \emptyset$, then there is also a possibly causal path in $\g$ from some $B_i \in \mb{B_i}$ to $\mb{Z}$. Thus by Lemma \ref{lem:concat}, there is a possibly causal path in $\g$ from $\mb{X}$ to $\mb{Z}$, which contradicts that $\mb{Z} \cap \PossDe(\mb{X}, \g) = \emptyset$.
\end{proofof}


\section{Proofs for Section \ref{sec:id-condition}: Identifiability Condition}
\label{app:id-condition}

This section includes the proof of Proposition \ref{prop:id-condition} found in Section \ref{sec:id-condition}. We note that one direction of the proof ($\Leftarrow$) relies heavily on a stronger claim that does not require $\mb{Z} \cap \PossDe(\mb{X},\g) = \emptyset$ (see Proposition \ref{prop:id-necessary}). But we do not include this claim in the main text, since its converse does not hold (see Examples \ref{ex:alg-condition1}-\ref{ex:alg-condition2}). The statement and proof of this claim are in the main results below. Two supporting results follow.


\subsection{Main Results}

\begin{proofof}[Proposition \ref{prop:id-condition}] 
    Follows from Theorem \ref{thm:id-formula} and Proposition \ref{prop:id-necessary}.
\end{proofof}

\begin{proposition}[Necessity Condition]
\label{prop:id-necessary}
    Let $\mb{X}$, $\mb{Y}$, and $\mb{Z}$ be pairwise disjoint node sets in a causal MPDAG $\g$. If the conditional causal effect of $\mb{X}$ on $\mb{Y}$ given $\mb{Z}$ is identifiable in $\g$, then there is no proper possibly causal path from $\mb{X}$ to $\mb{Y}$ in $\g$ that starts with an undirected edge and does not contain nodes in $\mb{Z}$.
\end{proposition}

\begin{proofof}[Proposition \ref{prop:id-necessary}] 
    We prove the contrapositive using similar logic to the proof of Proposition 3.2 in \cite{perkovic2020identifying}. Thus, suppose there is a proper possibly causal path from $\mb{X}$ to $\mb{Y}$ in $\g = (\mb{V},\mb{E})$ that starts with an undirected edge and does not contain nodes in $\mb{Z}$. Then by Lemma \ref{lem:twopaths-corrollary}, there is one such path---call it $q=\langle X=V_0, \dots, V_k=Y\rangle$, $X \in \mb{X}$, $Y \in \mb{Y}, k \ge 1$---where the corresponding paths in two DAGs in $[\g]$ take the forms $X \to \dots \to Y$ and $X \gets V_1 \to \dots \to Y$ ($X \gets Y$ when $k=1$). Call these DAGs $\g[D]^1$ and $\g[D]^2$ with paths $q_1$ and $q_2$, respectively.

    To prove that the conditional causal effect of $\mb{X}$ on $\mb{Y}$ given $\mb{Z}$ is not identifiable in $\g$, it suffices to show that there are two families of interventional densities over $\mb{V}$---call them $\mb{F^*_1}$ and $\mb{F^*_2}$, where for $i \in \{1,2\}$, we define $\mb{F^*_i} = \{ f_i(\mb{v}|do(\mb{x'})) : \mb{X'} \subseteq \mb{V} \}$---such that the following properties hold. 
    \begin{enumerate}[label=(\roman*)]
        \item $\g[D]^1$ and $\g[D]^2$ are compatible with $\mb{F^*_1}$ and $\mb{F^*_2}$, respectively.
            \footnote{For brevity, we say a DAG is ``compatible with'' a set of interventional densities and an interventional density is ``consistent with'' a DAG as shorthand for these claims holding only were the DAG to be causal.}
            \label{prop:id-necessary-1}
        \item $f_1(\mb{v}) = f_2(\mb{v})$. \label{prop:id-necessary-2}
        \item $f_1(\mb{y} | do(\mb{x}), \mb{z}) \neq f_2(\mb{y} | do(\mb{x}), \mb{z})$. \label{prop:id-necessary-3}
    \end{enumerate}

    To define such families, we start by introducing an additional DAG and an observational density $f(\mb{v})$. That is, let $\g[D]^{1'}$ be a DAG constructed by removing every edge from $\g[D]^1$ except for the edges on $q_1$. Then let $f(\mb{v})$ be the multivariate normal density under the following linear structural equation model (SEM). Each random variable $A \in \mb{V}$ has mean zero and is a linear combination of its parents in $\g[D]^{1'}$ and $\varepsilon_{A} \sim N(0, \sigma^2_A)$, where $\{\mb{\varepsilon}_{A}: A \in \mb{V}\}$ are mutually independent. The coefficients in this linear combination are defined by the edge coefficients of $\g[D]^{1'}$. We pick these edge coefficients in conjunction with $\{\mb{\sigma}^2_{A}: A \in \mb{V}\}$ in such a way that each coefficient is in $(0,1)$ and $\Var(A) = 1$ for all $A \in \mb{V}$.

    From this, we define $\mb{F^*_1} = \{ f_1(\mb{v}|do(\mb{x'})) : \mb{X'} \subseteq \mb{V} \}$ such that $\g[D]^{1'}$ is compatible with $\mb{F^*_1}$ and such that $f_1(\mb{v}) = f(\mb{v})$. Note that $f(\mb{v})$ is Markov compatible with $\g[D]^{1'}$ by construction, and we build the interventional densities in $\mb{F^*_1}$ by replacing the intervening random variables in the SEM with their interventional values \citep{pearl2009causality}.

    To construct the second family of interventional densities, we introduce the DAG $\g[D]^{2'}$, which we form by removing every edge from $\g[D]^2$ except for the edges on $q_2$. Then note that we could have defined $f(\mb{v})$ using a linear SEM based on the parents in $\g[D]^{2'}$. In this case, the resulting observational density would again be a multivariate normal with mean vector zero and a covariance matrix with ones on the diagonal. The off-diagonal entries would be the covariances between the variables in $\g[D]^{2'}$. But note that by Lemma \ref{lem:wright}, these values will equal the product of all edge coefficients between the relevant nodes in $\g[D]^{2'}$. Since $\g[D]^{1'}$ and $\g[D]^{2'}$ contain no paths with colliders, the observational density $f(\mb{v})$ built using $\g[D]^{2'}$ will be an identical density to that built under $\g[D]^{1'}$. Thus, in an analogous way to $\mb{F^*_1}$, we define $\mb{F^*_2} = \{ f_2(\mb{v}|do(\mb{x'})) : \mb{X'} \subseteq \mb{V} \}$ such that $\g[D]^{2'}$ is compatible with $\mb{F^*_2}$ and such that $f_2(\mb{v}) = f(\mb{v})$.
        
    Having defined $\mb{F^*_1}$ and $\mb{F^*_2}$, we check that their desired properties hold. Note that by construction, $\g[D]^{1'}$ and $\g[D]^{2'}$ are compatible with $\mb{F^*_1}$ and $\mb{F^*_2}$, respectively. Thus \ref{prop:id-necessary-1} holds by Lemma \ref{lem:markov-sub}. Similarly by construction, \ref{prop:id-necessary-2} holds. To show that \ref{prop:id-necessary-3} holds, it suffices to show that $E[Y | do(\mb{X}=\mb{1}), \mb{Z}]$ is not the same under $f_1$ and $f_2$. 

    To calculate these expectations, we first want to apply Rules 1-3 of the do calculus (Theorem \ref{thm:do-calc}). Since $f_i(\mb{v} | do(\mb{x}))$, $i \in \{1,2\}$, is consistent with $\g[D]^{i'}$, we apply these rules using graphical relationships in $\g[D]^{i'}$. Because the path in $\g[D]^{i'}$ corresponding to $q_i$, $i \in \{1,2\}$, does not contain nodes in $\mb{Z}$ or $\mb{X} \setminus \{X\}$, then $Y \dsepp \mb{Z} | \mb{X}$ and $Y \dsepp \mb{X} \setminus \{X\} | X$ in $\g[D]^{i'}_{\overline{\mb{X}}}$. Further, $Y \dsepp X$ in $\g[D]^{1'}_{\underline{X}}$ and $Y \dsepp X$ in $\g[D]^{2'}_{\overline{X}}$. Thus by Rules 1-3 of the do calculus (Theorem \ref{thm:do-calc}), the following hold.
    \begin{align*}
        E_1[Y | do(\mb{X}=\mb{1}), \mb{Z}] &= E_1[Y | do(X=1)] = E_1[Y|X=1] := a.\\
        E_2[Y | do(\mb{X}=\mb{1}), \mb{Z}] &= E_2[Y | do(X=1)] = E_2[Y] := b,
    \end{align*}    
    where $E_i, i \in \{1,2\}$ is the expectation under $f_i$. To calculate $a$ and $b$, we rely on the observational density $f(\mb{v})$, which was constructed using $\g[D]^{1'}$. By Lemma \ref{lem:mardia}, $a$ equals the covariance of $X$ and $Y$ under $f(\mb{v})$, and by Lemma \ref{lem:wright}, $\Cov(X,Y)$ equals the product of all edge coefficients in $\g[D]^{1'}$, which were chosen to be in $(0,1)$. Therefore, $a \neq 0$. But by definition of $f(\mb{v})$, $b=0$.    
\end{proofof}


\subsection{Supporting Results}

\begin{lemma}
\label{lem:twopaths-corrollary}
    Let $\mb{X}$, $\mb{Y}$, and $\mb{Z}$ be pairwise disjoint node sets in an MPDAG $\g=(\mb{V},\mb{E})$. Suppose that there is a proper possibly causal path from $\mb{X}$ to $\mb{Y}$ in $\g$ that starts with an undirected edge and does not contain nodes in $\mb{Z}$. Then there is one such path $\langle X=V_0, \dots, V_k=Y \rangle$, $X \in \mb{X}$, $Y \in \mb{Y}$, $k \ge 1$, where the corresponding paths in two DAGs in $[\g]$ take the forms $X \to \dots \to Y$ and $X \gets V_1 \to \dots \to Y$ ($X \gets Y$ when $k=1$), respectively.
\end{lemma}

\begin{proofof}[Lemma \ref{lem:twopaths-corrollary}]
    This lemma is similar to Lemma A.3 of \cite{perkovic2020identifying} and its proof borrows from the proof strategy of Lemma C.1 of \cite{perkovic2017interpreting}.
    
    Let $q^*$ be an arbitrary proper possibly causal path from $\mb{X}$ to $\mb{Y}$ in $\g$ that starts with an undirected edge and does not contain nodes in $\mb{Z}$. Then let $q=\langle X=V_0, \dots, V_k=Y\rangle$, $X \in \mb{X}$, $Y \in \mb{Y}$, $k \ge 1$, be a shortest subsequence of $q^*$ in $\g$ that also starts with an undirected edge. Note that $q$ is a proper possibly causal path from $\mb{X}$ to $\mb{Y}$ in $\g$ that starts with an undirected edge and does not contain nodes in $\mb{Z}$.

    Consider when $q$ is of definite status. Since $q$ is possibly causal, all non-endpoints of $q$ are definite non-colliders. Let $\g[D]^1$ be a DAG in $[\g]$ that contains $X \to V_1$. Then since $V_1$ is either $Y$ or a definite non-collider on $q$, the path corresponding to $q$ in $\g[D]^1$ takes the form $X \to \dots \to Y$ by induction. Let $\g[D]^2$ be a DAG in $[\g]$ with no additional edges into $V_1$ compared to $\g$ (Lemma \ref{lem:undirected-imply}). Since $\g$ contains $X - V_1$, $\g[D]^2$ contains $X \gets V_1$. When $k>1$, $\g$ contains either $V_1 - V_2$ or $V_1 \to V_2$, and so $\g[D]^2$ contains $X \gets V_1 \to V_2$. Thus by the same inductive reasoning as above, the path corresponding to $q$ in $\g[D]^2$ takes the form $X \gets V_1 \to \dots \to Y$ (or simply $X \gets Y$ when $k=1$).
    
    Consider instead when $q$ is not of definite status. Note that $k>1$. To see that $q$ contains $V_1 - V_2$, note that by the choice of $q$ and the fact that $q$ is possibly causal, $q(V_1,Y)$ is unshielded and possibly causal. Thus, $q(V_1,Y)$ is of definite status. However, $q$ is not of definite status, so $V_1$ must not be of definite status on $q$, which implies that $q$ cannot contain $V_1 \to V_2$. Since $q$ is possibly causal, it also cannot contain $V_1 \gets V_2$.

    To find two DAGs in $[\g]$ with paths corresponding to $q$ that fit our desired forms, we narrow our search to $[\g']$, where we let $\g'$ be an MPDAG constructed from $\g$ by adding $V_1 \to V_2$ and completing R1-R4 of \cite{meek1995causal}. We show below that the path corresponding to $q$ in $\g'$ takes the form $X - V_1 \to \dots \to Y$, and thus, there must be two DAGs in $[\g'] \subseteq [\g]$ with corresponding paths of the forms $X \to \dots \to Y$ and $X \gets V_1 \to \dots \to Y$.
            
    We first show that $\g'$ contains $X - V_1$ by the contraposition of Lemma \ref{lem:adding-edges-imply}. Note that we have already shown that $\g$ contains $V_1 - V_2$, that $\g'$ is formed by adding $V_1 \to V_2$ to $\g$, and that $\g$ contains $X - V_1$. It remains to show that $X,V_1 \notin \De(V_2, \g')$. To see this, note that $\g$ must contain an edge $\langle X,V_2 \rangle$, because $V_1$ is not of definite status on $q$. This edge must take the form $X \to V_2$ by the choice of $q$ and the fact that $q$ is possibly causal. Thus, $\g'$ contains $X \to V_2$ and $V_1 \to V_2$. Therefore, $X,V_1 \notin \De(V_2, \g')$. Finally, note that $\g'$ contains $V_1 \to \dots \to Y$ by R1 of \cite{meek1995causal}, since we constructed $\g'$ be adding $V_1 \to V_2$ to a path $q(V_1,Y)$ that is unshielded and possibly causal.
\end{proofof}

\begin{lemma}
\label{lem:markov-sub}
    Let $\mb{X}$, $\mb{Y}$, and $\mb{Z}$ be pairwise disjoint node sets in a causal DAG $\g[D] = (\mb{V},\mb{E})$. Then let $\g[D]^* = (\mb{V},\mb{E'})$ be a causal DAG constructed by removing edges from $\g[D]$, and let $f(\mb{v} | do(\mb{x}))$ be an interventional density over $\mb{V}$. If $f(\mb{v} | do(\mb{x}))$ is consistent with $\g[D]^*$, then it is consistent with $\g[D]$.
\end{lemma}

\begin{proofof}[Lemma \ref{lem:markov-sub}]
    Suppose that $f(\mb{v} | do(\mb{x}))$ is consistent with $\g[D]^*$. Then by definition, there exists a set of interventional densities $\mb{F^*}$ such that $\g[D]^*$ is compatible with $\mb{F^*}$. Let $f(\mb{v})$ be the density in $\mb{F^*}$ under a null intervention. Note that by the truncated factorization in Equation \eqref{eq:trunc-fact}, $f(\mb{v})$ is Markov compatible with $\g[D]^*$. Thus by Lemma \ref{lem:markov-equiv},
    \begin{align}
        V_i \ind \Big[ \mb{V} \setminus \big( \De(V_i, \g[D]^*) \cup \Pa(V_i,\g[D]^*) \big) \Big] | \Pa(V_i, \g[D]^*) \label{eq:markov-sub-1}
    \end{align}
    for all $V_i \in \mb{V}$, where $\ind$ indicates independence with respect to $f(\mb{v})$. Further, since $\De(V_i,\g[D]^*) \subseteq \De(V_i,\g[D])$, then $\De(V_i,\g[D]^*) \cap \Pa(V_i,\g[D]) = \emptyset$ and thus $\Pa(V_i,\g[D]) \subseteq \mb{V} \setminus \De(V_i, \g[D]^*)$. Therefore it follows from \eqref{eq:markov-sub-1} that
    \begin{align} 
        V_i \ind \Big[ \Pa(V_i,\g[D]) \setminus \Pa(V_i,\g[D]^*) \Big] \,\, \Big| \,\, \Pa(V_i,\g[D]^*). \label{eq:markov-sub-2}
    \end{align}
    
    Let $f(\mb{v} | do(\mb{x'}))$, $\mb{X'} \subseteq \mb{V}$, be an arbitrary density in $\mb{F^*}$. Then by definition and \eqref{eq:markov-sub-2}
    \begin{align*}
        f(\mb{v} | do(\mb{x'})) &= \prod_{V_i \in \mb{V} \setminus \mb{X'}} f(v_i|\pa(v_i,\g[D]^*)) \mathbbm{1}(\mb{X'} = \mb{x'})\\
                                &= \prod_{V_i \in \mb{V} \setminus \mb{X'}} f(v_i|\pa(v_i,\g[D])) \mathbbm{1}(\mb{X'} = \mb{x'}).
    \end{align*}
    Since $f(\mb{v} | do(\mb{x'}))$ was arbitrary, this holds for all densities in $\mb{F^*}$. Thus, $\g[D]$ is compatible with $\mb{F^*}$. Since $f(\mb{v} | do(\mb{x})) \in \mb{F^*}$, then by definition, it is consistent with $\g[D]$.
\end{proofof}


\section{Proofs for Section \ref{sec:do}: Do Calculus for MPDAGs}
\label{app:do}

\begin{figure}
    \centering
    \begin{tikzpicture}[>=stealth',shorten >=1pt,node distance=3cm, main node/.style={minimum size=0.4cm}]
    [>=stealth',shorten >=1pt,node distance=3cm,initial/.style    ={}]
    \node (R1b)     at (0,1.25)     {Lem. \ref{lem:zhang-modify-1b}};
    \node (R1a)     at (-2.25,1.25) {Lem. \ref{lem:zhang-modify-1a}};
    \node (R2b)     at (0,0)        {Lem. \ref{lem:zhang-modify-2b}};
    \node (R2a)     at (-2.25,0)    {Lem. \ref{lem:zhang-modify-2a}};
    \node (R3c)     at (0,-1.25)        {Lem. \ref{lem:zhang-modify-3c}};
    \node (R3b)     at (-2.25,-1.25)    {Lem. \ref{lem:zhang-modify-3b}};
    \node (R3a)     at (-2.25,-2.5) {Lem. \ref{lem:zhang-modify-3a}};
    \node (PC)      at (-4.6,-2.5)   {Lem. \ref{lem:pc-imply1b}};
    \node (T)       at (3.2,0)   {\textbf{Theorem \ref{thm:do-calc-mpdag}}};
    \node (H1)      at (-5.25,.75)   {Lem. \ref{lem:do-helper1}};
    \node (H2)      at (-5.25,.25)   {Lem. \ref{lem:do-helper2}};
    \node (H3)      at (-5.25,-.25)  {Lem. \ref{lem:do-helper3}};
    \node (H4)      at (-5.25,-.75)  {Lem. \ref{lem:do-helper4}};
    \draw [decorate,decoration = {calligraphic brace, raise=5pt, amplitude=5pt}, thick] (-4.6,1.05) -- (-4.6,-1.05);
    \draw [decorate,decoration = {calligraphic brace, raise=5pt, amplitude=5pt}, thick] (.6,1.5) -- (.6,-1.5);
    \draw[->] (R1a) to (R1b);
    \draw[->] (R2a) to (R2b);
    \draw[->] (R3b) to (R3c);
    \draw[->] (R3a) to (R3b);
    \draw[->] (R3a) to (R3c);
    \draw[->] (PC)  to (R3a);
    \draw[->] (-4,0.2)   to (-3.05,1.15);
    \draw[->] (-4,0)     to (-3.05,0);
    \draw[->] (-4,0.05)  to[bend left] (-.95,0.15);
    \draw[->] (-4,-0.1)  to (-3.05,-1.15);
    \draw[->] (1.1,0)     to (2,0);
    \end{tikzpicture}
    \vspace{.15in}
    \caption{Proof structure of Theorem \ref{thm:do-calc-mpdag}.}
    \label{fig:do-map}
\end{figure}
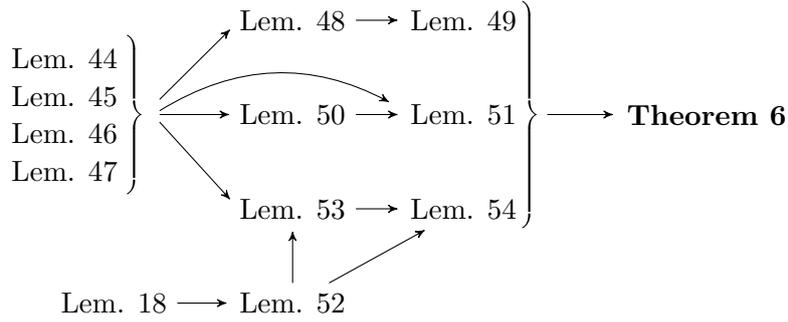

This section includes lemmas needed for the proof of our do calculus (Theorem \ref{thm:do-calc-mpdag}) found in Section \ref{sec:do}. We divide these results into those needed for all of our do rules (Lemmas \ref{lem:do-helper1}-\ref{lem:do-helper4}) and those needed for Rule 1 (Lemmas \ref{lem:zhang-modify-1a}-\ref{lem:zhang-modify-1b}), Rule 2 (Lemmas \ref{lem:zhang-modify-1a}-\ref{lem:zhang-modify-1b}), and Rule 3 (Lemmas \ref{lem:zhang-modify-3a}-\ref{lem:zhang-modify-3c}). Figure \ref{fig:do-map} provides a map of these results.

We began Section \ref{sec:do} with a discussion of the mutilated graphs seen in Theorem \ref{thm:do-calc-mpdag}, and we noted two differences between these graphs and their counterparts in Pearl's do calculus. Here, we highlight a third, more technical difference that will be useful in navigating the proofs below. Because the mutilated graphs in Theorem \ref{thm:do-calc-mpdag} might not be MPDAGs, the \textit{definite status} of a path in these graphs no longer conveys a certainty about the corresponding paths in related DAGs. To see this, consider an MPDAG $\g$ that contains the paths $V_1 - V_2 - X$ and $V_1 \to X$. The graph $\g_{\overline{X}}$ only contains the path $V_1 - V_2 - X$, where $V_2$ is a definite non-collider by definition. However, note that $[\g]$ contains a DAG with the path $V_1 \to V_2 \gets X$.

\subsection{Results for All do Rules}


\begin{lemma}
\label{lem:do-helper1}
{\normalfont (cf. Lemma 1 of \cite{zhang2006causal})}
    Let $\mb{X}$, $\mb{Y}$, $\mb{Z}$, and $\mb{W}$ be pairwise disjoint node sets in a DAG $\g[D]=(\mb{V,E})$, and let $\g$ be an MPDAG such that $\g[D] \in [\g]$. Additionally, let $p$ be a path in $\g[D]$ from $\mb{Z}$ to $\mb{Y}$ such that $p$ is d-connecting given $\mb{X} \cup \mb{W}$ but no subsequence of $p$ in $\g[D]$ is d-connecting given $\mb{X} \cup \mb{W}$. Then the following hold.
    \begin{enumerate}[label=(\roman*)]
        \item\label{lem:do-helper1-one}
        All colliders on $p$ are unshielded.
        
        \item\label{lem:do-helper1-two}
        The path in $\g$ corresponding to $p$ is of definite status.
    \end{enumerate}
\end{lemma}

\begin{proofof}[Lemma \ref{lem:do-helper1}]
    Let $p = \langle Z=V_1, \dots, V_k=Y \rangle$, $Z \in \mb{Z}$, $Y \in \mb{Y}$, $k > 1$.
    
    \ref{lem:do-helper1-one}
    Let $V_i$, $i \in \{2, \dots, k-1\}$, be an arbitrary collider on $p$. For sake of contradiction, suppose $\g[D]$ contains the edge $\langle V_{i-1}, V_{i+1} \rangle$. Define $q = p(Z,V_{i-1}) \oplus \langle V_{i-1}, V_{i+1} \rangle \oplus p(V_{i+1}, Y)$. By the choice of $p$, $q$ must be blocked by $\mb{X} \cup \mb{W}$. Since $V_{i-1}$ and $V_{i+1}$ are non-colliders on $p$, $q$ must be blocked by a collider at either $V_{i-1}$ or $V_{i+1}$. Without loss of generality, pick $V_{i-1}$ so that $\g[D]$ contains $V_{i-2} \to V_{i-1} \gets V_{i+1}$, where no descendant of $V_{i-1}$ is in $\mb{X} \cup \mb{W}$. However, since $p$ is d-connecting given $\mb{X} \cup \mb{W}$ and $p$ contains $V_{i-1} \to V_i \gets V_{i+1}$, at least one descendant of $V_i$---and therefore of $V_{i-1}$---in $\g[D]$ must be in $\mb{X} \cup \mb{W}$.

    \ref{lem:do-helper1-two}    
    Let $p^*$ be the path in $\g$ corresponding to $p$. For sake of contradiction, suppose there is a node $V_i$, $i \in \{2, \dots, k-1\}$, that is not of definite status on $p^*$. This implies that $\g$---and therefore $\g[D]$---must contain the edge $\langle V_{i-1}, V_{i+1} \rangle$. 

    Define $q = p(Z,V_{i-1}) \oplus \langle V_{i-1}, V_{i+1} \rangle \oplus p(V_{i+1}, Y)$ and consider the form of $p(V_{i-1},V_{i+1})$. Since $\g[D]$ contains $\langle V_{i-1}, V_{i+1} \rangle$, then by part \ref{lem:do-helper1-one}, it cannot contain $V_{i-1} \to V_i \gets V_{i+1}$. For sake of contradiction, suppose that $\g[D]$ contains $V_{i-1} \to V_i \to V_{i+1}$. This forces $\g[D]$ to contain $V_{i-1} \to V_{i+1}$. But since $p$ is d-connected given $\mb{X} \cup \mb{W}$, this implies that $q$ is d-connected given $\mb{X} \cup \mb{W}$, which contradicts the choice of $p$. By the same logic, $\g[D]$ cannot contain $V_{i-1} \gets V_i \gets V_{i+1}$. Therefore, $\g[D]$ must contain $V_{i-1} \gets V_i \to V_{i+1}$. 

    Without loss of generality, let $\langle V_{i-1}, V_{i+1} \rangle$ take the form $V_{i-1} \to V_{i+1}$ in $\g[D]$. Note that for $p$ to be d-connected and $q$ to be blocked given $\mb{X} \cup \mb{W}$, $V_{i-1}$ must be a collider on $p$ and a node in $\mb{X} \cup \mb{W}$. Then by part \ref{lem:do-helper1-one}, $V_{i-1}$ is unshielded on $p$. So by properties of MPDAGs, $V_{i-1}$ must also be a collider on $p^*$. Thus, $\g$ contains $V_{i-1} \gets V_i$. But this contradicts that $V_i$ is not of definite status on $p^*$.
\end{proofof}


\begin{lemma}
\label{lem:do-helper2}
    Let $\mb{X}$ and $\mb{W}$ be disjoint node sets in a DAG $\g[D]$ that has no edges into $\mb{X}$. If a path in $\g[D]$ is d-connecting given $\mb{X} \cup \mb{W}$, then it does not contain nodes in $\mb{X}$ and is d-connecting given $\mb{W}$. 
\end{lemma}

\begin{proofof}[Lemma \ref{lem:do-helper2}]
    Any path in $\g[D]$ that is d-connecting given $\mb{X} \cup \mb{W}$ cannot have non-colliders in $\mb{X}$ or $\mb{W}$. Further, since there are no edges into $\mb{X}$ in $\g[D]$, all of the colliders on such a path cannot be in $\mb{X}$ and must have descendants in $\mb{W}$.
\end{proofof}


\begin{lemma}
\label{lem:do-helper3}
    Let $\mb{X}$, $\mb{Y}$, $\mb{Z}$, and $\mb{W}$ be pairwise disjoint node sets in a DAG $\g[D]$, let $p$ be a path in $\g[D]$ from $\mb{Z}$ to $\mb{Y}$ that is d-connecting given $\mb{X} \cup \mb{W}$, and let $q$ be a subsequence of $p$ in $\g[D]$. Then all colliders on $q$ are in $\An(\mb{X} \cup \mb{W}, \g[D])$. 
\end{lemma}

\begin{proofof}[Lemma \ref{lem:do-helper3}]
    Let $p = \langle Z=V_1, \dots, V_k=Y \rangle$, $Z \in \mb{Z}$, $Y \in \mb{Y}$, $k >1$, and let $V_i$, $i \in \{2, \dots, k-1\}$, be an arbitrary collider on $q$.

    When $V_i$ is also a collider on $p$, then $V_i \in \An(\mb{X} \cup \mb{W}, \g[D])$, since $p$ is d-connecting given $\mb{X} \cup \mb{W}$. Consider when $V_i$ is not a collider on $p$. Then $p$ contains $V_{i-1} \gets V_i$ or $V_i \to V_{i+1}$. Without loss of generality, suppose $p$ contains $V_i \to V_{i+1}$. Note that $V_{i+1}$ is not on $q$, since $V_i$ is a collider on $q$. Therefore, let $V_j$, $j \in \{i+2, \dots, k\}$, be the node following $V_i$ on $q$ so that $q$ contains $V_i \gets V_j$. Since $\g[D]$ is acyclic, $p$ cannot contain $V_{i} \to \dots \to V_j$. Hence, let $C$ be the collider on $p(V_i,V_j)$ closest to $V_i$ so that $V_i \in \An(C, \g[D])$. Since $p$ is d-connecting given $\mb{X} \cup \mb{W}$, then $C \in \An(\mb{X} \cup \mb{W}, \g[D])$ and thus, $V_i \in \An(\mb{X} \cup \mb{W}, \g[D])$.
\end{proofof}


\begin{lemma}
\label{lem:do-helper4}
    Let $\mb{X}$ and $\mb{Z}$ be disjoint node sets in an MPDAG $\g$, let $\mb{X'} \subseteq \mb{X}$, let $\g[D]$ be a DAG in $[\g]$, and let $\overline{p}$ be a path in $\g[D]_{\overline{\mb{X}}\underline{\mb{Z}}}$. If the path in $\g$ corresponding to $\overline{p}$ is of definite status, then the sequence of nodes in $\g_{\overline{\mb{X'}}\underline{\mb{Z}}}$ corresponding to $\overline{p}$ forms a path where a node is a definite non-collider or collider if and only if it is a non-collider or collider, respectively, on $\overline{p}$.
\end{lemma}

\begin{proofof}[Lemma \ref{lem:do-helper4}]
    To see that the sequence of nodes in $\g_{\overline{\mb{X'}}\underline{\mb{Z}}}$ corresponding to $\overline{p}$ forms a path, note that the set of adjacencies in $\g[D]_{\overline{\mb{X}}\underline{\mb{Z}}}$ is a subset of the set of adjacencies in $\g_{\overline{\mb{X'}}\underline{\mb{Z}}}$. Then let $V$ be an arbitrary node on $\overline{p}$, and let $p$, $p^*$, and $\overline{p}^*$ be the paths in $\g[D]$, $\g$, and $\g_{\overline{\mb{X'}}\underline{\mb{Z}}}$, respectively, corresponding to $\overline{p}$. Consider the following. 
    \begin{enumerate}[label=(\alph*)]
        \item\label{lem:do-helper4-one}
        $V$ is a non-collider/collider on $\overline{p}$.
        
        \item\label{lem:do-helper4-two}
        $V$ is a non-collider/collider on $p$.

        \item\label{lem:do-helper4-three}
        $V$ is a definite non-collider/collider on $p^*$.

        \item\label{lem:do-helper4-four}
        $V$ is a definite non-collider/collider on $\overline{p}^*$.
    \end{enumerate}
    By definition of the graphs, \ref{lem:do-helper4-one} $\Leftrightarrow$ \ref{lem:do-helper4-two} and \ref{lem:do-helper4-two} $\Leftarrow$ \ref{lem:do-helper4-three}. Further, if $p^*$ is of definite status, \ref{lem:do-helper4-two} $\Rightarrow$ \ref{lem:do-helper4-three} and \ref{lem:do-helper4-three} $\Leftarrow$ \ref{lem:do-helper4-four}. Then since removing edges from $\g$ outside of $p^*$ cannot change a node of definite status on $p^*$, \ref{lem:do-helper4-three} $\Rightarrow$ \ref{lem:do-helper4-four}.
\end{proofof}


\subsection{Results for do Rule 1}


\begin{lemma}
\label{lem:zhang-modify-1a}
    Let $\mb{X,Y,Z}$, and $\mb{W}$ be pairwise disjoint node sets in an MPDAG $\g= (\mb{V,E})$, and let $\g[D]$ be a DAG in $[\g]$. Additionally, let $\overline{p}$ be a path in $\g[D]_{\overline{\mb{X}}}$ from $\mb{Z}$ to $\mb{Y}$ such that $\overline{p}$ is d-connecting given $\mb{X} \cup \mb{W}$ but no subsequence of $\overline{p}$ in $\g[D]_{\overline{\mb{X}}}$ is d-connecting given $\mb{X} \cup \mb{W}$. Then the sequence of nodes in $\g_{\overline{\mb{X}}}$ corresponding to $\overline{p}$ forms a path where a node is a definite non-collider or collider if and only if it is a non-collider or collider, respectively, on $\overline{p}$.
\end{lemma}

\begin{proofof}[Lemma \ref{lem:zhang-modify-1a}]
    Let $p$ be the path in $\g[D]$ corresponding to $\overline{p}$. If $p$ satisfies the assumptions of Lemma \ref{lem:do-helper1}, then the path in $\g$ corresponding to $\overline{p}$ is of definite status, and the claim follows by Lemma \ref{lem:do-helper4}. We show these assumptions hold below.

    First note that $p$ must be d-connecting given $\mb{X} \cup \mb{W}$, since $\overline{p}$ is d-connecting given $\mb{X} \cup \mb{W}$ and adding edges to a graph cannot block existing d-connecting paths. Next, for sake of contradiction, suppose that there is a subsequence of $p$ in $\g[D]$---call it $q$---that is d-connecting given $\mb{X} \cup \mb{W}$. By Lemma \ref{lem:do-helper2}, $\overline{p}$---and therefore $q$---does not contain nodes in $\mb{X}$. Thus, the sequence of nodes in $\g[D]_{\overline{\mb{X}}}$ corresponding to $q$ forms a path---call it $\overline{q}$. Since $q$ is d-connecting given $\mb{X} \cup \mb{W}$, none of the non-colliders on $q$---and therefore on $\overline{q}$---are in $\mb{X} \cup \mb{W}$. Further, since $\overline{p}$ is d-connecting given $\mb{X} \cup \mb{W}$, then by Lemma \ref{lem:do-helper3}, all colliders on $\overline{q}$ are in $\An(\mb{X} \cup \mb{W}, \g[D]_{\overline{\mb{X}}})$. But this implies that $\overline{q}$, which is a subsequence of $\overline{p}$, is d-connecting given $\mb{X} \cup \mb{W}$, which contradicts the choice of $\overline{p}$.
\end{proofof}


\begin{lemma}
\label{lem:zhang-modify-1b}
    Let $\mb{X,Y,Z}$, and $\mb{W}$ be pairwise disjoint node sets in an MPDAG $\g = (\mb{V,E})$. If $(\mb{Z} \dsepp \mb{Y} |\mb{X}, \mb{W})_{\g_{\overline{\mb{X}}}}$, then $(\mb{Z} \dsepp \mb{Y} |\mb{X}, \mb{W})_{\g[D]_{\overline{\mb{X}}}}$ for all DAGs $\g[D] \in [\g]$.
\end{lemma}

\begin{proofof}[Lemma \ref{lem:zhang-modify-1b}]
    We prove the contrapositive. Suppose there exists a DAG $\g[D] \in [\g]$ such that $\g[D]_{\overline{\mb{X}}}$ contains a path from $\mb{Z}$ to $\mb{Y}$ that is d-connecting given $\mb{X} \cup \mb{W}$. Let $\mb{S}$ be the set of all shortest such paths, and let $\overline{p} = \langle Z=V_1, \dots, V_k=Y \rangle$, $Z \in \mb{Z}$, $Y \in \mb{Y}$, $k >1$, be a path in $\mb{S}$ with the shortest distance to $\mb{X} \cup \mb{W}$ (Definition \ref{def:distance}).

    To complete the proof, we want to show that the sequence of nodes in $\g_{\overline{\mb{X}}}$ corresponding to $\overline{p}$ forms a path that is d-connecting given $\mb{X} \cup \mb{W}$. Note that by definition, $\overline{p}$ is d-connecting given $\mb{X} \cup \mb{W}$ and no subsequence of $\overline{p}$ in $\g[D]_{\overline{\mb{X}}}$ is d-connecting given $\mb{X} \cup \mb{W}$. Thus by Lemma \ref{lem:zhang-modify-1a}, we only need to show the following.
    \begin{enumerate}[label=(\roman*)]
        \item\label{lem:zhang-modify-1b-one}
        No non-collider on $\overline{p}$ is in $\mb{X} \cup \mb{W}$.
        
        \item\label{lem:zhang-modify-1b-two}
        Every collider on $\overline{p}$ is in $\An(\mb{X} \cup \mb{W}, \g_{\overline{\mb{X}}})$.
    \end{enumerate}
    Claim \ref{lem:zhang-modify-1b-one} holds since $\overline{p}$ is d-connecting given $\mb{X} \cup \mb{W}$. The remainder of the proof shows that claim \ref{lem:zhang-modify-1b-two} holds.

    Let $V_i$, $i \in \{2, \dots, k-1\}$, be an arbitrary collider on $\overline{p}$. Since $\overline{p}$ is d-connecting given $\mb{X} \cup \mb{W}$ in $\g[D]_{\overline{\mb{X}}}$, then $V_i \in \An(\mb{W}, \g[D]_{\overline{\mb{X}}})$. When $V_i \in \mb{W}$, then $V_i \in \An(\mb{X} \cup \mb{W}, \g_{\overline{\mb{X}}})$ and we are done.
    
    When $V_i \notin \mb{W}$, then let $\overline{q} = \langle V_i=Q_1, \dots, Q_m=W \rangle, W \in \mb{W}, m > 1$, be a shortest directed path in $\g[D]_{\overline{\mb{X}}}$ from $V_i$ to $\mb{W}$, and let $q^*$ be the path in $\g$ corresponding to $\overline{q}$. Note that no node on $q^*$ is in $\mb{X}$, since $\g[D]_{\overline{\mb{X}}}$ contains $V_{i-1} \to Q_1 \to \dots \to Q_m$. Thus if $q^*$ is directed, then the corresponding sequence of nodes in $\g_{\overline{\mb{X}}}$ will also form a directed path, which again implies that $V_i \in \An(\mb{X} \cup \mb{W}, \g_{\overline{\mb{X}}})$ and the proof is complete. We show below that $q^*$ is directed.

    We start by showing that $q^*$ contains $Q_1 \to Q_2$. Note that by Lemma \ref{lem:zhang-modify-1a}, $\g_{\overline{\mb{X}}}$---and therefore $\g$---contains $V_{i-1} \to Q_1 \gets V_{i+1}$. For sake of contradiction, suppose that $\g$---and therefore $\g[D]$---contains the edges $\langle V_{i-1,} Q_2 \rangle$ and $\langle V_{i+1,} Q_2 \rangle$. Since $\g[D]$ must contain $V_{i-1} \to Q_1 \to Q_2$ and $V_{i+1} \to Q_1 \to Q_2$, it must contain $V_{i-1} \to Q_2 \gets V_{i+1}$. And since no node on $q^*$ is in $\mb{X}$, then $\g[D]_{\overline{\mb{X}}}$ must also contain $V_{i-1} \to Q_2 \gets V_{i+1}$.

    When $Q_2$ is not on $\overline{p}$, consider the path $\overline{p}(Z, V_{i-1}) \oplus \langle V_{i-1}, Q_2, V_{i+1} \rangle \oplus \overline{p}(V_{i+1}, Y)$ in $\g[D]_{\overline{\mb{X}}}$. Note that this is a path from $\mb{Z}$ to $\mb{Y}$ that is d-connecting given $\mb{X} \cup \mb{W}$, is the same length as $\overline{p}$, and has a shorter distance than $\overline{p}$ to $\mb{X} \cup \mb{W}$, which contradicts the choice of $\overline{p}$. 
    
    We complete the contradiction by noting that there are no further cases---that is, $Q_2$ cannot be a node on $\overline{p}$. Suppose for sake of contradiction that it is, and without loss of generality, let $Q_2$ follow $V_i$ on $\overline{p}$. By the choice of $\overline{p}$ as a shortest path, the path $\overline{r} = \overline{p}(Z, Q_1) \oplus \langle Q_1, Q_2 \rangle \oplus \overline{p}(Q_2, Y)$ in $\g[D]_{\overline{\mb{X}}}$ must be blocked given $\mb{X} \cup \mb{W}$. Since $\overline{p}$ is d-connecting given $\mb{X} \cup \mb{W}$, $V_i \notin \mb{X} \cup \mb{W}$, and $\overline{r}$ contains $Q_1 \to Q_2$, then $Q_2$ must be a non-collider on $\overline{p}$ and a collider on $\overline{r}$ such that $Q_2 \notin \An(\mb{W}, \g[D]_{\overline{\mb{X}}})$. But since $V_i \in \An(\mb{W}, \g[D]_{\overline{\mb{X}}})$ and $V_i \notin \mb{W}$, then $Q_2 \in \An(\mb{W}, \g[D]_{\overline{\mb{X}}})$, which is a contradiction.
    
    Thus, $\g$ does not contain both edges $\langle V_{i-1,} Q_2 \rangle$ and $\langle V_{i+1,} Q_2 \rangle$. Since $\g$ contains $V_{i-1} \to Q_1 \gets V_{i+1}$, then by Rule 1 of \cite{meek1995causal}, $q^*$ contains $Q_1 \to Q_2$. If $m=2$, then $q^*$ is directed and we are done.

    Consider when $m>2$, and suppose for sake of contradiction that $q^*$ is shielded. That is, there exists an edge $\langle Q_j, Q_{j+2} \rangle$, $1 \le j \le m-2$, in $\g$. Since no node on $q^*$ is in $\mb{X}$, then $\g[D]_{\overline{\mb{X}}}$ also contains $\langle Q_j, Q_{j+2} \rangle$. Because $\g[D]_{\overline{\mb{X}}}$ contains $Q_j \to Q_{j+1} \to Q_{j+2}$, then it must contain $Q_j \to Q_{j+2}$. But then the path $\overline{q}(Q_1, Q_j) \oplus \langle Q_j, Q_{j+2} \rangle \oplus \overline{q}(Q_{j+2}, Q_m)$ contradicts the choice of $\overline{q}$. Thus when $m>2$, $q^*$ is unshielded and begins with $Q_1 \to Q_2$. So by Rule 1 of \cite{meek1995causal}, $q^*$ is again directed and we are done.
\end{proofof}


\subsection{Results for do Rule 2}


\begin{lemma}
\label{lem:zhang-modify-2a}
    Let $\mb{X,Y,Z}$, and $\mb{W}$ be pairwise disjoint node sets in an MPDAG $\g = (\mb{V,E})$ and let $\g[D]$ be a DAG in $[\g]$. Additionally, let $\overline{p} = \langle Z=V_1, \dots, V_k=Y \rangle$, $Z \in \mb{Z}$, $Y \in \mb{Y}$, $k > 1$, be a proper path from $\mb{Z}$ to $\mb{Y}$ in ${\g[D]_{\overline{\mb{X}}\underline{\mb{Z}}}}$ such that $\overline{p}$ is d-connecting given $\mb{X} \cup \mb{W}$ and no subsequence of $\overline{p}$ in ${\g[D]_{\overline{\mb{X}}\underline{\mb{Z}}}}$ is d-connecting given $\mb{X} \cup \mb{W}$. Let $p^*$ be the corresponding path in $\g$. 
    \begin{enumerate}[label = (\roman*)]
        \item\label{lem:zhang-modify-2a-two}
        If $\g$ has no edge $Z - V_i$, $i \in \{2, \dots, k\}$, then the sequence of nodes in $\g_{\overline{\mb{X}}\underline{\mb{Z}}}$ corresponding to $\overline{p}$ forms a path where a node is a definite non-collider or collider if and only if it is a non-collider or collider, respectively, on $\overline{p}$.

        \item\label{lem:zhang-modify-2a-three} 
        If $\g$ has an edge $Z - V_i$, $i \in \{2, \dots, k\}$, let $V_i$ be the closest such node to $Y$ on $p^*$. Then the sequence of nodes in $\g_{\overline{\mb{X}}\underline{\mb{Z}}}$ corresponding to $\langle Z, V_i \rangle \oplus \overline{p}(V_i, Y)$ forms a path where a node is a definite non-collider or collider if and only if it is a non-collider or collider, respectively, on $\overline{p}$.
    \end{enumerate}
\end{lemma}

\begin{proofof}[Lemma \ref{lem:zhang-modify-2a}]
    Let $p$ be the path in $\g[D]$ corresponding to $\overline{p}$. Before proving parts \ref{lem:zhang-modify-2a-two} and \ref{lem:zhang-modify-2a-three}, we first show that $V_1$ and $V_3, \dots, V_k$ are nodes of definite status on $p^*$. This clearly holds for the endpoints $V_1$ and $V_k$. Next, pick an arbitrary $V_j$, $j \in \{3, \dots, k-1\}$, and for sake of contradiction, suppose $V_j$ is not of definite status on $p^*$. This implies that $\g$ contains an edge $\langle V_{j-1}, V_{j+1} \rangle$. Since $\overline{p}$ is proper and has no nodes in $\mb{X}$ (Lemma \ref{lem:do-helper2}), then $\g[D]_{\overline{\mb{X}}\underline{\mb{Z}}}$ must contain $\langle V_{j-1}, V_{j+1} \rangle$.

    Define the path $\overline{r} = \overline{p}(Z, V_{j-1}) \oplus \langle V_{j-1}, V_{j+1} \rangle \oplus \overline{p}(V_{j+1}, Y)$ in $\g[D]_{\overline{\mb{X}}\underline{\mb{Z}}}$. Note that by the choice of $\overline{p}$, $\overline{r}$ cannot be d-connecting given $\mb{X} \cup \mb{W}$. But by Lemma \ref{lem:do-helper3}, any collider on $\overline{r}$ is in $\An(\mb{X} \cup \mb{W}, \g[D]_{\overline{\mb{X}}\underline{\mb{Z}}})$. Hence, the only way for $\overline{p}$ to be d-connecting but $\overline{r}$ to be blocked given $\mb{X} \cup \mb{W}$ is if $V_{j-1}$ or $V_{j+1}$ is a node in $\mb{X} \cup \mb{W}$, a collider on $\overline{p}$, and a non-collider on $\overline{r}$. 

    Without loss of generality, pick $V_{j-1}$. By Lemma \ref{lem:do-helper1}\ref{lem:do-helper1-one}, $V_{j-1}$ is unshielded on $\overline{p}$. Further, since $\overline{p}$ has no nodes in $\mb{X}$ and $\g[D]_{\overline{\mb{X}}\underline{\mb{Z}}}$ contains $V_{j-2} \to V_{j-1} \gets V_j$, then $V_{j-2}, V_j \notin \mb{X} \cup \mb{Z}$. Therefore, $V_j$ is also an unshielded collider on $p$, and by properties of MPDAGs, $\g$ must contain $V_{j-2} \to V_{j-1} \gets V_j$. But this contradicts that $V_j$ is not of definite status on $p^*$.

       \vskip .1in

    \ref{lem:zhang-modify-2a-two}
    Suppose $\g$ has no edge $Z - V_i$, $i \in \{2, \dots, k\}$, so that $p^*$ cannot start undirected. Further, note that $p^*$ cannot start with $Z \to V_2$, since this would imply $\g[D]_{\overline{\mb{X}}\underline{\mb{Z}}}$ contains $Z \to V_2$, which is a contradiction. Thus, $p^*$ must start with $Z \gets V_2$, which makes $V_2$ a node of definite status on $p^*$. Since we have already shown $V_1$ and $V_3, \dots, V_k$ are nodes of definite status on $p^*$, we have that $p^*$ is of definite status. The claim follows by Lemma \ref{lem:do-helper4}.    

       \vskip .1in
    
    \ref{lem:zhang-modify-2a-three}
    Suppose there is an edge $Z - V_i$, $i \in \{2, \dots, k\}$, in $\g$ where $V_i$ is the closest such node to $Y$ on $p^*$. Then let $q^*$ be the the path $\langle Z, V_i \rangle \oplus p^*(V_i, Y)$ in $\g$. Note that we have already shown $V_{i+1}, \dots, V_k$ are nodes of definite status on $p^*$. So by Lemma \ref{lem:do-helper4}, the sequence of nodes in $\g_{\overline{\mb{X}}\underline{\mb{Z}}}$ corresponding to $\overline{p}(V_i, Y)$ forms a path where a node is a definite non-collider or collider if and only if it is a non-collider or collider, respectively, on $\overline{p}$.

    Note that $\g_{\overline{\mb{X}}\underline{\mb{Z}}}$ also contains $Z - V_i$, and consider the path in $\g_{\overline{\mb{X}}\underline{\mb{Z}}}$ corresponding to $\langle Z, V_i \rangle \oplus \overline{p}(V_i, Y)$---call this $\overline{q}^*$. We have shown the claim holds for $\overline{q}^*(V_i, Y)$. Then since $V_1$ is an endpoint on $\overline{q}^*$, it remains to consider the status of $V_i$ on $\overline{q}^*$.
    
    For sake of contradiction, suppose that $V_i$ is not of definite status on $\overline{q}^*$. Then note that $V_i$ cannot be of definite status on $q^*$, since removing edges from $\g$ outside of $q^*$ cannot change a node of definite status of $q^*$. It follows that $\g$ must contain an edge $\langle Z,V_{i+1} \rangle$, which by choice of $V_i$, must be directed. 
    
    Consider when $\g$ contains $Z \to V_{i+1}$ so that $\g_{\overline{\mb{X}}\underline{\mb{Z}}}$ contains $Z - V_i$ but not the edge $\langle Z,V_{i+1} \rangle$. Since $V_i$ is not of definite status on $\overline{q}^*$, then $\g_{\overline{\mb{X}}\underline{\mb{Z}}}$ cannot contain $Z - V_i - V_{i+1}$ or $Z - V_i \to V_{i+1}$. Thus, $\g_{\overline{\mb{X}}\underline{\mb{Z}}}$ contains $Z - V_i \gets V_{i+1}$. But this implies that $\g$ contains $Z - V_i \gets V_{i+1}$ and $Z \to V_{i+1}$, which contradicts Rule 2 of \cite{meek1995causal}.
    
    Consider instead when $\g$ contains $Z \gets V_{i+1}$. Then $\g[D]_{\overline{\mb{X}}\underline{\mb{Z}}}$ also contains $Z \gets V_{i+1}$, since $Z \notin \mb{X}$ and $\overline{p}$ is proper. By choice of $\overline{p}$, the path $\overline{t} = \langle Z, V_{i+1} \rangle \oplus \overline{p}(V_{i+1}, Y)$ in $\g[D]_{\overline{\mb{X}}\underline{\mb{Z}}}$ must be blocked by $\mb{X} \cup \mb{W}$. Since $\overline{p}$ is d-connecting given $\mb{X} \cup \mb{W}$ and $V_{i+1}$ is a non-collider on $\overline{t}$, then $V_{i+1}$ must be a node in $\mb{X} \cup \mb{W}$ and a collider on $\overline{p}$. By Lemma \ref{lem:do-helper1}\ref{lem:do-helper1-one}, $V_{i+1}$ must be an unshielded collider on $\overline{p}$. Since $\overline{p}$ has no nodes in $\mb{X}$ (Lemma \ref{lem:do-helper2}) and $\g[D]_{\overline{\mb{X}}\underline{\mb{Z}}}$ contains $V_i \to V_{i+1} \gets V_{i+2}$, then $V_i, V_{i+2} \notin \mb{X} \cup \mb{Z}$. Therefore, $V_{i+1}$ is also an unshielded collider on $p$, and by properties of MPDAGs, $\g$ must contain $V_i \to V_{i+1} \gets V_{i+2}$. But this contradicts that $V_i$ is not of definite status on $q^*$.

    Thus, $V_i$ is of definite status on $\overline{q}^*$. Finally, we show that $V_i$ is a definite non-collider on both $\overline{q}^*$ and $\overline{p}$. Since $\overline{q}^*$ starts undirected and $V_i$ is of definite status on $\overline{q}^*$, then $V_i$ is a definite non-collider on $\overline{q}^*$. For sake of contradiction, suppose that $V_i$ is a collider on $\overline{p}$. By Lemma \ref{lem:do-helper1}\ref{lem:do-helper1-one}, $V_i$ must be an unshielded collider on $\overline{p}$. Since $\overline{p}$ has no nodes in $\mb{X}$ and $\g[D]_{\overline{\mb{X}}\underline{\mb{Z}}}$ contains $V_{i-1} \to V_i \gets V_{i+1}$, then $V_{i-1}, V_{i+1} \notin \mb{X} \cup \mb{Z}$. Therefore, $V_i$ is also an unshielded collider on $p$, and by properties of MPDAGs, $\g$---and therefore $\g_{\overline{\mb{X}}\underline{\mb{Z}}}$---must contain $V_{i-1} \to V_i \gets V_{i+1}$. But then $\g_{\overline{\mb{X}}\underline{\mb{Z}}}$ contains $Z - V_i \gets V_{i+1}$, which contradicts that $V_i$ is of definite status on $\overline{q}^*$.
\end{proofof}


\begin{lemma}
\label{lem:zhang-modify-2b}
    Let $\mb{X,Y,Z}$, and $\mb{W}$ be pairwise disjoint node sets in an MPDAG $\g = (\mb{V,E})$. If $(\mb{Z} \dsepp \mb{Y} |\mb{X}, \mb{W})_{\g_{\overline{\mb{X}}\underline{\mb{Z}}}}$, then $(\mb{Z} \dsepp \mb{Y} |\mb{X}, \mb{W})_{\g[D]_{\overline{\mb{X}}\underline{\mb{Z}}}}$ for all DAGs $\g[D] \in [\g]$.
\end{lemma}

\begin{proofof}[Lemma \ref{lem:zhang-modify-2b}]
    We prove the contrapositive. Suppose there exists a DAG $\g[D] \in [\g]$ such that $\g[D]_{\overline{\mb{X}}\underline{\mb{Z}}}$ contains a path from $\mb{Z}$ to $\mb{Y}$ that is d-connecting given $\mb{X} \cup \mb{W}$. Let $\mb{S}$ be the set of all shortest such paths, and let $\overline{p} = \langle Z=V_1, \dots, V_k=Y \rangle$, $Z \in \mb{Z}$, $Y \in \mb{Y}$, $k >1$, be a path in $\mb{S}$ with the shortest distance to $\mb{X} \cup \mb{W}$ (Definition \ref{def:distance}). 

    Note that by definition, $\overline{p}$ is proper, $\overline{p}$ is d-connecting given $\mb{X} \cup \mb{W}$, and no subsequence of $\overline{p}$ in $\g[D]_{\overline{\mb{X}}\underline{\mb{Z}}}$ is d-connecting given $\mb{X} \cup \mb{W}$. Thus by Lemma \ref{lem:zhang-modify-2a}, $\g_{\overline{\mb{X}}\underline{\mb{Z}}}$ contains a definite status path from $\mb{Z}$ to $\mb{Y}$ where a node is a definite non-collider or collider if and only if it is a non-collider or collider, respectively, on $\overline{p}$. To complete the proof, we want to show that this path is d-connecting given $\mb{X} \cup \mb{W}$. To do this, we only need to show the following.
    \begin{enumerate}[label=(\roman*)]
        \item\label{lem:zhang-modify-2b-one}
        No non-collider on $\overline{p}$ is in $\mb{X} \cup \mb{W}$.
        
        \item\label{lem:zhang-modify-2b-two}
        Every collider on $\overline{p}$ is in $\An(\mb{X} \cup \mb{W}, \g_{\overline{\mb{X}}\underline{\mb{Z}}})$.
    \end{enumerate}
    Claim \ref{lem:zhang-modify-2b-one} holds since $\overline{p}$ is d-connecting given $\mb{X} \cup \mb{W}$. The remainder of the proof shows that claim \ref{lem:zhang-modify-2b-two} holds.

    Let $V_i$, $i \in \{2, \dots, k-1\}$, be an arbitrary collider on $\overline{p}$. Since $\overline{p}$ is d-connecting given $\mb{X} \cup \mb{W}$ in $\g[D]_{\overline{\mb{X}}\underline{\mb{Z}}}$, then $V_i \in \An(\mb{W}, \g[D]_{\overline{\mb{X}}\underline{\mb{Z}}})$. When $V_i \in \mb{W}$, then $V_i \in \An(\mb{X} \cup \mb{W}, \g_{\overline{\mb{X}}\underline{\mb{Z}}})$ and we are done.
    
    When $V_i \notin \mb{W}$, then let $\overline{q} = \langle V_i=Q_1, \dots, Q_m=W \rangle, W \in \mb{W}, m > 1$, be a shortest directed path in $\g[D]_{\overline{\mb{X}}\underline{\mb{Z}}}$ from $V_i$ to $\mb{W}$, and let $q^*$ be the path in $\g$ corresponding to $\overline{q}$. Note that no node on $q^*$ is in $\mb{X} \cup \mb{Z}$, since $\g[D]_{\overline{\mb{X}}\underline{\mb{Z}}}$ contains $V_{i-1} \to Q_1 \to \dots \to Q_m$ and $Q_m \in \mb{W}$. Thus if $q^*$ is directed, then the corresponding sequence of nodes in $\g_{\overline{\mb{X}}\underline{\mb{Z}}}$ will also form a directed path, which again implies that $V_i \in \An(\mb{X} \cup \mb{W}, \g_{\overline{\mb{X}}\underline{\mb{Z}}})$ and the proof is complete. We show below that $q^*$ is directed.

    We start by showing that $q^*$ contains $Q_1 \to Q_2$. Note that $V_i$ is an unshielded collider on $\overline{p}$ by Lemma \ref{lem:do-helper1}\ref{lem:do-helper1-one}. Since $\overline{p}$ has no nodes in $\mb{X}$ (Lemma \ref{lem:do-helper2}) and $\g[D]_{\overline{\mb{X}}\underline{\mb{Z}}}$ contains $V_{i-1} \to V_i \gets V_{i+1}$, then $V_{i-1}, V_{i+1} \notin \mb{X} \cup \mb{Z}$. Therefore, $V_i$ is also an unshielded collider on the path in $\g[D]$ corresponding to $\overline{p}$, and by properties of MPDAGs, $\g$ contains $V_{i-1} \to Q_1 \gets V_{i+1}$. 
    
    For sake of contradiction, suppose that $\g$---and therefore $\g[D]$---contains the edges $\langle V_{i-1,} Q_2 \rangle$ and $\langle V_{i+1,} Q_2 \rangle$. Since $\g[D]$ must contain $V_{i-1} \to Q_1 \to Q_2$ and $V_{i+1} \to Q_1 \to Q_2$, it must contain $V_{i-1} \to Q_2 \gets V_{i+1}$. And since no node on $q^*$ is in $\mb{X}$ and $V_{i-1}, V_{i+1} \notin \mb{Z}$, then $\g[D]_{\overline{\mb{X}}\underline{\mb{Z}}}$ must also contain $V_{i-1} \to Q_2 \gets V_{i+1}$. When $Q_2$ is not on $\overline{p}$, consider the path $\overline{p}(Z, V_{i-1}) \oplus \langle V_{i-1}, Q_2, V_{i+1} \rangle \oplus \overline{p}(V_{i+1}, Y)$ in $\g[D]_{\overline{\mb{X}}\underline{\mb{Z}}}$. Note that this is a path from $\mb{Z}$ to $\mb{Y}$ that is d-connecting given $\mb{X} \cup \mb{W}$, is the same length as $\overline{p}$, and has a shorter distance than $\overline{p}$ to $\mb{X} \cup \mb{W}$, which contradicts the choice of $\overline{p}$.
    
    We complete the contradiction by noting that there are no further cases---that is, $Q_2$ cannot be a node on $\overline{p}$. Suppose for sake of contradiction that it is, and without loss of generality, let $Q_2$ follow $V_i$ on $\overline{p}$. By the choice of $\overline{p}$ as a shortest path, the path $\overline{r} = \overline{p}(Z, Q_1) \oplus \langle Q_1, Q_2 \rangle \oplus \overline{p}(Q_2, Y)$ in $\g[D]_{\overline{\mb{X}}\underline{\mb{Z}}}$ must be blocked given $\mb{X} \cup \mb{W}$. Since $\overline{p}$ is d-connecting given $\mb{X} \cup \mb{W}$, $V_i \notin \mb{X} \cup \mb{W}$, and $\overline{r}$ contains $Q_1 \to Q_2$, then $Q_2$ must be a non-collider on $\overline{p}$ and a collider on $\overline{r}$ such that $Q_2 \notin \An(\mb{W}, \g[D]_{\overline{\mb{X}}\underline{\mb{Z}}})$. But since $V_i \in \An(\mb{W}, \g[D]_{\overline{\mb{X}}\underline{\mb{Z}}})$ and $V_i \notin \mb{W}$, then $Q_2 \in \An(\mb{W}, \g[D]_{\overline{\mb{X}}\underline{\mb{Z}}})$, which is a contradiction.
    
    Thus, $\g$ does not contain both edges $\langle V_{i-1,} Q_2 \rangle$ and $\langle V_{i+1,} Q_2 \rangle$. Since $\g$ contains $V_{i-1} \to Q_1 \gets V_{i+1}$, then by Rule 1 of \cite{meek1995causal}, $q^*$ contains $Q_1 \to Q_2$. If $m=2$, then $q^*$ is directed and we are done.

    Consider when $m>2$, and suppose for sake of contradiction that $q^*$ is shielded. That is, there exists an edge $\langle Q_j, Q_{j+2} \rangle$, $1 \le j \le m-2$, in $\g$. Since no node on $q^*$ is in $\mb{X} \cup \mb{Z}$, then $\g[D]_{\overline{\mb{X}}\underline{\mb{Z}}}$ also contains $\langle Q_j, Q_{j+2} \rangle$. Because $\g[D]_{\overline{\mb{X}}\underline{\mb{Z}}}$ contains $Q_j \to Q_{j+1} \to Q_{j+2}$, then it must contain $Q_j \to Q_{j+2}$. But then the path $\overline{q}(Q_1, Q_j) \oplus \langle Q_j, Q_{j+2} \rangle \oplus \overline{q}(Q_{j+2}, Q_m)$ contradicts the choice of $\overline{q}$. Thus when $m>2$, $q^*$ is unshielded and begins with $Q_1 \to Q_2$. So by Rule 1 of \cite{meek1995causal}, $q^*$ is again directed and we are done.
\end{proofof}


\subsection{Results for do Rule 3}


\begin{lemma}
\label{lem:zhang-modify-3a}
    Let $\mb{X}$, $\mb{Y}$, $\mb{Z}$, and $\mb{W}$ be pairwise disjoint node sets in an MPDAG $\g = (\mb{V,E})$ and let $\g[D]$ be a DAG in $[\g]$. Define $\mb{Z(W)} = \mb{Z} \setminus \An(\mb{W}, \g[D]_{\overline{\mb{X}}})$ and $\mb{Z'(W)} = \mb{Z} \setminus \PossAn(\mb{W}, \g_{\mb{V \setminus X}})$. Then $\mb{Z'(W)} \subseteq \mb{Z(W)}$.
\end{lemma}

\begin{proofof}[Lemma \ref{lem:zhang-modify-3a}]
    Note that we can write $\mb{Z(W)} = \mb{Z} \setminus \big[ \An(\mb{W}, \g[D]_{\overline{\mb{X}}}) \setminus \mb{X} \big]$. Thus to prove this claim, we show that $\An(\mb{W}, \g[D]_{\overline{\mb{X}}}) \setminus \mb{X} \subseteq \PossAn(\mb{W}, \g_{\mb{V \setminus X}})$.
    
    Pick any $A \in \An(\mb{W}, \g[D]_{\overline{\mb{X}}}) \setminus \mb{X}$ and let $\overline{p}$ be an arbitrary causal path in $\g[D]_{\overline{\mb{X}}}$ from $A$ to $\mb{W}$. Since the path in $\g[D]$ corresponding to $\overline{p}$ must be causal, then by Lemma \ref{lem:pc-imply1b}, the corresponding path in $\g$ must be possibly causal. Note that no node on $\overline{p}$ is in $\mb{X}$. Thus, the sequence of nodes in $\g_{\mb{V \setminus X}}$ corresponding to $\overline{p}$ forms a path. Further, since removing edges from $\g$ that are outside of a possibly causal path cannot make that path non-causal, then the path in $\g_{\mb{V \setminus X}}$ corresponding to $\overline{p}$ must be possibly causal. Therefore, $A \in \PossAn(\mb{W}, \g_{\mb{V \setminus X}})$.
\end{proofof}


\begin{lemma}
\label{lem:zhang-modify-3b}
    Let $\mb{X,Y,Z}$, and $\mb{W}$ be pairwise disjoint node sets in an MPDAG $\g = (\mb{V,E})$ and let $\g[D]$ be a DAG in $[\g]$. Define $\mb{Z(W)} = \mb{Z} \setminus \An(\mb{W}, \g[D]_{\overline{\mb{X}}})$. Then let $\overline{p}$ be a proper path from $\mb{Z}$ to $\mb{Y}$ in $\g[D]_{\overline{\mb{X \, Z(W)}}}$ such that $\overline{p}$ is d-connecting given $\mb{X} \cup \mb{W}$ but no subsequence of $\overline{p}$ in $\g[D]_{\overline{\mb{X \, Z(W)}}}$ is d-connecting given $\mb{X} \cup \mb{W}$. Define $\mb{Z'(W)} = \mb{Z} \setminus \PossAn(\mb{W}, \g_{\mb{V \setminus X}})$. Then the sequence of nodes in $\g_{\overline{\mb{X \, Z'(W)}}}$ corresponding to $\overline{p}$ forms a path where a node is a definite non-collider or collider if and only if it is a non-collider or collider, respectively, on $\overline{p}$.
\end{lemma}

\begin{proofof}[Lemma \ref{lem:zhang-modify-3b}]
    Let $p$ be the path in $\g[D]$ corresponding to $\overline{p}$. If $p$ satisfies the assumptions of Lemma \ref{lem:do-helper1}, then the path in $\g$ corresponding to $\overline{p}$ is of definite status, and the claim follows by Lemmas \ref{lem:zhang-modify-3a} and \ref{lem:do-helper4}. We show these assumptions hold below.

    First note that $p$ must be d-connecting given $\mb{X} \cup \mb{W}$, since $\overline{p}$ is d-connecting given $\mb{X} \cup \mb{W}$ and adding edges to a graph cannot block existing d-connecting paths. Next, for sake of contradiction, suppose that there is a subsequence of $p$ in $\g[D]$---call it $q$---that is d-connecting given $\mb{X} \cup \mb{W}$. 

    Consider when the sequence of nodes in $\g[D]_{\overline{\mb{X \, Z(W)}}}$ corresponding to $q$ forms a path---call it $\overline{q}$. Since $q$ is d-connecting given $\mb{X} \cup \mb{W}$, none of the non-colliders on $q$---and therefore on $\overline{q}$---are in $\mb{X} \cup \mb{W}$. Further, since $\overline{p}$ is d-connecting given $\mb{X} \cup \mb{W}$, then by Lemma \ref{lem:do-helper3}, all colliders on $\overline{q}$ are in $\An(\mb{X} \cup \mb{W}, \g[D]_{\overline{\mb{X \, Z(W)}}})$. But this implies that $\overline{q}$, which is a subsequence of $\overline{p}$, is d-connecting given $\mb{X} \cup \mb{W}$, which contradicts the choice of $\overline{p}$.

    Consider instead when the sequence of nodes in $\g[D]_{\overline{\mb{X \, Z(W)}}}$ corresponding to $q$ does not form a path. This implies that $q$ must contain an edge into $\mb{X} \cup \mb{Z(W)}$. By Lemma \ref{lem:do-helper2}, $\overline{p}$---and therefore $q$---is proper and has no nodes in $\mb{X}$. Thus, $q$ must start with an edge into $Z$ and $Z$ must be a node in $\mb{Z(W)}$. But since $\overline{p}$ is a path in $\g[D]_{\overline{\mb{X \, Z(W)}}}$, then $\overline{p}$---and therefore $p$---must start with an edge out of $Z$. Let $Q_2$ and $V_2$ be the distinct second nodes on $q$ and $p$, respectively, so that $\g[D]$ contains $Z \gets Q_2$ and $Z \to V_2 \dots Q_2$. By the acyclicity of $\g[D]$, there must be a collider on $p(Z, Q_2)$---and therefore on $\overline{p}(Z, Q_2)$. 
    
    Let $C$ be the closest collider to $Z$ on $\overline{p}$. Since $\overline{p}$ is d-connecting given $\mb{X} \cup \mb{W}$ in $\g[D]_{\overline{\mb{X \, Z(W)}}}$, then $C \in \An(\mb{W},\g[D]_{\overline{\mb{X \, Z(W)}}})$. Further, since $\g[D]$ contains $Z \to \dots \to C$, then $Z \in \An(\mb{W},\g[D]_{\overline{\mb{X \, Z(W)}}}) \subseteq \An(\mb{W}, \g[D]_{\overline{\mb{X}}})$, which contradicts that $Z \in \mb{Z(W)}$.
\end{proofof}


\begin{lemma}
\label{lem:zhang-modify-3c}
    Let $\mb{X,Y,Z}$, and $\mb{W}$ be pairwise disjoint node sets in an MPDAG $\g = (\mb{V,E})$. If $(\mb{Z} \dsepp \mb{Y} |\mb{X}, \mb{W})_{\g_{\overline{\mb{X \, Z'(W)}}}}$, then $(\mb{Z} \dsepp \mb{Y} |\mb{X}, \mb{W})_{\g[D]_{\overline{\mb{X \, Z(W)}}}}$ for all DAGs $\g[D] \in [\g]$, where we define $\mb{Z'(W)} = \mb{Z} \setminus \PossAn(\mb{W}, \g_{\mb{V \setminus X}})$ and $\mb{Z(W)} = \mb{Z} \setminus \An(\mb{W}, \g[D]_{\overline{\mb{X}}})$.
\end{lemma}

\begin{proofof}[Lemma \ref{lem:zhang-modify-3c}]
    We prove the contrapositive. Suppose there exists a DAG $\g[D] \in [\g]$ such that $\g[D]_{\overline{\mb{X \, Z(W)}}}$ contains a path from $\mb{Z}$ to $\mb{Y}$ that is d-connecting given $\mb{X} \cup \mb{W}$. Let $\mb{S}$ be the set of all shortest such paths, and let $\overline{p} = \langle Z=V_1, \dots, V_k=Y \rangle$, $Z \in \mb{Z}$, $Y \in \mb{Y}$, $k >1$, be a path in $\mb{S}$ with the shortest distance to $\mb{X} \cup \mb{W}$ (Definition \ref{def:distance}).

    To complete the proof, we want to show that the sequence of nodes in $\g_{\overline{\mb{X \, Z'(W)}}}$ corresponding to $\overline{p}$ forms a path that is d-connecting given $\mb{X} \cup \mb{W}$. Note that by definition, $\overline{p}$ is proper, $\overline{p}$ is d-connecting given $\mb{X} \cup \mb{W}$, and no subsequence of $\overline{p}$ in $\g[D]_{\overline{\mb{X \, Z(W)}}}$ is d-connecting given $\mb{X} \cup \mb{W}$. Thus by Lemma \ref{lem:zhang-modify-3b}, we only need to show the following.
    \begin{enumerate}[label=(\roman*)]
        \item\label{lem:zhang-modify-3c-one}
        No non-collider on $\overline{p}$ is in $\mb{X} \cup \mb{W}$.
        
        \item\label{lem:zhang-modify-3c-two}
        Every collider on $\overline{p}$ is in $\An(\mb{X} \cup \mb{W}, \g_{\overline{\mb{X \, Z'(W)}}})$.
    \end{enumerate}
    Claim \ref{lem:zhang-modify-3c-one} holds since $\overline{p}$ is d-connecting given $\mb{X} \cup \mb{W}$. The remainder of the proof shows that claim \ref{lem:zhang-modify-3c-two} holds.

    Let $V_i$, $i \in \{2, \dots, k-1\}$, be an arbitrary collider on $\overline{p}$. Since $\overline{p}$ is d-connecting given $\mb{X} \cup \mb{W}$ in $\g[D]_{\overline{\mb{X \, Z(W)}}}$, then $V_i \in \An(\mb{W}, \g[D]_{\overline{\mb{X \, Z(W)}}})$. When $V_i \in \mb{W}$, then $V_i \in \An(\mb{X} \cup \mb{W}, \g_{\overline{\mb{X \, Z'(W)}}})$ and we are done.

    When $V_i \notin \mb{W}$, then let $\overline{q} = \langle V_i=Q_1, \dots, Q_m=W \rangle, W \in \mb{W}, m > 1$, be a shortest directed path in $\g[D]_{\overline{\mb{X \, Z(W)}}}$ from $V_i$ to $\mb{W}$, and let $q^*$ be the path in $\g$ corresponding to $\overline{q}$. Note that no node on $q^*$ is in $\mb{X} \cup \mb{Z(W)}$, since $\g[D]_{\overline{\mb{X \, Z(W)}}}$ contains $V_{i-1} \to Q_1 \to \dots \to Q_m$. By Lemma \ref{lem:zhang-modify-3a}, this implies that no node on $q^*$ is in $\mb{X} \cup \mb{Z'(W)}$. Thus, if $q^*$ is directed, then the corresponding sequence of nodes in $\g_{\overline{\mb{X \, Z'(W)}}}$ will also form a directed path, which again implies that $V_i \in \An(\mb{X} \cup \mb{W}, \g_{\overline{\mb{X \, Z'(W)}}})$ and the proof is complete. We show below that $q^*$ is directed.

    We start by showing that $q^*$ contains $Q_1 \to Q_2$. Note that by Lemma \ref{lem:zhang-modify-3b}, $\g_{\overline{\mb{X \, Z'(W)}}}$---and therefore $\g$---contains $V_{i-1} \to Q_1 \gets V_{i+1}$. For sake of contradiction, suppose that $\g$---and therefore $\g[D]$---contains the edges $\langle V_{i-1,} Q_2 \rangle$ and $\langle V_{i+1,} Q_2 \rangle$. Since $\g[D]$ must contain $V_{i-1} \to Q_1 \to Q_2$ and $V_{i+1} \to Q_1 \to Q_2$, it must contain $V_{i-1} \to Q_2 \gets V_{i+1}$. And since no node on $q^*$ is in $\mb{X} \cup \mb{Z(W)}$, then $\g[D]_{\overline{\mb{X \, Z(W)}}}$ must also contain $V_{i-1} \to Q_2 \gets V_{i+1}$.

    When $Q_2$ is not on $\overline{p}$, consider the path $\overline{p}(Z, V_{i-1}) \oplus \langle V_{i-1}, Q_2, V_{i+1} \rangle \oplus \overline{p}(V_{i+1}, Y)$ in $\g[D]_{\overline{\mb{X \, Z(W)}}}$. Note that this is a path from $\mb{Z}$ to $\mb{Y}$ that is d-connecting given $\mb{X} \cup \mb{W}$, is the same length as $\overline{p}$, and has a shorter distance than $\overline{p}$ to $\mb{X} \cup \mb{W}$, which contradicts the choice of $\overline{p}$. 

    We complete the contradiction by noting that there are no further cases---that is, $Q_2$ cannot be a node on $\overline{p}$. Suppose for sake of contradiction that it is, and without loss of generality, let $Q_2$ follow $V_i$ on $\overline{p}$. By the choice of $\overline{p}$ as a shortest path, the path $\overline{r} = \overline{p}(Z, Q_1) \oplus \langle Q_1, Q_2 \rangle \oplus \overline{p}(Q_2, Y)$ in $\g[D]_{\overline{\mb{X \, Z(W)}}}$ must be blocked given $\mb{X} \cup \mb{W}$. Since $\overline{p}$ is d-connecting given $\mb{X} \cup \mb{W}$, $V_i \notin \mb{X} \cup \mb{W}$, and $\overline{r}$ contains $Q_1 \to Q_2$, then $Q_2$ must be a non-collider on $\overline{p}$ and a collider on $\overline{r}$ such that $Q_2 \notin \An(\mb{W}, \g[D]_{\overline{\mb{X \, Z(W)}}})$. But since $V_i \in \An(\mb{W}, \g[D]_{\overline{\mb{X \, Z(W)}}})$ and $V_i \notin \mb{W}$, then $Q_2 \in \An(\mb{W}, \g[D]_{\overline{\mb{X \, Z(W)}}})$, which is a contradiction.

    Thus, $\g$ does not contain both edges $\langle V_{i-1,} Q_2 \rangle$ and $\langle V_{i+1,} Q_2 \rangle$. Since $\g$ contains $V_{i-1} \to Q_1 \gets V_{i+1}$, then by Rule 1 of \cite{meek1995causal}, $q^*$ contains $Q_1 \to Q_2$. If $m=2$, then $q^*$ is directed and we are done.

    Consider when $m>2$, and suppose for sake of contradiction that $q^*$ is shielded. That is, there exists an edge $\langle Q_j, Q_{j+2} \rangle$, $1 \le j \le m-2$, in $\g$. Since no node on $q^*$ is in $\mb{X} \cup \mb{Z(W)}$, then $\g[D]_{\overline{\mb{X \, Z(W)}}}$ also contains $\langle Q_j, Q_{j+2} \rangle$. Because $\g[D]_{\overline{\mb{X \, Z(W)}}}$ contains $Q_j \to Q_{j+1} \to Q_{j+2}$, then it must contain $Q_j \to Q_{j+2}$. But then the path $\overline{q}(Q_1, Q_j) \oplus \langle Q_j, Q_{j+2} \rangle \oplus \overline{q}(Q_{j+2}, Q_m)$ contradicts the choice of $\overline{q}$. Thus when $m>2$, $q^*$ is unshielded and begins with $Q_1 \to Q_2$. So by Rule 1 of \cite{meek1995causal}, $q^*$ is again directed and we are done.
\end{proofof}


\section{Proofs for Section \ref{sec:alg-condition}: Identification Algorithm}
\label{app:alg}

This section includes the proof of Theorem \ref{thm:alg-complete} found in Section \ref{sec:alg-condition}. Four supporting results needed for this proof follow.


\subsection{Main Result}

\begin{proofofnoqed}[Theorem \ref{thm:alg-complete}]
    Suppose there is a path in $\g = (\mb{V, E})$ from $X \in \mb{X}$ to $\mb{Y} \cup \mb{Z}$ that does not contain $\mb{X} \setminus \{X\}$, is possibly causal, starts undirected---and where $\mb{Y} \not\dsepp X \,|\, \mb{X} \setminus \{X\}, \mb{Z}$ in $\g_{\overline{\mb{X} \setminus \{X\}} \underline{X}}$. We want to show that the conditional effect of $\mb{X}$ on $\mb{Y}$ given $\mb{Z}$ is not identifiable in $\g$. Since ${\mb{Y} \not\dsepp X \,|\, \mb{X} \setminus \{X\}, \mb{Z}}$ in $\g_{\overline{\mb{X} \setminus \{X\}} \underline{X}}$, the following sets must be non-empty:
    \begin{align*}
        \mb{S_1} &= \left\{\begin{array}{ll}
                        \text{definite status paths in $\g_{\overline{\mb{X} \setminus \{X\}} \underline{X}}$ from $X$ to $\mb{Y}$ that are} \\
                        \text{d-connecting given $\mb{X} \setminus \{X\} \cup \mb{Z}$}
                    \end{array}\right\} \text{ and}\\
        \mb{S_2} &= \hspace{.05in}\{ \hspace{.05in}\text{paths in $\mb{S_1}$ with the fewest colliders} \hspace{.05in}\}.
    \end{align*}
    Then consider the following set:
    \begin{align*}
        \mb{S_3} &= \hspace{.05in}\{ \hspace{.05in}\text{paths in } \mb{S_2} \text{ that start undirected}\hspace{.05in}\}. \hspace{1.8in}
    \end{align*}
    We divide the remainder of the proof into two cases based on whether $\mb{S_3} = \emptyset$. In both cases, we consider DAGs $\g[D]^1, \g[D]^2 \in [\g]$ with paths in the sets above. To show the conditional effect is not identifiable in $\g$, it suffices to show that each $\g[D]^i$ is compatible with a family of interventional densities $\mb{F^*_i} = \{ f_i(\mb{v}|do(\mb{x'})) : \mb{X'} \subseteq \mb{V} \}$ such that the following hold.
    \begin{align}
        f_1(\mb{v}) &\,=\, f_2(\mb{v}). \label{thm:alg-complete-1}\\
        f_1(\mb{y} \,|\, do(\mb{x}), \mb{z}) & \,\neq\, f_2(\mb{y} \,|\, do(\mb{x}), \mb{z}). \label{thm:alg-complete-2}
    \end{align}

    Before proceeding to cases, we briefly show that all paths in $\mb{S_1}$ are proper paths from $\mb{X}$ to $\mb{Y}$ that are open given $\mb{Z}$. To see this, note that since every path in $\mb{S_1}$ is of definite status and open given $\mb{X} \setminus \{X\} \cup \mb{Z}$, it cannot have definite non-colliders in $\mb{X} \setminus \{X\} \cup \mb{Z}$. Further, since there are no edges into $\mb{X} \setminus \{X\}$ in $\g_{\overline{\mb{X} \setminus \{X\}} \underline{X}}$, all of the colliders on any path in $\mb{S_1}$ cannot be in $\mb{X} \setminus \{X\}$ and must have descendants in $\mb{Z}$.

    \vskip .2in

    \noindent
    \textbf{CASE 1:}
    Consider when $\mb{S_3} \neq \emptyset$, and define the following sets.
    \begin{align*}
        \mb{S_4} &= \{ \text{paths in $\mb{S_3}$ with the fewest edges} \}.\\
        \mb{S_5} &= \{ \text{paths in $\mb{S_4}$ with a shortest distance to $\mb{Z}$ in $\g_{\overline{\mb{X} \setminus \{X\}} \underline{X}}$} \}.
    \end{align*}
    Pick an arbitrary path $\overline{p} := \langle X = P_1, \dots, P_k = Y \rangle$, $k > 1$, $Y \in \mb{Y}$, in $\mb{S_5} \subseteq \mb{S_1}$. Note that the sequence of nodes in $\g$ corresponding to $\overline{p}$ in $\g_{\overline{\mb{X} \setminus \{X\}} \underline{X}}$ forms a path $p$ in $\g$ that contains the same edge orientations as $\overline{p}$. Then note that since $\overline{p} \in \mb{S_1}$, we have shown that $\overline{p}$---and therefore $p$---is a proper path from $\mb{X}$ to $\mb{Y}$. Further, both paths start undirected, since $\overline{p} \in \mb{S_3}$.

    Consider first when $p$ is possibly causal so that $p$---and therefore $\overline{p}$---has no colliders. Since $\overline{p} \in \mb{S_1}$ is of definite status and we have shown that paths in $\mb{S_1}$ are open given $\mb{Z}$, then $\overline{p}$---and therefore $p$---does not contain nodes in $\mb{Z}$. Thus, $p$ is a proper, possibly causal path from $\mb{X}$ to $\mb{Y}$ in $\g$ that starts undirected and does not contain nodes in $\mb{Z}$. It follows by Proposition \ref{prop:id-necessary} that the conditional effect of $\mb{X}$ on $\mb{Y}$ given $\mb{Z}$ is not identifiable in $\g$, and we are done.

    Hence, we consider when $p$ is not possibly causal for the remainder of this case. For sake of contradiction, suppose $\overline{p}$ does not contain colliders. Since $\overline{p}$ is of definite status and starts undirected, then $\overline{p}$---and therefore $p$---cannot contain an edge $P_i \gets P_{i+1}$, $i \in \{1, \dots, k-1\}$. Thus by Lemma \ref{lem:pc-def-stat}, $p$ is not of definite status. The only way for $\overline{p}$ to be of definite status while $p$ is not is for the paths to contain at least one subpath $P_i - P_{i+1} - P_{i+2}$ that is unshielded in $\overline{p}$ but shielded in $p$. This forces $i=1$, since $\overline{p}$ is a proper path in $\g_{\overline{\mb{X} \setminus \{X\}} \underline{X}}$. That is, $\overline{p}$ begins $X - P_2 - P_3$ and $\g$ contains $X \to P_3$. It follows that $p(P_2, P_k)$ is of definite status. Since $p$---and therefore $p(P_2, P_k)$---does not contain an edge $P_i \gets P_{i+1}$, then by Lemma \ref{lem:pc-def-stat}, $p(P_2, P_k)$ is possibly causal. However, $p$ is not possibly causal, so $\g$ must contain an edge $X \gets P_j$, $j \in \{4, \dots, k\}$, so that $\g$ contains $P_3 \gets X \gets P_j$. But this contradicts that $p(P_2, P_k)$ is possibly causal (Lemma \ref{lem:nobackpaths}).

    Thus, $\overline{p}$ has at least one collider. Let $\mb{C} = \{C_1, \dots, C_m \}$, $1 \le m \le k-3$, be the set of all colliders on $\overline{p}$ in order of their appearance on $\overline{p}$. Since we have shown that paths in $\mb{S_1}$ are open given $\mb{Z}$, every collider in $\mb{C}$ is in $\An(\mb{Z},\g_{\overline{\mb{X} \setminus \{X\}} \underline{X}})$. Then for each $i \in \{1, \dots, m\}$, let $\overline{q_i} = \langle C_i = Q_{i_1}, \dots, Q_{i_{k_i}}=Z_i \rangle$ be a shortest causal path from $C_i \in \mb{C}$ to $\mb{Z}$ in $\g_{\overline{\mb{X} \setminus \{X\}} \underline{X}}$.
    
    Using these paths, we turn to the task of finding DAGs in $[\g]$ that are compatible with families of interventional densities such that \eqref{thm:alg-complete-1}-\eqref{thm:alg-complete-2} hold. To do this, we consider the subcases below.
    \begin{itemize}
        \item \textbf{SUBCASE 1:}
        Let $p$ be of definite status. Consider any two DAGs $\g[D]^1, \g[D]^2 \in [\g]$ that include the edges $X \to P_2$ and $X \gets P_2$, respectively. Then construct $\g[D]^{1'}$ and $\g[D]^{2'}$ by removing every edge from $\g[D]^1$ and $\g[D]^2$ other than those on paths corresponding to $\overline{p}$ and $\overline{q_1}, \dots, \overline{q_m}$. Since $p$ is of definite status, note that the paths in $\g[D]^{1'}$ and $\g[D]^{2'}$ corresponding to $p$ begin $X \to P_2 \to \dots \to C_1$, and $X \gets P_2 \dots C_1$, respectively.

        \item \textbf{SUBCASE 2:}
        Let $p$ be of non-definite status. As before, this forces $\overline{p}$ to begin $X - P_2 - P_3$; $\g$ to contain $X \to P_3$; and $P_2$ to be the only node on $p$ not of definite status. To choose DAGs in $[\g]$, we first restrict the class. Thus, construct the MPDAG $\g^*$ by adding $P_2 \to P_3$ to $\g$ and completing \citeauthor{meek1995causal}'s (\citeyear{meek1995causal}) orientation rules. Note that by Lemma \ref{lem:adding-edges-imply}, $\g^*$ contains $X - P_2$. Now consider any two DAGs $\g[D]^1, \g[D]^2 \in [\g^*] \subseteq [\g]$ that include the edges $X \to P_2$ and $X \gets P_2$, respectively. And construct $\g[D]^{1'}$ and $\g[D]^{2'}$ by removing every edge from $\g[D]^1$ and $\g[D]^2$ other than those on paths corresponding to $\overline{p}$ and $\overline{q_1}, \dots, \overline{q_m}$. Note that the paths in $\g[D]^{1'}$ and $\g[D]^{2'}$ corresponding to $p$ begin $X \to P_2 \to \dots \to C_1$, and $X \gets P_2 \to \dots \to C_1$, respectively.
    \end{itemize}
    
    Now that we have chosen $\g[D]^1, \g[D]^2 \in [\g]$, we construct a family of interventional densities $\mb{F^*_i} = \{ f_i(\mb{v}|do(\mb{x'})) : \mb{X'} \subseteq \mb{V} \}$ compatible with each DAG. Start by considering the following linear structural equation model (SEM). Each random variable $V_i \in \mb{V}$ is a linear combination of its parents in $\g[D]^{1'}$ and $\varepsilon_{v_i} \sim N(0, \sigma^2_{v_i})$, where $\mb{\sigma}^2_{v_i} \in (0,1]$ and $\{\mb{\varepsilon}_{v_i}: V_i \in \mb{V}\}$ are mutually independent:
    \begin{align}
    \label{thm:alg-complete-3}
        V_i \ \longleftarrow \mspace{-25mu} \sum_{V_j \in \Pa(V_i, \g[D]^{1'})} \mspace{-25mu} A_{ij} V_j + \varepsilon_{v_i}.
    \end{align}
    Each coefficient $A_{ij}$ in this linear combination is defined by the edge coefficient of each $V_i \gets V_j$ in $\g[D]^{1'}$. We pick these edge coefficients in conjunction with $\{\mb{\sigma}^2_{v_i}: V_i \in \mb{V}\}$ in such a way that each $A_{ij} \in (0,1)$ and $\Var(V_i)=1$. 

    We build $\mb{F^*_1}$ by letting $f_1(\mb{v})$ be the density of the multivariate normal generated by the SEM in \eqref{thm:alg-complete-3}. For the remaining densities in $\mb{F^*_1}$, we let $f_1(\mb{v} \,|\, do(\mb{x'})) := f_*(\mb{v})$, where $f_*$ is  the density of the multivariate normal generated by taking the SEM in \eqref{thm:alg-complete-3} and replacing $\mb{X'}$ with its interventional value $\mb{x'}$ \citep{pearl2009causality}. Note that $\g[D]^{1'}$ is compatible with $\mb{F^*_1}$ by construction---and therefore, so is $\g[D]^1$ (Lemma \ref{lem:markov-sub}).

    To construct $\mb{F^*_2}$, define a linear SEM identical to the SEM in \eqref{thm:alg-complete-3} but based on the parents in $\g[D]^{2'}$. Let $f_2(\mb{v})$ be the density of the resulting distribution, which is another multivariate normal with mean vector zero and a covariance matrix with ones on the diagonal. We show below that the off-diagonal entries---which house the variables' covariances---are identical to the covariances from the SEM in \eqref{thm:alg-complete-3}. Thus, in an analogous way to $\mb{F^*_1}$, we can define $\mb{F^*_2} = \{ f_2(\mb{v}|do(\mb{x'})) : \mb{X'} \subseteq \mb{V} \}$ such that $\g[D]^{2'}$---and therefore $\g[D]^2$---is compatible with $\mb{F^*_2}$ and such that $f_2(\mb{v}) = f_1(\mb{v})$.

    To see that the covariances are identical, note that by Lemma \ref{lem:wright}, the covariances based on $\g[D]' \in \{ \g[D]^{1'}, \g[D]^{2'} \}$ will equal the product of edge coefficients along non-collider paths in $\g[D]'$. Since $\g[D]^{1'}$ and $\g[D]^{2'}$ have identical adjacencies, we only need to show that these DAGs have an identical set of colliders. Let $q'_1, \dots, q'_m$, and $p'$ be the paths in $\g[D]'$ corresponding to $\overline{p}$ and $\overline{q_1}, \dots, \overline{q_m}$ in $\g_{\overline{\mb{X} \setminus \{X\}} \underline{X}}$. Recall that $\mb{C}$ is the set of all colliders on $\overline{p}$. Then note that $\mb{C}$ is also the set of all colliders on $p'$, since $P_1, P_3, \dots, P_k$ are nodes of definite status on $p$ in $\g$ and since $P_2$ is always a non-collider on $p'$. Further, by Lemma \ref{lem:collider-paths}, $C_i$ is the only node on both $p'$ and $q_i'$, $i \in \{1, \dots, m\}$, and there is no shared node between the causal paths $q_i'$ and $q_j'$, $i \neq j$. Therefore, $\mb{C}$ is the set of all colliders on any path in either $\g[D]^{1'}$ or $\g[D]^{2'}$.

    Thus we have two families of interventional densities such that $\g[D]^i$ is compatible with $\mb{F^*_i}$ and such that $f_1(\mb{v})=f_2(\mb{v})$ so that \eqref{thm:alg-complete-1} holds. To complete Case 1, we show that \eqref{thm:alg-complete-2} holds. It suffices to show that $E[Y \,|\, do(\mb{X}=\mb{1}), \mb{Z}=\mb{0}]$ is not the same under $f_1$ and $f_2$. We start by calculating this expectation under $f_2$:
    \begin{align*}
        \E_{f_2}[Y \,|\, do(&\mb{X}=\mb{1}), \mb{Z} = \mb{0}]\\
            &\,=\, \left. \E_{f_2}[Y \,|\, Z_1, \dots, Z_m] \,\, \right\rvert_{\, \mb{Z}=\mb{0}}\\
            &\,=\, \left. \begin{bmatrix}
                                \Cov_{f_2}(Y, Z_1) \,\, \cdots \,\, \Cov_{f_2}(Y, Z_m) 
                          \end{bmatrix}
                          \, \mb{\Sigma}^{-1}
                          \,\, \mb{Z} \,\, \right\rvert_{\, \mb{Z}=\mb{0}}\\
            &\,=\, 0,
    \end{align*}
    where $\mb{\Sigma}$ is the covariance matrix of $\mb{Z}$ under $f_2$ (and where $\mb{\Sigma}$ is invertible since $\mb{V}$ is non-degenerate). The first equality follows from Rule 3 of Pearl's do calculus (Theorem \ref{thm:do-calc}), since $Y \dsepp \mb{X} \,|\, \mb{Z}$ in $\g[D]^{2'}_{\overline{\mb{X}}}$ and since $f_2$ is consistent with $\g[D]^{2'}$. The second equality follows from properties of multivariate normals (see Lemma \ref{lem:mardia}).

    Next, we calculate the same expectation under $f_1$:
    \begin{align*}
        \E_{f_1}[Y \,|\, do(&\mb{X}= \mb{1}), \mb{Z} = \mb{0}]\\
            &\,=\, \left. \E_{f_1}[Y \,|\, X, Z_1, \dots, Z_m] \,\, \right\rvert_{\, X=1, \, \mb{Z}=\mb{0}}\\
            &\,=\, \left. \begin{bmatrix}
                                \Cov_{f_1}(Y, X) & \Cov_{f_1}(Y,Z_1) \,\, \cdots \,\, \Cov_{f_1}(Y, Z_m) 
                          \end{bmatrix} \,
                          \mb{\Sigma}^{-1} \,\, 
                          \begin{bmatrix}
                                X \\
                                \mathbf{Z}
                          \end{bmatrix} \,\, \right\rvert_{\, X=1, \, \mb{Z}=\mb{0}}\\
            &\,=\, \begin{bmatrix}
                                0 \,\, \cdots \,\, 0 & \Cov_{f_1}(Y, Z_m) 
                          \end{bmatrix} \,
                          \mb{\Sigma}_1^{-1} \,\, 
                          \begin{bmatrix}
                                1 \\
                                \mathbf{0}
                          \end{bmatrix}\\
            &\,=\, \Sigma_{m+1,1} \cdot \Cov_{f_1}(Y, Z_m) \,\,,
    \end{align*} 
    where $\mb{\Sigma}$ is the invertible covariance matrix of $(X, Z_1, \dots, Z_m)^T$ under $f_1$ and where $\Sigma_{m+1,1}$ is the $(m+1,1)^{th}$ entry of $\mb{\Sigma}^{-1}$. The first equality follows from Rules 3 and 2 of Pearl's do calculus (Theorem \ref{thm:do-calc}), since $Y \dsepp \mb{X} \setminus \{X\} \,|\, X, \mb{Z}$ in $\g[D]^{1'}_{\overline{\mb{X}}}$, since $Y \dsepp X \,|\, \mb{Z}$ in $\g[D]^{1'}_{\underline{X}}$, and since $f_1$ is consistent with $\g[D]^{1'}$. The second equality follows from properties of multivariate normals (see Lemma \ref{lem:mardia}), and the third follows from applying Wright's Rule (Lemma \ref{lem:wright}) to $\g[D]^{1'}$.

    To complete Case 1, we show this expectation is non-zero so that \eqref{thm:alg-complete-2} holds. Note that $\Cov_{f_1}(Y, Z_m) \neq 0$, since by Lemma \ref{lem:wright}, it equals the product of all edge coefficients on $-(q'_m) \oplus p'(C_m,Y)$ in $\g[D]^{1'}$, which were chosen to be in $(0,1)$. To show that $\Sigma_{m+1,1} \neq 0$, we find the following.
    \begin{align*}
    \mb{\Sigma} \,=\, &
    \begin{blockarray}{ccccc}
        \begin{block}{[ccccc]}
          1                 & \Cov_{f_1}(X,Z_1)   &        &   & \bigzero \\
          \Cov_{f_1}(X,Z_1) & 1                   & \ddots &   & \\
      	                    & \Cov_{f_1}(Z_1,Z_2) & \ddots &   & \\
          \bigzero          &	                    & \ddots & 1 & \Cov_{f_1}(Z_{m-1}, Z_m) \\
                            &	                  &        & \Cov_{f_1}(Z_{m-1}, Z_m) & 1\\
        \end{block}
    \end{blockarray}.\\
        \Sigma_{m+1,1} &\,=\, \frac{\,\, (-1)^{m+2} \,\,}{\Det(\mb{\Sigma})} \cdot \Cov_{f_1}(X,Z_1) \cdot \Cov_{f_1}(Z_1,Z_2) \,\cdots\, \Cov_{f_1}(Z_{m-1},Z_{m}) \\
                  &\,\neq\, 0.
    \end{align*}
    The final equality follows by applying Wright's Rule (Lemma \ref{lem:wright}) to $\g[D]^{1'}$ and noting that the paths between $X$ and $Z_1$ as well as $Z_i$ and $Z_{i+1}$, $i \in \{1, \dots, m-1\}$, have edge coefficients in $(0,1)$ and no colliders.

    \vskip .2in

    \noindent
    \textbf{CASE 2:}
    Consider when $\mb{S_3} = \emptyset$, and define the following sets.
    \begin{align*}
        \mb{S_6} &= \{ \text{paths in $\mb{S_2}$ with the fewest edges} \}.\\
        \mb{S_7} &= \{ \text{paths in $\mb{S_6}$ with a shortest distance to $\mb{Z}$ in $\g_{\overline{\mb{X} \setminus \{X\}} \underline{X}}$} \}.
    \end{align*}
    Pick an arbitrary path $\overline{p} := \langle X = P_1, \dots, P_k = Y \rangle$, $k > 1$, $Y \in \mb{Y}$, in $\mb{S_7}$. Note that $\overline{p}$ begins $X \gets P_2$, since $\overline{p} \in \mb{S_2}$ is a path in $\g_{\overline{\mb{X} \setminus \{X\}} \underline{X}}$ and $\mb{S_3} = \emptyset$. Further, $\overline{p}$ is a proper path from $\mb{X}$ to $\mb{Y}$ that is open given $\mb{Z}$, since we have shown this holds for all paths in $\mb{S_1}$. Let $\mb{C} = \{C_1, \dots, C_m \}$, $1 \le m \le k-3$, be the set of all colliders on $\overline{p}$ in order of their appearance on $\overline{p}$, where possibly $\mb{C} = \emptyset$. Note that every node in $\mb{C}$ is in $\An(\mb{Z},\g_{\overline{\mb{X} \setminus \{X\}} \underline{X}})$, since $\overline{p}$ is open given $\mb{Z}$. Then let $\overline{q_i} = \langle Q_{i_1}, \dots, Q_{i_{k_i}} \rangle$, $i \in \{1, \dots m\}$, be a shortest causal path from $C_i$ to $\mb{Z}$ in $\g_{\overline{\mb{X} \setminus \{X\}} \underline{X}}$. Let $p$, $q_1, \dots, q_m$ be the paths in $\g$ corresponding to $\overline{p}$, $\overline{q_1}, \dots, \overline{q_m}$ in $\g_{\overline{\mb{X} \setminus \{X\}} \underline{X}}$.

    Next, recall from the beginning of the proof that there is a path in $\g$ from $X$ to $\mb{Y} \cup \mb{Z}$ that does not contain $\mb{X} \setminus \{X\}$, is possibly causal, and starts undirected. Let $r = \langle X = R_1, \dots, R_n \rangle, n >1$, be a shortest such path. When $r$ ends in $\mb{Y}$, it is a proper, possibly causal path from $\mb{X}$ to $\mb{Y}$ that starts undirected and does not contain nodes in $\mb{Z}$. It follows by Proposition \ref{prop:id-necessary} that the conditional effect of $\mb{X}$ on $\mb{Y}$ given $\mb{Z}$ is not identifiable in $\g$, and we are done. Hence, suppose $r$ ends in $\mb{Z}$ for the remainder of this proof.

    Using these paths, we turn to the task of finding DAGs in $[\g]$ that are compatible with families of interventional densities such that \eqref{thm:alg-complete-1}-\eqref{thm:alg-complete-2} hold. We start by considering a restriction of the class $[\g]$ using an MPDAG $\g^*$. When $X \notin \Adj(R_3, \g)$, then let $\g^* = \g$. When $X \in \Adj(R_3, \g)$, note that $\g$ cannot contain $X \gets R_3$, since $r$ is possibly causal. Neither can $\g$ contain $X - R_3$, since $\langle X, R_3 \rangle \oplus r(R_3,R_n)$ would contradict the definition of $r$ as a shortest proper possibly causal path in $\g$ from $X$ to $\mb{Z}$ that starts undirected. Thus, $\g$ contains $X \to R_3$. In this case, construct the MPDAG $\g^*$ by adding $R_2 \to R_3$ to $\g$ (if it does not already exist) and completing \citeauthor{meek1995causal}'s (\citeyear{meek1995causal}) orientation rules. Note by Lemma \ref{lem:adding-edges-imply} that $\g^*$ still contains $X - R_2$, since $X,R_2 \in \Pa(R_3,\g^*)$.

    With $\g^*$ defined, we let $\g[D]^1$ be a DAG in $[\g^*] \subseteq [\g]$ that contains $X \to R_2$. To see that the path in $\g[D]^1$ corresponding to $r$ in $\g$ takes the form $X \to R_2 \to \dots \to R_n$, note that $r(R_2,R_n)$ is unshielded by the definition of $r$ as a shortest possibly causal path. So when $X \notin \Adj(R_3, \g)$, the claim holds by R1 of \cite{meek1995causal} since we formed $\g[D]^1$ by adding $X \to R_2$ to $\g$. When $X \in \Adj(R_3, \g)$, then the same logic holds since we formed $\g[D]^1$ by adding $X \to R_2 \to R_3$ to $\g$. Further, note that by Lemma \ref{lem:collider-paths2}, $\g$---and therefore $\g[D]^1$---contains the edge $\langle P_2, R_2 \rangle$. Since $\g[D]^1$ is acyclic and has the path $P_2 \to X \to R_2$, then it contains $P_2 \to R_2$ so that $P_2$ is a non-collider on $\langle P_3, P_2, R_2 \rangle$ in $\g[D]^1$.

    Similarly, we let $\g[D]^2$ be a DAG in $[\g^*] \subseteq [\g]$ that contains $X \gets R_2 \to R_3$. (This DAG clearly exists when $\g^*$ contains $X - R_2 \to R_3$. When instead $\g^*$ contains $X - R_2 - R_3$, this DAG exists by Lemma \ref{lem:undirected-imply}.) Then note that the path in $\g[D]^2$ corresponding to $r$ in $\g$ takes the form $X \gets R_2 \to \dots \to R_n$ by R1 of \cite{meek1995causal}, since $r(R_2,R_n)$ is unshielded and possibly causal. To see that $P_2$ is a non-collider on $\langle P_3, P_2, R_2 \rangle$ in $\g[D]^1$, note that by Lemma \ref{lem:collider-paths2}, $\g$ must contain $P_2 \to R_2$; or $P_3 \gets P_2$; or $P_3 - P_2 - R_2$ such that $P_3 \notin \Adj(R_2, \g)$.

    Now that we have chosen $\g[D]^1, \g[D]^2 \in [\g]$, we construct a family of interventional densities $\mb{F^*_i} = \{ f_i(\mb{v}|do(\mb{x'})) : \mb{X'} \subseteq \mb{V} \}$ compatible with each DAG. Start by constructing a DAG $\g[D]^{1'}$ by removing every edge from $\g[D]^1$ except those on the paths corresponding to $p, q_1, \dots, q_m$, $r$, and $\langle P_2, R_2 \rangle$. Then consider the following linear SEM. Each random variable $V_i \in \mb{V}$ is a linear combination of its parents in $\g[D]^{1'}$ and $\varepsilon_{v_i} \sim N(0, \sigma^2_{v_i})$, where $\{\mb{\varepsilon}_{v_i}: V_i \in \mb{V}\}$ are mutually independent; where the coefficients in the SEM are all $\tfrac{1}{2}$; and where we choose $\sigma^2_{v_i}$ such that $\Var(V_i)=1$ for all $V_i \in \mb{V}$:
    \begin{align}
    \label{thm:alg-complete-4}
        V_i \ \longleftarrow \mspace{-25mu} \sum_{V_j \in \Pa(V_i, \g[D]^{1'})} \mspace{-25mu} \tfrac{1}{2} V_j + \varepsilon_{v_i}.
    \end{align}
    We build $\mb{F^*_1}$ by letting $f_1(\mb{v})$ be the density of the multivariate normal generated by this SEM. For the remaining densities in $\mb{F^*_1}$, we let $f_1(\mb{v} \,|\, do(\mb{x'})) := f_*(\mb{v})$, where $f_*$ is  the density of the multivariate normal generated by taking the SEM in \eqref{thm:alg-complete-4} and replacing $\mb{X'}$ with its interventional value $\mb{x'}$ \citep{pearl2009causality}. Note that $\g[D]^{1'}$ is compatible with $\mb{F^*_1}$ by construction---and therefore, so is $\g[D]^1$ (Lemma \ref{lem:markov-sub}).

    To construct $\mb{F^*_2}$ consider the following linear SEM. Each random variable $V_i \in \mb{V}$ is a linear combination of its parents in $\g[D]^{2'}$ and $\varepsilon_{v_i} \sim N(0, \sigma^2_{v_i})$, where $\{\mb{\varepsilon}_{v_i}: V_i \in \mb{V}\}$ are mutually independent:
    \begin{align}
    \label{thm:alg-complete-5}
        V_i \ &\longleftarrow \mspace{-25mu} \sum_{V_j \in \Pa(V_i, \g[D]^{2'})} \mspace{-25mu} A_{ij} V_j + \varepsilon_{v_i}.
    \end{align}
    Each coefficient $A_{ij}$ in this linear combination is defined by the edge coefficient of $\langle V_i, V_j \rangle$ in $\g[D]^{2'}$. We set the edge coefficients for $\langle X,P_2 \rangle$, $\langle X,R_2 \rangle$, $\langle P_2,R_2 \rangle$ to $-\tfrac{1}{7}$, $\tfrac{6}{7}$, $\tfrac{3}{4}$, respectively, and we set the remaining $A_{ij} = \tfrac{1}{2}$. Further, we choose $\{\mb{\sigma}^2_{v_i}: V_i \in \mb{V}\}$ so that $\Var(V_i)=1$. Then let $f_2(\mb{v})$ be the density of the resulting distribution, which is another multivariate normal with mean vector zero and a covariance matrix with ones on the diagonal. We show below that the off-diagonal entries---which house the variables' covariances---are identical to the covariances from the SEM in \eqref{thm:alg-complete-4}. Thus, in an analogous way to $\mb{F^*_1}$, we can define $\mb{F^*_2} = \{ f_2(\mb{v}|do(\mb{x'})) : \mb{X'} \subseteq \mb{V} \}$ such that $\g[D]^{2'}$---and therefore $\g[D]^2$---is compatible with $\mb{F^*_2}$ and such that $f_2(\mb{v}) = f_1(\mb{v})$.
    
    To see that the covariances are identical, note that by Lemma \ref{lem:wright}, the covariances based on $\g[D]' \in \{ \g[D]^{1'}, \g[D]^{2'} \}$ will equal the product of edge coefficients along non-collider paths in $\g[D]'$. Consider the paths in $\g[D]'$ and their colliders. By Lemmas \ref{lem:collider-paths}-\ref{lem:collider-paths2}, the paths in $\g[D]'$ corresponding to $p$, $q_1, \dots, q_m$, $r$ do not overlap anywhere except at $X$ and $C_i$, $i \in \{1, \dots, m\}$. Then note that $\g[D]'$ contains no colliders on $q_1, \dots, q_m$, $\langle R_1, \dots, R_n \rangle$, or $\langle P_3, P_2, R_2, R_3 \rangle$. To see that that $\mb{C}$ is the set of all colliders on $\langle P_1, \dots, P_k \rangle$, note that $p$ is of definite status in $\g$, since $\overline{p} \in \mb{S_1}$ is a definite status path in $\g_{\overline{\mb{X} \setminus \{X\}} \underline{X}}$ that begins $X \gets P_2$ and has no nodes in $\mb{X} \setminus \{X\}$. However, note that $\g[D]^{1'}$ contains $X \to R_2 \gets P_2$, whereas $\g[D]^{2'}$ contains $P_2 \to X \gets R_2$. Therefore, the claim that the covariances from \eqref{thm:alg-complete-4} and \eqref{thm:alg-complete-5} are identical holds after calculating the following for both models: $\Cov(P_2,X)=\tfrac{1}{2}$, $\Cov(P_2,R_2)=\tfrac{3}{4}$, $\Cov(P_2,R_3)=\tfrac{3}{8}$, $\Cov(X,R_2)=\tfrac{3}{4}$.

    Thus we have two families of interventional densities such that $\g[D]^i$ is compatible with $\mb{F^*_i}$ and such that \eqref{thm:alg-complete-1} holds. To complete Case 2, we show that \eqref{thm:alg-complete-2} holds. It suffices to show that $E[Y \,|\, do(\mb{X}=\mb{1}), \mb{Z}=\mb{0}]$ is not the same under $f_1$ and $f_2$. We start by calculating this expectation under $f_2$. For ease of notation, let $Z_0=R_n$ and let $Z_i = Q_{i_{k_i}}$ for all $i \in \{1, \dots, m\}$.
    \begin{align*}
        \E_{f_2}[Y \,|\, do(&\mb{X}=\mb{1}), \mb{Z} = \mb{0}]\\[.1cm]
            &\,=\, \left. \E_{f_2}[Y \,|\, Z_0, \dots, Z_m] \,\, \right\rvert_{\, \{Z_0, \dots, Z_m\}=\mb{0}}\\[-.4cm]
            &\,=\, \left. \begin{bmatrix}
                                \Cov_{f_2}(Y, Z_0) \,\, \cdots \,\, \Cov_{f_2}(Y, Z_m) 
                          \end{bmatrix} 
                          \, \mb{\Sigma}^{-1}
                          \begin{bmatrix}
                                Z_0 \\ \vdots \\ Z_m
                          \end{bmatrix} 
                          \,\, \right\rvert_{\, \{Z_0, \dots, Z_m\}=\mb{0}}\\[-.5cm]
            &\,=\, 0,
    \end{align*}
    where $\mb{\Sigma}$ is the covariance matrix of $\{Z_0, \dots, Z_m\}^T$ under $f_2$ (and where $\mb{\Sigma}$ is invertible since $\mb{V}$ is non-degenerate). The first equality follows from Rules 3 and 1 of Pearl's do calculus (Theorem \ref{thm:do-calc}), since $Y \dsepp \mb{X} \,|\, \mb{Z}$ in $\g[D]^{2'}_{\overline{\mb{X}}}$; since $Y \dsepp \mb{Z} \setminus \{Z_0, \dots, Z_m\} \,|\, Z_0, \dots, Z_m$ in $\g[D]^{2'}$; and since $f_2$ is consistent with $\g[D]^{2'}$. The second equality follows from properties of multivariate normals (see Lemma \ref{lem:mardia}).

    Next, we calculate the same expectation under $f_1$. In each of the subcases below, we show that $\E_{f_1}[Y | do(\mb{X}= \mb{1}), \mb{Z} = \mb{0}]$ is non-zero so that \eqref{thm:alg-complete-2} holds. For these calculations, note that $f_1(y \,|\, do(\mb{x}), \mb{z} )$ is the conditional density of $Y \,|\, \mb{Z}$ under $f_*(\mb{v})$.
    \begin{itemize}
        \item \textbf{SUBCASE 1:}
        When $p'$ has no colliders in $\g[D]^{1'}$,
        \begin{align*}
            \E_{f_1}[Y \,|\, do(\mb{X}= \mb{1}), \mb{Z}= \mb{0}] 
                    &\,=\, \left. \E_{f_*}[Y \,|\, \mb{Z}] \, \right\rvert_{\, \mb{x}=\mb{1}, \, \mb{Z}=\mb{0}}\\
                    &\,=\,  \left. \E_{f_*}[Y \,|\, Z_0] \, \right\rvert_{\, x=1, \, Z_0=0}\\
                    &\,=\,  \left. \E_{f_*}[Y] + \frac{\Cov_{f_*}(Y,Z_0)}{\Var_{f_*}(Z_0)} \cdot \Big(Z_0 - \E_{f_*}[Z_0]\Big) \, \right\rvert_{\, x=1, \, Z_0=0}\\
                    &\,=\,  \frac{\Cov_{f_*}(Y,Z_0)}{\Var_{f_*}(Z_0)} \cdot \big(- \tfrac{1}{2}^{n -1}\big) \\
                    &\,\neq\, 0.
        \end{align*}
        The third equality follows from properties of multivariate normals (see Lemma \ref{lem:mardia}). For the final equality, consider Lemma \ref{lem:sem-covariance}, and note that the SEM that defines $f_*(\mb{v})$ under $do(\mb{X})$ has a matrix of coefficients that are non-zero exactly when $\g[D]^{1'}_{\overline{\mb{X}}}$ contains an edge. It follows that $\Cov_{f_*}(Y,Z_0) \neq 0$, since $\g[D]^{1'}_{\overline{\mb{X}}}$ contains a node in $\{P_2, \dots, P_k\}$ with causal paths to $Y$ and $Z_0$.

        \item \textbf{SUBCASE 2:}
        When $p'$ has at least one collider,
        \begin{align*}
            \E_{f_1}[Y &\,|\, do(\mb{X}= \mb{1}), \mb{Z}= \mb{0}]\\[.1cm]
                    &\,=\, \left. \E_{f_*}[Y \,|\, \mb{Z}] \, \right\rvert_{\, \mb{x}=\mb{1}, \, \mb{Z}=\mb{0}}\\[.1cm]
                    &\,=\, \left. \E_{f_*}[Y \,|\, Z_0, \dots, Z_m] \, \right\rvert_{\, x=1, \, \{Z_0, \dots, Z_m\}=\mb{0}}\\[-.4cm]
                    &\,=\, \left. \E_{f_*}[Y] + 
                                  \begin{bmatrix}
                                        \Cov_{f_*}(Y,Z_0) \,\, \cdots \,\, \Cov_{f_*}(Y, Z_m) 
                                  \end{bmatrix} \,
                                        \mb{\Sigma}^{-1} \,\, 
                                  \begin{bmatrix}
                                        Z_0-\E_{f_*}[Z_0]\\
                                        \vdots\\
                                        Z_m-\E_{f_*}[Z_m]
                                  \end{bmatrix} \,\, \right|_{\, \begin{subarray}{l} \,\, x=1, \\ \{Z_0, \dots, Z_m\}=\mb{0}\end{subarray}}\\[-.5cm]
                    &\,=\, \begin{bmatrix}
                                0 \,\, \cdots \,\, 0 & \Cov_{f_*}(Y, Z_m)
                            \end{bmatrix}\,
                                \mb{\Sigma}^{-1} \,\,
                            \begin{bmatrix}
                                -\E_{f_*}[Z_0] \\
                                \mathbf{0}
                            \end{bmatrix}\\
                    &= \tfrac{1}{2}^{k_m'} \cdot \Sigma_{m+1,1} \cdot \big(- \tfrac{1}{2}^{n -1}\big),
        \end{align*}      
        where $k_m'$ is the number of edges on $(-q_m')(Z_m,C_m) \oplus p'(C_m,Y)$; where $\mb{\Sigma}$ is the invertible covariance matrix of $(Z_0, \dots, Z_m)^T$ under $f_*$; and where $\Sigma_{m+1,1}$ is the $(m+1,1)^{th}$ entry of $\mb{\Sigma}^{-1}$. The third equality follows from properties of multivariate normals (see Lemma \ref{lem:mardia}). The last two equalities follow by Lemma \ref{lem:sem-covariance}, since $\g[D]^{1'}_{\overline{\mb{X}}}$ contains no node with causal paths to $Y$ and $Z_i$, $i \in \{0, \dots, m-1\}$, but $\g[D]^{1'}_{\overline{\mb{X}}}$ does contain a node with causal paths to $Y$ and $Z_0$.

        To complete Subcase 2, we show that $\Sigma_{m+1,1} \neq 0$.        
        \begin{align*}
        \mb{\Sigma} \,=\, &
        \begin{blockarray}{cccccc}
        \begin{block}{[cccccc]}
            1   			       & \Cov_{f_*}(Z_0,Z_1) & & & & \\
      	    \Cov_{f_*}(Z_0,Z_1) & 1                      & \ddots & & & \bigzero \\
                                   & \Cov_{f_*}(Z_1,Z_2) & \ddots & & &	& \\
           	\bigzero               &                        & \ddots & & \Cov_{f_*}(Z_{m-1}, Z_m) 	& & \\
                                   & & & & 1                     & \Cov_{f_*}(Z_m, Y) \\
                                   & & & & \Cov_{f_*}(Z_m, Y) & 1 \\
        \end{block}
        \end{blockarray}.\\
        \Sigma_{m+1,1} &= \frac{(-1)^{m+2}}{\Det(\mb{\Sigma})} \Cov_{f_*}(Z_0,Z_1) \cdots \Cov_{f_*}(Z_{m-1},Z_m) \\
        &\neq 0.
        \end{align*}
        The final equality follows from Lemma \ref{lem:sem-covariance}, since for each $i \in \{0, \dots, m-1\}$, $\g[D]^{1'}_{\overline{\mb{X}}}$ contains a node in $\{P_2, \dots, P_k\}$ with causal paths to $Z_i$ and $Z_{i+1}$.
    \hfill \BlackBox
    \end{itemize}
\end{proofofnoqed}


\subsection{Supporting Results}


\begin{lemma}
\label{lem:helper-for-annoying-case}
    Let $p = \langle V_1, \dots, V_k \rangle, k> 1$, be a definite status path in $\g = (\mb{V, E})$, where $\g$ is either a causal MPDAG or there exists a causal MPDAG $\g^{'}$ and node sets $\mb{A} \subseteq \mb{V}$, and $\mb{B} \subseteq \mb{V} \setminus \mb{A}$ such that $\g \equiv \g^{'}_{\overline{A} \underline{B}}$. Further, let $\mb{C} \subseteq \mb{V}$.

    Suppose $p$ is d-connecting given $\mb{C}$ in $\g$ and suppose $\g$ contains an edge $\langle V_i, V_j \rangle$, $1 \le i < j \le k$, such that $t = p(V_1, \dots, V_i) \oplus \langle V_i, V_j \rangle \oplus p(V_j, V_k)$ is also a definite status path in $\g$. If $V_i$ and $V_j$ are definite non-colliders or endpoints on $p$, then $t$ is also d-connecting given $\mb{C}$.
\end{lemma}

\begin{proofof}[Lemma \ref{lem:helper-for-annoying-case}]
    If $V_i$ and $V_j$ are still definite non-colliders (or endpoints) on $t$, the statement clearly holds. Hence, suppose that one of $V_i$ or $V_j$ is a collider on $t$. Without loss of generality, we will assume that it is $V_i$ that is a collider on $t$. Then $V_{i-1} \to V_i \to V_{i+1}$ must be on $p$, whereas $V_{i-1} \to V_i \gets V_j$ is on $\g$. 

    Consider $p(V_i, V_j)$. This path cannot take the form $V_i \to \dots \to V_j$ by the acyclicity of $\g'$. Further, $p$ cannot contain $V_l \to V_{l+1} - V_{l+1}$, it is of definite status. Hence, there must be a collider on $p(V_i, V_j)$ in $\De(V_i,\g)$. Since this collider is in $\An (\mb{C}, \g)$, it also holds that $V_i \in \An(\mb{C}, \g)$. Therefore, $t$ is still d-connecting given $\mb{C}$.
\end{proofof}


\begin{lemma}
\label{lem:unishielded-colliders}
    Let $\mb{X,Y,Z}$ be pairwise disjoint node sets in a causal MPDAG $\g = (\mb{V,E})$. Suppose there is no possibly causal path in $\g$ from $X$ to $\mb{Y}$ that starts undirected and does not contain $\mb{X} \setminus \{X\} \cup \mb{Z}$. Further, suppose there is a possibly causal path in $\g$ from $X$ to $\mb{Z}$ that starts undirected and does not contain $\mb{X} \setminus \{X\} \cup \mb{Y}$, where $(X \not \dsepp \mb{Y} | \mb{X} \setminus \{X\}, \mb{Z})_{\g_{\overline{\mb{X} \setminus \{X\}} \underline{X}}}$. Then consider the following sets.
    \begin{align*}
        \mb{S_1} &= \left\{\begin{array}{ll}
                                \text{definite status paths in $\g_{\overline{\mb{X} \setminus \{X\}} \underline{X}}$ from $X$ to $\mb{Y}$ that are} \\
                                \text{d-connecting given $\mb{X} \setminus \{X\} \cup \mb{Z}$}
                           \end{array}\right\}.\\
        \mb{S_2} &= \{ \text{paths in $\mb{S_1}$ with the fewest colliders} \}.\\[.4cm]
        \mb{S_3} &= \{ \text{paths in } \mb{S_2} \text{ that start undirected}\}.\\
        \mb{S_4} &= \{ \text{paths in $\mb{S_3}$ with the fewest edges} \}.\\
        \mb{S_5} &= \{ \text{paths in $\mb{S_4}$ with a shortest distance to $\mb{Z}$ in $\g_{\overline{\mb{X} \setminus \{X\}} \underline{X}}$} \}.\\[.4cm]
        \mb{S_6} &= \{ \text{paths in $\mb{S_2}$ with the fewest edges} \}.\\
        \mb{S_7} &= \{ \text{paths in $\mb{S_6}$ with a shortest distance to $\mb{Z}$ in $\g_{\overline{\mb{X} \setminus \{X\}} \underline{X}}$} \}.
    \end{align*}
    When $\mb{S_3} \neq \emptyset$, pick an arbitrary path in $\mb{S_5}$. When $\mb{S_3} = \emptyset$, pick an arbitrary path in $\mb{S_7}$. Call this path $p := \langle X = P_1, \dots, P_k = Y \rangle$, $k > 1$, $Y \in \mb{Y}$, and let $p^*$ be the path in $\g$ corresponding to $p$ in $\g_{\overline{\mb{X} \setminus \{X\}} \underline{X}}$. Then the following hold.
    \begin{enumerate}[label = (\roman*)]
        \item\label{unishielded-coll-case0} $p$ contains no nodes in $\mb{X} \setminus \{X\}$ or $\mb{Y} \setminus \{Y\}$.
        \item \label{unishielded-coll-case1} $\g$ does not contain $X - P_i$ for any $i > 2$.
        \item \label{unishielded-coll-case2} Every collider on $p$ or $p^*$ is unshielded in $\g_{\overline{\mb{X} \setminus \{X\}} \underline{X}}$ or $\g$, respectively. 
    \end{enumerate}
\end{lemma}

\begin{proofofnoqed}[Lemma \ref{lem:unishielded-colliders}]
    \begin{enumerate}[label = (\roman*)]
        \item[\ref{unishielded-coll-case0}] Since $p$ is a d-connecting given $\mb{X} \setminus \{X\} \cup \mb{Z}$, no definite non-collider on $p$ is in $\mb{X} \setminus \{X\}$. Also, since all edges into $\mb{X} \setminus \{X\}$ are removed in $\g_{\overline{\mb{X} \setminus \{X\}} \underline{X}}$, no collider on $p$ is in $\mb{X} \setminus \{X\}$. To show that $P_k = Y$ is the only node from $\mb{Y}$ on $p$, it is enough to notice that otherwise, for $P_l \in \mb{Y}, l \in \{2, \dots, k-1\}$ we could choose $p(X, P_l)$ instead of $p$.

        \item[\ref{unishielded-coll-case1}] We show that $X - P_i$ is not in $\g$ for any $i > 2$ by contradiction. Hence, let $j \in \{3, \dots, k\}$ be the largest index such that $X - P_j$ is in $\g$. Now, $j \neq k$ since otherwise, $X - Y$ is in $\g$, which contradicts our assumptions. 

        Next, suppose that $2< j < k$ and $X - P_j$ is in $\g$. If $P_j \to P_{j+1}$ is on $p$, we have a contradiction with the choice of $p$, since we could have chosen $\langle X, P_j \rangle \oplus p(P_j, Y)$. We get the same contradiction if $P_j - P_{j+1}$ is in $\g$ and $X \notin \Adj(P_{j+1}, \g_{\overline{\mb{X} \setminus \{X\}} \underline{X}})$. 

        Two cases are left to consider, the case when $P_j - P_{j+1}$ and $X \gets P_{j+1}$ are in $\g$ and $ \g_{\overline{\mb{X} \setminus \{X\}} \underline{X}}$, and the case when $P_j \gets P_{j+1}$ and $X \gets P_{j+1}$ are in $\g$ and $ \g_{\overline{\mb{X} \setminus \{X\}} \underline{X}}$ (by choice of $j$, we cannot have $X - P_{j+1}$ in $\g$). In both cases, $s = \langle X, P_{j+1} \rangle \oplus p(P_{j+1}, Y)$, $s \in \mb{S_1}$ (as $P_{j+1}$ is a definite non-collider on both $p$ and $s$). Furthermore, $s$ cannot have more colliders than $p$, so $s \in \mb{S_2}$. 

        Suppose next that $\mb{S_3} = \emptyset$, that is $p \in \mb{S_7}$. Since $s$ has fewer edges than $p$, we have a contradiction with the choice of $p$.

        Otherwise, $\mb{S_3} \neq \emptyset$. Now, for $p$ to be in $\mb{S_5} \subseteq \mb{S_3} \subseteq \mb{S_2}$, we need to have that $p(X,P_{j+1})$ contains no colliders and that $p$ starts undirected. Furthermore, since $p$ is of definite status, we have that $P_{i-1} - P_i \gets P_{i+1}$, $i \in \{2, \dots, j\}$ cannot be on $p(X,P_{j+1})$. Combining the information that $P_{i} \to P_{i+1} \gets P_{i+2}$ and $P_{i} - P_{i+1} \gets P_{i+2}$ cannot be on $p(X,P_{j+1})$, for $i \in \{1, j-1\}$ and that $X - P_2$ is on $p(X,P_{j+1})$ leads us to conclude that $P_l \gets P_{l+1}$, $l \in \{1,\dots, j\}$ cannot be on $p(X,P_{j+1})$, either. 

        If $p^{*}$ is of definite status, then since $P_l \gets P_{l+1}$, $l \in \{1,\dots, j\}$ is not on $p^{*}(X,P_{j+1})$, we have that $p^{*}(X,P_{j+1})$ is possibly causal in $\g$ (Lemma \ref{lem:pc-def-stat}). Therefore, $X \gets P_{j+1}$ being in $\g$ is a contradiction. 

        Otherwise, $p^{*}$ is not of definite status in $\g$. This implies that $X \to P_{3}$ is in $\g$ (and not in $\g_{\overline{\mb{X} \setminus \{X\}} \underline{X}}$). Moreover, case \ref{unishielded-coll-case0} implies that $p^{*}(P_2, P_{j+1})$ is of definite status. Hence, since $P_l \gets P_{l+1}$, $l \in \{1,\dots, j\}$ is not on $p^{*}(X,P_{j+1})$, we have that $p^{*}(P_2,P_{j+1})$ is possibly causal in $\g$ (Lemma \ref{lem:pc-def-stat}). But then the fact that $p^{*}(P_3,P_{j+1})$ is possibly causal and $P_{j+1} \to X \to P_{3}$ is in $\g$, contradicts Lemma \ref{lem:nobackpaths}.

        \item[\ref{unishielded-coll-case2}]
        Next, we show that every collider on $p$ is unshielded. Hence, for the rest of the proof we will assume that there is at least one collider on $p$. By case \ref{unishielded-coll-case0}, $p$ does not contain any node in $\mb{X} \setminus \{X\}$. So for the colliders on $p^{*}$ to be unshielded, it is enough to show that colliders on $p$ are unshielded. Also, note that any path in $\mb{S_1}$ must be d-connecting given $\mb{Z}$ (since all edges into $\mb{X} \setminus \{X\}$ are removed in $\g_{\overline{\mb{X} \setminus \{X\}} \underline{X}}$).

        Let $\mb{C} = \{C_1, \dots, C_m \}$, $1 \le m \le k-3$, be the set of all colliders on $p$, labeled such that if $1 \le i< j \le m$, $C_i \equiv P_l$, $C_j \equiv P_r$, $2 \le l < r \le k-1$. Then each $C_i$, $i \in \{1, \dots, m\}$ is in $\An(\mb{Z},\g_{\overline{\mb{X} \setminus \{X\}} \underline{X}})$. Let $q_i = \langle C_i = Q_{i_1}, \dots, Q_{i_{k_i}} \rangle$, be a shortest directed path from $C_i$ to $\mb{Z}$ in $\g_{\overline{\mb{X} \setminus \{X\}} \underline{X}}$. Note that $Q_{i_{k_i}}$ is the only node in $\mb{Z}$ on $q_i$ and that $q_i$ cannot contain nodes in $\mb{X}$.

        Suppose for a contradiction that a collider $C_i$ on $p$ is shielded, that is for $C_i \equiv P_j$, $j \in \{3, \dots, k-1\}$, we have that $\langle P_{j-1}, P_{j+1} \rangle$ is in $\g_{\overline{\mb{X} \setminus \{X\}} \underline{X}}$. Then consider the path $s = p(X,P_{j-1}) \oplus \langle P_{j-1}, P_{j+1} \rangle \oplus p(P_{j+1}, Y)$ in $\g_{\overline{\mb{X} \setminus \{X\}} \underline{X}}$. If $s$ is of definite status in $\g_{\overline{\mb{X} \setminus \{X\}} \underline{X}}$, then by Lemma \ref{lem:helper-for-annoying-case}, $s$ must be d-connecting given $\mb{Z}$. Then since $s$ starts with the same edge as $p$, but is shorter than $p$, we have a contradiction with the choice of $p$.

        Otherwise, $s$ is not of definite status, meaning that $\langle P_{j-2}, P_{j+1} \rangle$ or $\langle P_{j-1}, P_{j+2} \rangle$, or both. are in $\g_{\overline{\mb{X} \setminus \{X\}} \underline{X}}$. We now split the proof into four parts depending on the position of collider $C_i$ on $p$ and the number of colliders on $p$.
        \begin{enumerate}[label = (\arabic*)]
            \item\label{case1:unshielded-collider}
            Let $i = 1 = k$. That means $C_i = P_j$ is the first and only collider on $p$. Hence, let $P_l, l\ge1$ be the closest node to $X$ on $p(X,P_{j-1})$ that is adjacent to a node on $p(P_{j+1},Y)$. Furthermore, let $P_r$, be the closest node to $Y$ on $p(P_{j+1}, Y)$ such that $\langle P_l, P_{r} \rangle$ is in $\g_{\overline{\mb{X} \setminus \{X\}} \underline{X}}$. Furthermore, consider the path $t = p(X,P_l) \oplus \langle P_l, P_r \rangle \oplus p(P_r, Y)$. Note that it is possible that $l =1$ and $r = k$, meaning it is possible that $p(X,P_l), p(P_r, Y)$ are of length zero.

            Then $t$ must be of definite status path by construction. Also, since neither $P_l$ nor $P_r$ are colliders on $p$, by Lemma \ref{lem:helper-for-annoying-case}, $t$ must also be d-connecting given $\mb{X} \setminus \{X\} \cup \mb{Z}$. Now, if $l = 1$, then $t$ starts with an edge into $X$ (by case \ref{unishielded-coll-case1}) and has fewer colliders than $p$, so we have a contradiction. If $l > 1$, then we again have a contradiction with the choice of $p$, because $t$ and $p$ start with the same edge, and $t$ either has fewer colliders than $p$, or the same number of colliders but fewer edges than $p$. 

            \item\label{case2:unshielded-collider}
            Let $i = 1 < k$. In this case, we assume $i = 1 < k$, so $C_i = P_j$ is the collider closest to $X$ on $p$, but not the only collider on $p$. We will now proceed similarly to the case \ref{case1:unshielded-collider}. 

            First, let $C_2 \equiv P_w$ for some $j +1<w<k$. Then, let $P_l$ be the closest node to $X$ on $p(X,P_{j-1})$ that is adjacent to a node on $p(P_{j+1},P_{w-1})$. Furthermore, let $P_r$, be the closest node to $P_{w-1}$ on $p(P_{j+1}, P_{w-1})$ such that $\langle P_l, P_{r} \rangle$ is in $\g_{\overline{\mb{X} \setminus \{X\}} \underline{X}}$. Furthermore, consider the path $t = p(X,P_l) \oplus \langle P_l, P_r \rangle \oplus p(P_r, Y)$. 

            Note that $t$ must be of definite status by construction, even in the case where $P_r \equiv P_{w-1}$, simply because $p(P_{w-1}, Y)$ is out of $P_{w-1}$. Furthermore, since both $P_l$ and $P_r$ are non-colliders on $p$, we have that $t$ is d-connecting given $\mb{X} \setminus \{X\} \cup \mb{Z}$ by Lemma \ref{lem:helper-for-annoying-case}. Then if $l =1$, $t$ starts with an edge into $X$ (by case \ref{unishielded-coll-case1}) and has one fewer collider than $p$, which leads us to a contradiction. Otherwise,  $l \neq 1$, and $t$ starts with the same edge as $p$, but $t$ either has fewer colliders than $p$, or the same number of colliders as $p$ but fewer edges than $p$ which gives us the desired contradiction. 

            \item\label{case3:unshielded-collider}
            Let $1 < i < k$. That is, $C_i$ is not the closest collider to $X$ or to $Y$ on $p$. Let $P_u = C_{i-1}$ and let $P_w = C_{i+1}$ and note that, $1 < u < j-1$ and $j +1< w <k$. Then, let $P_l$ be the closest node to $P_{u+1}$ on $p(P_{u+1},P_{j-1})$ that is adjacent to a node on $p(P_{j+1},P_{w-1})$. Furthermore, let $P_r$, be the closest node to $P_{w-1}$ on $p(P_{j+1}, P_{w-1})$ such that $\langle P_l, P_{r} \rangle$ is in $\g_{\overline{\mb{X} \setminus \{X\}} \underline{X}}$. Furthermore, consider the path $t = p(X,P_l) \oplus \langle P_l, P_r \rangle \oplus p(P_r, Y)$. 

            As in the cases above, $t$ must be of definite status by construction, even in the case where $P_r = P_{u+1}$, as $-p(P_{u+1}, X)$ is out of $P_{u+1}$ and even if $P_l \equiv P_{w-1}$, $p(P_{w-1}, Y)$ is out of $P_{w-1}$. Now, similarly to the above cases, we can show that $t$ must also be d-connecting given $\mb{X} \setminus \{X\} \cup \mb{Z}$ (Lemma \ref{lem:helper-for-annoying-case}). And similarly to the above cases $t$ then contradicts the choice of $p$.

            \item\label{case4:unshielded-collider}
            Let $1 < i = k$. That is, $C_i$ is the closest collider to $Y$ on $p$. Let $P_u = C_{i-1}$ and note that, $1 < u < j-1$. Then, let $P_l$ be the closest node to $P_{u+1}$ on $p(P_{u+1},P_{j-1})$ that is adjacent to a node on $p(P_{j+1},Y)$. Furthermore, let $P_r$, be the closest node to $Y$ on $p(P_{j+1}, Y)$ such that $\langle P_l, P_{r} \rangle$ is in $\g_{\overline{\mb{X} \setminus \{X\}} \underline{X}}$. Furthermore, consider the path $t = p(X,P_l) \oplus \langle P_l, P_r \rangle \oplus p(P_r, Y)$. Now, using Lemma \ref{lem:helper-for-annoying-case}, path $t$ leads us to a contradiction similar to above. \hfill \BlackBox
        \end{enumerate}
    \end{enumerate}
\end{proofofnoqed}


\begin{lemma}
\label{lem:collider-paths}
    Let $\mb{X,Y,Z}$ be pairwise disjoint node sets in a causal MPDAG $\g = (\mb{V,E})$. Suppose there is no possibly causal path in $\g$ from $X$ to $\mb{Y}$ that starts undirected and does not contain $\mb{X} \setminus \{X\} \cup \mb{Z}$. Further, suppose there is a possibly causal path in $\g$ from $X$ to $\mb{Z}$ that starts undirected and does not contain $\mb{X} \setminus \{X\} \cup \mb{Y}$, where $(X \not \dsepp \mb{Y} | \mb{X} \setminus \{X\}, \mb{Z})_{\g_{\overline{\mb{X} \setminus \{X\}} \underline{X}}}$. Then consider the following sets.
    \begin{align*}
        \mb{S_1} &= \left\{\begin{array}{ll}
                                \text{definite status paths in $\g_{\overline{\mb{X} \setminus \{X\}} \underline{X}}$ from $X$ to $\mb{Y}$ that are} \\
                                \text{d-connecting given $\mb{X} \setminus \{X\} \cup \mb{Z}$}
                           \end{array}\right\}.\\
        \mb{S_2} &= \{ \text{paths in $\mb{S_1}$ with the fewest colliders} \}.\\[.4cm]
        \mb{S_3} &= \{ \text{paths in } \mb{S_2} \text{ that start undirected}\}.\\
        \mb{S_4} &= \{ \text{paths in $\mb{S_3}$ with the fewest edges} \}.\\
        \mb{S_5} &= \{ \text{paths in $\mb{S_4}$ with a shortest distance to $\mb{Z}$ in $\g_{\overline{\mb{X} \setminus \{X\}} \underline{X}}$} \}.\\[.4cm]
        \mb{S_6} &= \{ \text{paths in $\mb{S_2}$ with the fewest edges} \}.\\
        \mb{S_7} &= \{ \text{paths in $\mb{S_6}$ with a shortest distance to $\mb{Z}$ in $\g_{\overline{\mb{X} \setminus \{X\}} \underline{X}}$} \}.
    \end{align*}
    When $\mb{S_3} \neq \emptyset$, pick an arbitrary path in $\mb{S_5}$. When $\mb{S_3} = \emptyset$, pick an arbitrary path in $\mb{S_7}$. Call this path $p := \langle X = P_1, \dots, P_k = Y \rangle$, $k > 1$, $Y \in \mb{Y}$. Let $\mb{C} = \{C_1, \dots, C_m \}$, $1 \le m \le k-3$, be the set of all colliders on $p$ in order of their appearance on $p$, and let $q_i = \langle C_i = Q_{i_1}, \dots, Q_{i_{k_i}} \rangle$ be a shortest causal path from $C_i$ to $\mb{Z}$ in $\g_{\overline{\mb{X} \setminus \{X\}} \underline{X}}$. Further, let $\g[C]$ be the CPDAG that respresents all the DAGs in $[\g]$, and let $p^*$ and $q_i^*$, $i \in \{1, \dots, m \}$, be the paths in $\g[C]$ corresponding to $p$ and $q_i$ in $\g_{\overline{\mb{X} \setminus \{X\}} \underline{X}}$.

    \vskip .1in
    \noindent
    Then if $\mb{C} \neq \emptyset$, the following hold.
    \begin{enumerate}[label = (\roman*)]
        \item\label{q-prop1}
        Each $q_i^*$ is causal.
        
        \item\label{q-prop2}
        $C_i$ is the only node on both $p$ and $q_i$.
        
        \item\label{q-prop3}
        $q_i$ and $q_j$ do not share any nodes, where $i \neq j$.
    \end{enumerate}
\end{lemma}

\begin{proofofnoqed}[Lemma \ref{lem:collider-paths}]    
    \begin{enumerate}
        \item[\ref{q-prop1}]
        Note that none of the nodes on $q_i$ are in $\mb{X}$, as $q_i$ is a path in $\g_{\overline{\mb{X} \setminus \{X\}} \underline{X}}$. Also, the only node on $q_i$ that is in $\mb{Z}$ is $Q_{i_{k_i}}$ (otherwise, we can choose a shorter path). Therefore, $q_{i}^{*}$ must be a possibly directed unshielded path from $C_i$ to $\mb{Z}$ in $\g[C].$ 

        Furthermore, by Lemma \ref{lem:unishielded-colliders}, all colliders on $p$ are unshielded and none of nodes on $p$ are in $\mb{X}\setminus \{X\}$. Hence, let $P_j = C_i$, $j > 2$, then  $P_{j-1} \to P_j \gets P_{j+1}$ is an unshielded collider on $p^{*}$ in $\g[C]$. Then if either $P_{j-1} \notin \Adj(Q_{i_2}, \g[C])$, or $P_{j+1} \notin \Adj(Q_{i_2}, \g[C])$, $Q_{i_1} \to Q_{i_2}$ is in $\g[C]$ by Lemma \ref{lem:prop1meek}. Then in turn,  $q_i^{*}$ must be a directed path in $\g[C]$.
     
        Otherwise, $P_{j-1} \to Q_{i_2} \gets P_{j+1}$ must be in $\g[C]$ (and $\g$). But in this case, depending on whether $Q_{i_2}$ is on $p$, we can derive a contradiction with the choice of $p$.
   
        For instance, if $Q_{i_2}$ is not a node on $p$, then clearly $p(X,P_{j-1}) \oplus \langle P_{j-1}, Q_{i_2}, P_{j+1} \rangle \oplus p(P_{j+1}, Y)$ is a path with exactly the same properties as $p$, but with a shorter distance to $\mb{Z}$, which leads us to a contradiction with the choice fo $p$. Otherwise, $Q_{i_2}$ is on $p$, so either \ref{qonpx} $Q_{i_2}$ is on $p(X,P_{j-1})$, or \ref{qonpy} $Q_{i_2}$ is on $p(P_{j+1},Y)$
        \begin{enumerate}[label=(\alph*)]
            \item\label{qonpx}
            If $Q_{i_2}$ is a collider on $p(X,P_{j-1})$, then $p(X, Q_{i_2}) \oplus \langle Q_{i_2}, P_{j+1} \rangle \oplus p(P_{j+1}, Y)$ gives us the desired path to derive the contradiction. If $Q_{i_2}$ is a definite non-collider on $p$ and if $-p(Q_{i_2}, X)$ is out of $Q_{i_2}$, we again have that $p(X, Q_{i_2}) \oplus \langle Q_{i_2}, P_{j+1} \rangle \oplus p(P_{j+1}, Y)$ gives us the desired contradiction. Otherwise, $Q_{i_2}$ is a definite non-collider on $p$ and $-p(Q_{i_2}, X)$ starts with an undirected edge, that is $Q_{i_2} - P_{u} \dots X$, $1 < u < < j-1$. But then since $P_{j+1} \to Q_{i_2}$ is in $\g[C]$, Lemma \ref{lem:prop1meek}, lets us conclude that $P_{j+1} \to P_{l}$ for every $1 \le l < u$ such that $-p^{*}(P_u, P_l)$ is an undirected path. Then either $P_{j+1} \to X$ is in $\g[C]$, or there exists an $1< r < u$, such that $-p^{*}(P_r, X)$ starts with an edge out of $P_r$, and $P_{j+1} \to P_r$ is in $\g[C].$ Then we have that either $\langle X, P_{j+1} \rangle \oplus p(P_{j+1}, Y)$ (which has one fewer collider than $p$), or $p(X,P_r) \oplus \langle P_r, P_{j+1} \rangle \oplus p(P_{j+1}, Y)$ that leads us to the desired contradiction. 

            \item\label{qonpy}
            If $Q_{i_2}$ is a collider on $p(P_{j+1},Y)$, then $p(X, P_{j-1}) \oplus \langle P_{j-1}, Q_{i_2} \rangle \oplus p(Q_{i_2}, Y)$ gives us the desired path to derive the contradiction. If $Q_{i_2}$ is a definite non-collider on $p(P_{j+1},Y)$ and if $p(Q_{i_2}, Y)$ is out of $Q_{i_2}$, we have that $p(X, P_{j-1}) \oplus \langle P_{j-1}, Q_{i_2} \rangle \oplus p(Q_{i_2}, Y)$ gives us the desired contradiction. Otherwise, $Q_{i_2}$ is a definite non-collider on $p(P_{j+1},Y)$ and $p(Q_{i_2}, Y)$ starts with an undirected edge, that is $Q_{i_2} - P_{u} \dots Y$, $j+1 < u < k$. But then since $P_{j-1} \to Q_{i_2}$ is in $\g[C]$, Lemma \ref{lem:prop1meek}, lets us conclude that $P_{j-1} \to P_{l}$ for every $u< l \le k$ such that $p^{*}(P_u, P_l)$ is an undirected path. Then either $P_{j-1} \to Y$ is in $\g[C]$, or there exists an $u< r <k$, such that $p^{*}(P_r, Y)$ starts with an edge out of $P_r$, and $P_{j-1} \to P_r$ is in $\g[C].$ Then we have that either $p( X, P_{j-1})\oplus \langle P_{j-1}, Y \rangle$, or $p(X, P_{j-1}) \oplus \langle P_{j-1}, P_{r} \rangle \oplus p(P_{r}, Y)$ that leads us to the desired contradiction. 
        \end{enumerate} 
        
        \item[\ref{q-prop2}]
        Next, we show that the only node that $q_i$ and $p$ have in common is $C_i$. Hence, suppose for a contradiction that $Q_{i_d}$, $d \in \{2, \dots, Q_{i_{k_i}}\}$ is the closest node to $C_i$ on $q$ that is also $p$. Then $Q_{i_d}$ is either \ref{case1qid} on $p(X,P_{j-1})$, or \ref{case2qid} on $p(P_{j+1}, Y)$.
        \begin{enumerate}[label = (\alph*)]
            \item\label{case1qid}
            In this case, $Q_{i_d}$ is either a collider or a definite non-collider on $p(X, P_{j-1})$. If $Q_{i_d}$ is a collider on $p(X, P_{j-1})$, then $p(X, Q_{i_d}) \oplus (-q_i( Q_{i_d}, P_{j})) \oplus p(P_{j}, Y)$ gives us the desired path to derive the contradiction. If $Q_{i_d}$ is a definite non-collider on $p(X, P_{j-1})$ and $-p(Q_{i_d}, X)$ is a path that starts with a directed edge out of $Q_{i_d}$, we again have that $p(X, Q_{i_d}) \oplus (-q_i( Q_{i_d}, P_{j})) \oplus p(P_{j}, Y)$ gives us the desired contradiction. Otherwise, $Q_{i_d}$ is a definite non-collider on $p(X, P_{j-1})$ and $-p(Q_{i_d}, X)$ starts with an undirected edge, that is $Q_{i_d} - P_{u} \dots X$, for some $u \in \{1, \dots, j-2\}$. But then, we have  by case \ref{q-prop1} that   $Q_{i_{d-1}} \to Q_{i_d}$ is in $\g[C]$. Then  Lemma \ref{lem:prop1meek}, lets us conclude that $ Q_{i_{d-1}} \to P_{l}$ for every $1 \le l < u$ such that $-p^{*}(P_u, P_l)$ is an undirected path. Then either $ Q_{i_{d-1}} \to X$ is in $\g[C]$, or there exists an $1< r < u$, such that $-p^{*}(P_r, X)$ starts with an edge out of $P_r$, and $ Q_{i_{d-1}} \to P_r$ is in $\g[C].$ Then we have that either $\langle X, Q_{i_{d-1}} \rangle \oplus (-q( Q_{i_{d-1}}, P_{j})) \oplus p(P_{j}, Y)$, or $p(X,P_r) \oplus \langle P_r, Q_{i_2}\rangle \oplus (-q( Q_{i_{d-1}}, P_{j})) \oplus p(P_{j}, Y)$ contradict the choice of $p$, since both of these paths contain one fewer collider compared to $p$.

            \item\label{case2qid}
            In this case, $Q_{i_d}$ is either a collider or a definite non-collider on $p(P_{j+1}, Y)$. If $Q_{i_d}$ is a collider on $p(P_{j+1}, Y)$, then $p(X, P_{j-1}) \oplus q_i(P_{j-1}, Q_{i_d}) \oplus p(Q_{i_d}, Y)$ gives us the desired path to derive the contradiction. If $Q_{i_d}$ is a definite non-collider on $p(P_{j+1}, Y)$ and $p(Q_{i_d}, Y)$ is a path that starts with a directed edge out of $Q_{i_d}$, we again have that $p(X, P_{j}) \oplus q_i(P_{j} Q_{i_d})) \oplus p(Q_{i_d}, Y)$ gives us the desired contradiction. Otherwise, $Q_{i_d}$ is a definite non-collider on $p(P_{j+1}, Y)$ and $p(Q_{i_d}, Y)$ starts with an undirected edge, that is $Q_{i_d} - P_{u} \dots Y$, for some $u \in \{j+2, \dots, k\}$. But then, since   by case \ref{q-prop1} $Q_{i_{d-1}} \to Q_{i_d}$ is in $\g[C]$, Lemma \ref{lem:prop1meek}, lets us conclude that $ Q_{i_{d-1}} \to P_{l}$ for every $u \le l < k$ such that $p^{*}(P_u, P_l)$ is an undirected path. Then either $ Q_{i_{d-1}} \to Y$ is in $\g[C]$, or there exists an $u< r < k$, such that $p^{*}(P_r, Y)$ starts with a directed edge out of $P_r$, and $ Q_{i_{d-1}} \to P_r$ is in $\g[C].$ Then we have that either $p(X, P_{j-1}) \oplus q(P_{j-1},Q_{i_{d-1}}) \oplus \langle Q_{i_{d-1}}, Y \rangle$, or $p(X, P_{j-1}) \oplus q(P_{j-1},Q_{i_{d-1}}) \oplus \langle Q_{i_{d-1}}, P_{r} \rangle \oplus p(P_r, Y)$ contradict the choice of $p$, since both of these paths contain one fewer collider compared to $p$. 
        \end{enumerate}

        \item[\ref{q-prop3}]
        Lastly, we show that $q_i$ and $q_j$ such that $i, j \in \{1, \dots, m\}$, $i<j$ cannot contain a node in common by contradiction. Hence, let $Q_{i_d}$, $d >1$, be the closest node to $Q_{i_1}$ on $q_i$ that is also on $q_j$. Then the path $p(X,C_i) \oplus q_i(C_i, Q_{i_d}) \oplus (-q_j(Q_{i_d}, C_j)) \oplus p(C_j, Y)$ contradicts the choice of $p$ as it contains one fewer collider than $p$. \hfill \BlackBox
    \end{enumerate}
\end{proofofnoqed}


\begin{lemma}
\label{lem:collider-paths2}
    Let $\mb{X,Y,Z}$ be pairwise disjoint node sets in a causal MPDAG $\g = (\mb{V,E})$. Suppose there is no possibly causal path in $\g$ from $X$ to $\mb{Y}$ that starts undirected and does not contain $\mb{X} \setminus \{X\} \cup \mb{Z}$. Further, suppose there is a possibly causal path in $\g$ from $X$ to $\mb{Z}$ that starts undirected and does not contain $\mb{X} \setminus \{X\} \cup \mb{Y}$, where $(X \not \dsepp \mb{Y} | \mb{X} \setminus \{X\}, \mb{Z})_{\g_{\overline{\mb{X} \setminus \{X\}} \underline{X}}}$. Let $r^* = \langle X = R_1, \dots, R_{k_0}\rangle, k_0 >1$, be a shortest such path, and let $r$ be the corresponding sequence of nodes in $\g_{\overline{\mb{X} \setminus \{X\}} \underline{X}}$. Then consider the following sets.
    \begin{align*}
        \mb{S_1} &= \left\{\begin{array}{ll}
                                \text{definite status paths in $\g_{\overline{\mb{X} \setminus \{X\}} \underline{X}}$ from $X$ to $\mb{Y}$ that are} \\
                                \text{d-connecting given $\mb{X} \setminus \{X\} \cup \mb{Z}$}
                           \end{array}\right\}.\\
        \mb{S_2} &= \{ \text{paths in $\mb{S_1}$ with the fewest colliders} \}.\\
        \mb{S_3} &= \{ \text{paths in } \mb{S_2} \text{ that start undirected}\}.\\[.4cm]
        \mb{S_6} &= \{ \text{paths in $\mb{S_2}$ with the fewest edges} \}.\\
        \mb{S_7} &= \{ \text{paths in $\mb{S_6}$ with a shortest distance to $\mb{Z}$ in $\g_{\overline{\mb{X} \setminus \{X\}} \underline{X}}$} \}.
    \end{align*}
    Suppose  $\mb{S_3} = \emptyset$, and pick an arbitrary path $p := \langle X = P_1, \dots, P_k = Y \rangle$, $k > 1$, $Y \in \mb{Y}$ in $\mb{S_7}$. Let $\mb{C} = \{C_1, \dots, C_m \}$, $1 \le m \le k-3$, be the set of all colliders on $p$ in order of their appearance on $p$, and let $q_i := \langle C_i = Q_{i_1}, \dots, Q_{i_{k_i}} \rangle$ be a shortest causal path from $C_i$ to $\mb{Z}$ in $\g_{\overline{\mb{X} \setminus \{X\}} \underline{X}}$. Then the following hold.
    \begin{enumerate}[label = (\roman*)]
        \item\label{rprop0} $r$ is a path in $\g_{\overline{\mb{X} \setminus \{X\}} \underline{X}}$.
        
        \item\label{rprop1} $r^*$ is unshielded in $\g$ and does not contain nodes in $\mb{Z} \setminus \{R_{k_0}\}$.

        \item\label{rprop2} $\g_{\overline{\mb{X} \setminus \{X\}} \underline{X}}$ does not contain $X - R_i$ for any $i \neq 2$.

        \item\label{rprop3} Either $\g_{\overline{\mb{X} \setminus \{X\}} \underline{X}}$ contains $P_2 \to R_2$, or it contains $P_2 - R_2$ where $P_2 \notin \{R_2, \dots, R_{k_0}\}$. Further, if $P_2 \in \Adj(R_i, \g_{\overline{\mb{X} \setminus \{X\}} \underline{X}})$, $i >2$, then $\g$ contains $P_2 \to R_i$ or $P_2 - R_i$.

        \item\label{rprop4p} If $P_i$ is a definite non-collider on $p$, then $R_2 \notin \Adj(P_i, \g)$ for any $i \neq 2$.

        \item\label{rprop4} $X$ is the only node on both $r$ and $p$.

        \item\label{rprop5} $r$ and $q_i$ do not share any nodes.
    \end{enumerate}
\end{lemma}

\begin{proofofnoqed}[Lemma \ref{lem:collider-paths2}]
    \begin{itemize}
        \item[\ref{rprop0}]
        Since $r$ does not contain a node from $\mb{X}\setminus \{X\}$ and since $X - R_2$ is in $\g$, $r$ is a path in $\g_{\overline{\mb{X} \setminus \{X\}} \underline{X}}$.
    
        \item[\ref{rprop1}]
        By choice of $r$, $R_{k_0}$ is the only node on $r$ that is in $\mb{Z}$ and $r$ must be unshielded $\g_{\overline{\mb{X} \setminus \{X\}} \underline{X}}$ otherwise, we can choose a shorter path as $r^{*}$. 

        \item[\ref{rprop2}]
        $X - R_i$ is not in $\g_{\overline{\mb{X} \setminus \{X\}} \underline{X}}$ for any $i \neq 2$, otherwise we can choose a shorter path $r^{*}$.

        \item[\ref{rprop3}]
        Since $r^{*}$ is a possibly causal path, $P_2 \notin \{R_2, \dots, R_{k_0}\}$. Also, since $P_2 \to X- R_2$ is in $\g_{\overline{\mb{X} \setminus \{X\}} \underline{X}}$ and since $P_2 \notin \mb{X}$, we have that $P_2 \in \Adj(R_2, \g_{\overline{\mb{X} \setminus \{X\}} \underline{X}})$ (since otherwise, $\g$ is not an MPDAG). We also cannot have $R_2 \to P_2 \to X$ and $X - R_2$ in $\g$, due to Lemma \ref{lem:nobackpaths}. Then $P_2 \to R_2$ or $P_2 - R_2$ is in $\g$.

        Similarly if $P_2 \in \Adj(R_i, \g_{\overline{\mb{X} \setminus \{X\}} \underline{X}})$ for any $i \in \{3, \dots, k_0\}$, we know that $R_i \to P_2$ cannot be in $\g$ otherwise, $R_i \to P_2 \to X$ and Lemma \ref{lem:nobackpaths} would contradict that $r^{*}$ is a possibly causal path.

        \item[\ref{rprop4p}]
        Suppose for a contradiction that $R_2 \in \Adj(P_i, \g)$ for some definite non-collider $P_i$ on $p$, and $i > 2$. Choose the definite non-collider $P_j$ on $p$ with the largest index $j \in \{3, \dots, k\}$ such that $R_2 \in \Adj(P_j, \g)$. Then consider path $t = \langle X, R_2, P_j \rangle \oplus p(P_j, Y)$. Note that $R_2$ is of definite status on $t$ in $\g_{\overline{\mb{X} \setminus \{X\}} \underline{X}}$, since otherwise $X \gets P_j$ is in $\g$ (due to Lemma \ref{lem:unishielded-colliders}) and $\langle X, P_j \rangle \oplus p(P_j, Y)$ contradicts the choice of $p$. Furthermore, $P_j$ must be of definite status on $t$, otherwise, $R_2 \in \Adj(P_{j+1}, \g_{\overline{\mb{X} \setminus \{X\}} \underline{X}})$ and $R_2 - P_j - P_{j+1}$ or $R_2 - P_j \gets P_{j+1}$ is in $\g_{\overline{\mb{X} \setminus \{X\}} \underline{X}}$, but that contradicts the choice of $P_j$ on $p$. Hence, $t$ is of definite status in $\g_{\overline{\mb{X} \setminus \{X\}} \underline{X}}$. Now, it is enough to show that $t$ is d-connecting given $\mb{Z}$ to derive the contradiction. Hence, if $P_j$ is a non-collider on $t$, we have our contradiction. Otherwise, $P_j$ is a collider on $t$, which further means that $P_{j-1} \gets P_j \gets P_{j+1}$ is on $p$ and $R_2 \to P_j$ is on $t$. Note that $X \gets \dots \gets P_j$ cannot be the form of $-p(P_j, X)$, otherwise $\langle R_2, P_j \rangle \oplus (-p(P_j, R_2)$ and $X - R_2$ contradict Lemma \ref{lem:nobackpaths}. Hence, there must be collider that is a descendant of $P_j$ on $p(X, P_j)$, which further implies that $P_j \in \An(\mb{Z}, \g_{\overline{\mb{X} \setminus \{X\}} \underline{X}})$. Therefore, $t$ is d-connecting given $\mb{Z}$.

        \item[\ref{rprop4}]
        Next, we show that the only node on both $p$ and $r$ is $X$. We will assume for a contradiction that there is a node on $p$ and $r$ that is not $X$. Hence let $P_u$ be chosen as the node on $p$ with the largest index $u \in \{3, \dots, k\}$ that is also on $r$, let $P_u \equiv R_l, l \in \{3, \dots, k_0\}$. Then consider path $t = r(X, P_u) \oplus p(P_u, Y)$. If $t$ is a path is of definite status and $P_u$ is a definite non-collider on $t$, then as $t \in \mb{S_3}$, we have a contradiction with the assumption that  $\mb{S_3} =\emptyset$.  If $t$ is a path of definite status and $P_u$ is a collider on $t$, we still obtain a contradiction as long as there is a collider on $p(X,P_u)$. Note that there must be such a collider on $p(X,P_u)$ otherwise, $p(X, P_u)$ is of the form $X \gets \dots \gets P_u$ and since $P_u \equiv R_l$ and $r^{*}(X,R_l)$ is a possibly causal path from $X$ to $R_l$ we would have a contradiction with Lemma \ref{lem:nobackpaths}. 

        Otherwise, $t$ is not of definite status due to $P_u$, and $R_{l-1} - P_u - P_{u+1}$ or $R_{l-1} - P_u \gets P_{u+1}$ and $\langle R_{l-1}, P_u \rangle$ is in $\g_{\overline{\mb{X} \setminus \{X\}} \underline{X}}.$ Let $P_{d}$, $d \in \{u+1, \dots, k\}$ be chosen so that $P_{d+1}$ is the closest collider to $P_u$ on $p(P_u, Y)$ or if there is no collider on $p(P_u, Y)$, then let, $P_d = Y$. Consider that none of the nodes among $P_{u+1}, \dots , P_d$ are adjacent to $X$ as that would imply that $X - P_h$ or $X \gets P_h$, for $h \in \{u+1, \dots, d \}$ is in $ \g_{\overline{\mb{X} \setminus \{X\}} \underline{X}}$ and neither of that is possible (the former due to Lemma \ref{lem:unishielded-colliders}, and the latter due to the choice of $p$, as $\langle X, P_h \rangle \oplus p(P_h, Y)$ has all the same properties as $p$ but is shorter). 

        Hence, choose as $P_h$ $h \in \{u+1, \dots, d \}$ a node with the largest index that is adjacent to a node on $r(X,R_{l-1})$. Next, choose as $R_o$, $o \in \{1, \dots, l-1\}$ a node with the smallest index such that $R_o \in \Adj(P_h,\g_{\overline{\mb{X} \setminus \{X\}} \underline{X}})$. Then consider path $t = r(X,R_{o}) \langle R_{o}, P_{h} \rangle \oplus p(P_{h} ,Y)$. Path $t$ has the same properties as $p$ but starts with an undirected edge, which contradicts that $\mb{S_3} = \emptyset$.

        \item[\ref{rprop5}] Lastly we show that there is no node that is common to $r$ and $q_i$ for any $i \in \{1, \dots, m\}$. Suppose for a contradiction that there exists $i \in \{1, \dots , m\}$ such that $q_i$ and $r$ have a node in common and choose the smallest such $i$. By above, we have that $Q_{i_1}$ is not on $r$, so let $Q_{i_l}$ be the closest node to $Q_{i_1}$ on $q_i$ that is on both $r$ and $q_i$, $Q_{i_l} \equiv R_d$, $d \in \{2, \dots, k_0\}$. Consider path $t = r(X,Q_{i_l}) \oplus (-q(Q_{i_l}, Q_{i_1})) \oplus p(Q_{i_1}, Y)$. If $t$ is of definite status, then $t \in \mb{S_3}$ which contradicts our assumptions about $\mb{S_3} = \emptyset$. Otherwise, $R_{d-1} - Q_{i_l} \gets Q_{i_{l-1}}$ is in $\g_{\overline{\mb{X} \setminus \{X\}} \underline{X}}$ and by Lemma \ref{lem:collider-paths} also in the CPDAG $\g[C]$. Hence, by Lemma \ref{lem:prop1meek}, $R_{d-1} \gets Q_{i_{l-1}}$. Furthermore, since $r^{*}$ is a possibly causal path and $r$ is an unshielded path $X - \dots - R_{d-1}$ is in $\g_{\overline{\mb{X} \setminus \{X\}} \underline{X}}$. But this would then imply that $X \gets Q_{i_{l-1}}$ by successive applications of Lemma \ref{lem:prop1meek}. However, now we have that, $\langle X, Q_{i_{l-1}} \rangle \oplus (-q(Q_{i_{l-1}}, Q_{i_1})) \oplus p(Q_{i_1}, Y)$ contradicts the choice of $p$. \hfill \BlackBox
    \end{itemize}
\end{proofofnoqed}

\section{A Note on Jaber's (2022) \texttt{CIDP} Algorithm}
\label{rem-cidp}

Despite the generality of PAGs, the \texttt{CIDP} algorithm of \cite{jaber2022causal} does not hold in the general MPDAG setting. We demonstrate this in the example below, where the \texttt{CIDP} algorithm cannot identify an identifiable conditional effect given an MPDAG.

\begin{example}
\label{ex:cidp}
    Reconsider Example \ref{ex:id-rule3}, where a causal MPDAG $\g$ is known. We will attempt to use a naive translation of \citeauthor{jaber2022causal}'s \texttt{CIDP} algorithm to identify the conditional effect of $X$ on $Y$ given $Z$ in this setting. Start by defining $\mb{D} = \PossAn(\mb{Y} \cup \mb{Z}, \g_{\mb{V} \setminus \mb{X}})$ so that $\mb{D} = \{V_1, V_2, V_3, Y, Z\}$. Then define the ordered bucket decomposition (Definition \ref{def:obd}) of $\mb{V}$ in $\g$ as $(\mb{B_1},\mb{B_2}, \mb{B_3}, \mb{B_4}) = (\{V_1\}, \{Y\}, \{X, V_2, V_3 \}, \{Z\})$. Since $\mb{B_3} \cap \mb{D} \neq \emptyset$ and $\mb{B_3} \not\subseteq \mb{D}$, we enter the \texttt{while} loop of the \texttt{CIDP} algorithm. But $\mb{B_3} \cap \{X\} = \{X\}$, where $X \not \dsepp Y | Z$ in $\g_{\underline{X}}$. Thus, the algorithm outputs a \texttt{FAIL}, implying the conditional effect is not identifiable in $\g$. However, we showed in Example \ref{ex:id-rule3} that $f(y \,|\, do(x),z) = f(y \,|\, z)$.
\end{example}

\vskip 0.2in
\bibliographystyle{apalike}
\bibliography{main}

\end{document}